%% file: neurips_cameraready.tex
\documentclass{article}

\usepackage[nonatbib,final]{neurips_2022}

\usepackage[utf8]{inputenc} 
\usepackage[T1]{fontenc}    
\usepackage{url}            
\usepackage{booktabs}       
\usepackage{amsfonts}       
\usepackage{nicefrac}       
\usepackage{microtype}      
\usepackage{xcolor}         

\usepackage{amsmath,amssymb,amsthm,commath,esint,tikz-cd,tikz, mathtools, physics}
\usetikzlibrary{shapes,arrows,positioning}
\usepackage[mathscr]{euscript}
\usepackage{graphicx}
\usepackage{float}
\usepackage{fancyhdr}
\usepackage{bm}
\usepackage{bbm}
\usepackage[ruled,vlined]{algorithm2e}

\numberwithin{equation}{section}

\input{custom_definitions}

\title{Generic bounds on the approximation error for physics-informed (and) operator learning}

\author{
Tim De Ryck
\thanks{Seminar for Applied Mathematics (SAM), D-MATH, ETH Z\"urich, Switzerland} \AND
Siddhartha Mishra
\thanks{
Seminar for Applied Mathematics (SAM), D-MATH and ETH AI center, 
ETH Zürich, Switzerland
}
}

\begin{document}

\maketitle

\begin{abstract}
We propose a very general framework for deriving rigorous bounds on the approximation error for physics-informed neural networks (PINNs) and operator learning architectures such as DeepONets and FNOs as well as for physics-informed operator learning. These bounds guarantee that PINNs and (physics-informed) DeepONets or FNOs will efficiently approximate the underlying solution or solution operator of generic partial differential equations (PDEs). Our framework utilizes existing neural network approximation results to obtain bounds on more involved learning architectures for PDEs. We illustrate the general framework by deriving the first rigorous bounds on the approximation error of physics-informed operator learning and by showing that PINNs (and physics-informed DeepONets and FNOs) mitigate the curse of dimensionality in approximating nonlinear parabolic PDEs. 
\end{abstract}

\section{Introduction}


 The efficient numerical approximation of partial differential equations (PDEs) is of paramount importance as PDEs mathematically describe an enormous range of interesting phenomena in the sciences and engineering. Machine learning techniques, particularly deep learning, are playing an increasingly important role in this context. For instance, given their universal approximation properties, deep neural networks serve as ansatz spaces for supervised learning of a variety of (parametric) PDEs \cite{HEJ1,SZ1, Kuty,LMR1,LMPR1} and references therein. In this setting, large amounts of training data might be required. However, this data is often acquired from expensive computer simulations  or physical measurements \cite{LMR1}, necessitating the design of learning frameworks that work with limited data.  \emph{Physics-informed neural networks (PINNs)}, proposed by \cite{DPT,Lag1,Lag2} and popularized by \cite{KAR1,KAR2}, 
are a prominent example of such a learning framework as the residual of the underlying PDE is minimized within the class of neural networks and in principle, little (or even no) training data is required. PINNs and their variants have proven to be a very powerful and computationally efficient framework for approximating solutions to PDEs, \cite{KAR4,KAR5,KAR6,KAR7,KAR8,jag1,jag2,MM1,MM2,MM3,BKMM1} and references therein.

Often in the context of PDEs, one needs to approximate the underlying solution operator that maps one infinite-dimensional function space into another \cite{guss2019universal,NO}. As neural networks can only map between finite dimensional spaces, a new field of \emph{operator learning} is emerging wherein novel learning frameworks need to be designed in order to approximate operators. These include deep operator networks (DeepONets) \cite{ChenChen1995,deeponets} and their variants as well as \emph{neural operators} \cite{NO}, which generalize neural networks to this setting. A variety of neural operators have been proposed, see \cite{GKO,Mpole} but arguably, the most efficient form of neural operators is provided by the so-called \emph{Fourier neural operators} (FNOs) \cite{FNO}. Both DeepONets and FNOs have been very successfully deployed in scientific computing \cite{fair,donet1,donet2,donet3,FNO1,FNO2} and references therein. Finally, one can combine PINNs and operator learning to design physics-informed DeepONets/FNOs \cite{wang2021learning, FNO1,wang2021long, goswami2022physics}. 

From a theoretical perspective, one needs to provide a rigorous guarantee that the learning framework can approximate the underlying PDE solution (operator) to desired accuracy. More precisely, given an error tolerance $\epsilon > 0$, we need to rigorously prove that the approximation error of the neural network (operator) can be made smaller than $\epsilon$. For \emph{efficient approximation}, one has to further ensure that the computational complexity (measured in terms of the size) of the learning architecture grows at most \emph{polynomially} in $\epsilon^{-1}$. In particular, \emph{exponential} growth has to be ruled out. As neural networks, DeepONets and FNOs are all \emph{universal approximators} \cite{cybenko1989approximation, ChenChen1995, deeponets, LMK1, kovachki2021universal} of the underlying functions or operators, it is possible to show that the approximation error can be made as small as desired. However, these results do not guarantee efficient approximation as the underlying network size could still grow exponentially with decreasing error, see \cite{yarotsky2017error} for neural networks in very high spatial dimensions, \cite{LMK1} for DeepONets and \cite{kovachki2021universal} for FNOs. Hence, the real theoretical challenge in this context lies in proving efficient approximation results for the different learning architectures in scientific computing. Such efficient approximation results have mostly been obtained for neural networks in the supervised learning setting e.g. \cite{grohs2018proof, jentzen2018proof, hutzenthaler2020proof, beck2020overcoming_elliptic, opschoor2020deep} and references therein. In contrast, there is a relative scarcity of such efficient approximation results for PINNs and operator learning with notable exceptions being  \cite{guhring2020error, deryck2021approximation, deryck2021pinn} (for PINNs), \cite{LMK1} (for DeepONets) and  \cite{kovachki2021universal} (for FNOs). Moreover, the underlying proofs in these works are often on a case-by-case basis and the overall abstract structure is not clearly identified. Finally, no similar rigorous approximation results for physics informed operator learning are available till date.

This paucity of generic efficient approximation results for PINNs and operator learning for PDEs sets the stage for the current paper where our main contribution is to propose a very general \emph{framework} (Section \ref{sec:general}) for proving bounds on the approximation error for space-time neural networks, PINNs, DeepONets, FNOs and physics-informed DeepONets and FNOs for very general PDEs. Consequently, we obtain the first rigorous bounds for physics-informed operator learning in literature. 
Our framework is based on the observation that error estimates for different types of neural network architectures can all be obtained from one another. As error estimates for neural network approximations of PDE solutions at a fixed time are the easiest to obtain, and hence constitute the largest proportion of currently available estimates, we devote particular attention to demonstrating how these available estimates can be used to obtain novel bounds on the approximation error for space-time networks, PINNs and (physics-informed) operator learning. 
Our results provide a \emph{roadmap} for deriving mathematical guarantees for deep learning methods in scientific computing by simplifying the proofs, as the needed work essentially reduces to verifying a small number of assumptions. We demonstrate how the generic error bounds from Section \ref{sec:general} can be applied in practice in Section \ref{sec:applications}, among others by giving short alternative proofs for known results and also proving a number of novel results. In particular, we show in Section \ref{sec:CoD} that PINNs can overcome the curse of dimensionality for nonlinear parabolic PDEs such as the Allen-Cahn equation i.e., that the network size does not grow exponentially with increasing spatial dimension. Moreover, dimension-independent convergence rates are also obtained for (physics-informed) DeepONets and FNOs, provided that the PDE solutions are sufficiently smooth. These are the first results of their kind. We note that many of the proofs and some examples are deferred to the supplementary material (\textbf{SM}). 
\section{Preliminaries}\label{sec:preliminaries}

\subsection{Setting}\label{sec:setting}

Given $T>0$ and $D\subset \R^d$ compact, consider the function $u:[0,T]\times D\to \R^m$, for $m \geq 1$, that {\color{black} belongs to a function space $\cH$} and solves the following (time-dependent) PDE,
\begin{equation}\label{eq:PDE}
    \cL_a(u)(t,x) = 0\quad\text{and}\quad u(x, 0) = u_0\qquad \forall (t,x)\in[0,T]\times D, 
\end{equation}
where $u_0\in \cY \subset L^2(D)$ is the initial condition and $\cL_a{\color{black}:\cH\to L^2([0,T]\times D)}$ is a differential operator that can depend on a parameter (function) $a\in\cZ \subset L^2(D)$. In our notation, we will often suppress the dependence of $\cL:=\cL_a$ on $a$ for simplicity. Depending on the context, one might want to recover one of the following mathematical objects: for fixed $a$ and $u_0$, one might want to approximate $u(T,\cdot)$ or $u (\cdot,\cdot)$ with a neural network; a more challenging task would be to learn the \emph{solution operator} $\cG: \cX \to L^2(\Omega):v\mapsto u$, where $v\in \{u_0,a\}$,  $\cX\in\{\cY,\cZ\}$ and $\Omega=D$ or $\Omega=[0,T]\times D$. We will use this notation consistently throughout the paper, see \textbf{SM} \ref{app:glossary} for an overview.

\subsection{Approximating PDEs with neural networks}\label{sec:pde-nn}

\paragraph{Neural networks}\label{sec:prel-nn}

A (feedforward) neural network $u_{\theta}:\R^{d_0}\to\R^{d_L}$ is defined as a concatenation of affine maps $\cA_l: \R^{ d_{l-1}}\to \R^{d_{l}}: z\mapsto W_lz+b_l$
and an activation function $\sigma:\R\to\R$ that is applied component-wise, resulting in,
\begin{equation}
\label{eq:ann1}
u_{\theta}(y) = \cA_L \circ\sigma \circ \cA_{L-1}\ldots \ldots \ldots \circ\sigma \circ \cA_2 \circ \sigma \circ \cA_1(y).
\end{equation} 
The weights and biases of the affine maps $\theta = \{W_l, b_l\}_{1\leq l \leq L}$ are the trainable parameters. We will quantify the size of a neural network by its $\depth(u_\theta) := L$ and its $\width(u_\theta) := \max_l d_l$.
In order to obtain a neural network that approximates the solution $u$ of PDE \eqref{eq:PDE} at time $t=T$, one chooses the parameters of $u_\theta:D\to \R$ such that a discretization (quadrature) of $\cJ(\theta) =\Vert u(T)-u_\theta\Vert_{L^2(D)}$ is minimized. 
The training data is acquired from either measurements or potentially expensive simulations. 

\paragraph{PINNs}\label{sec:prel-pinn} Physics-informed neural networks (PINNs) are neural networks that are trained with a different, residual-based loss function. As the PDE solution $u$ satisfies $\cL(u)=0$, the goal of physics-informed learning is to find a neural network $u_\theta:[0,T]\times D\to \R$ for which the PDE residual is approximately zero, $\cL(u_\theta)\approx 0$. To ensure uniqueness, one also needs to require that the initial condition is satisfied i.e., $u_\theta(0,x) \approx u_0(x)$, and similarly for boundary conditions. In practice one minimizes a quadrature approximation of $\cJ(\theta) =\Vert\cL(u_\theta)\Vert_{L^2([0,T]\times D)}^2+\Vert u_\theta(0,\cdot)-u_0\Vert_{L^2(D)}^2$,
where additional terms can be added to (approximately) impose boundary conditions and augment the loss function using data. A desirable property of PINNs is that only very little or even no training data is needed to construct the loss function. 

\paragraph{Operator learning}\label{sec:prel-deeponets}

In order to approximate operators, one needs to allow the input and output of the learning architecture to be infinite-dimensional. A possible approach is to use \emph{deep operator networks} (DeepONets), as proposed in \cite{ChenChen1995, deeponets}. 
Given $m$, fixed sensor locations $\{x_j\}_{j=1}^m \subset D$ and the corresponding \emph{sensor values} $\{v(x_j)\}_{j=1}^m$ as input,
a DeepONet can be formulated in terms of two (deep) neural networks: a \emph{branch net} $\branch: \R^m \to \R^{p}$ and a \emph{trunk net} $\trunk :D\to\R^{p+1}$.
The branch and trunk nets are then combined to approximate the underlying nonlinear operator as the following \emph{DeepONet} $\cG_\theta:\cX \to L^2(D)$, with $\cG_\theta(v)(y) = \tr_0(y)+\sum_{k=1}^p \beta_k(v) \tr_k(y)$. A second approach is that of \emph{neural operators}, which generalize hidden layers by including a non-local integral operator \cite{GKO}, of which particularly \emph{Fourier neural operators} (FNOs) \cite{FNO} are already well-established. The practical implementation (i.e. discretization) of an FNO maps from and to the space of trigonometric polynomials of degree at most $N\in\N$, denoted by $L^2_N$, and can be identified with a finite-dimensional mapping that is a composition of affine maps and nonlinear layers of the form $\mathfrak{L}_l(z)_j = \sigma(W_l v_j + b_{l,j}
\cF^{-1}_N(P_l(k)\cdot \cF_N(z)(k)_j))$, where the $P_l(k)$ are coefficients that define a non-local convolution operator via the discrete Fourier transform $\cF_N$, see \cite{kovachki2021universal}. 

\paragraph{Physics-informed operator learning} Both DeepONets and FNOs are trained by choosing a suitable probability measure $\mu$ on $\cX$ and minimizing a quadrature approximation of $\cJ(\theta) = \Vert\cG_\theta(v)-\cG(v)\Vert_{L^2_{\mu \times dx}(\cX\times \Omega)}$. Generating training sets might require many calls to an expensive PDE solver, leading to an enormous computational cost. In order to reduce or even fully eliminate the need for training data, \emph{physics-informed operator learning} has been proposed in \cite{wang2021learning} for DeepONets and in \cite{FNO1} for FNOs. Similar to PINNs, the training procedure aims to minimize a quadrature approximation of $\cJ(\theta) =\Vert\cL(\cG_\theta)\Vert_{L^2_{\mu \times dx}(\cX\times \Omega)}$.

\section{General results}\label{sec:general}

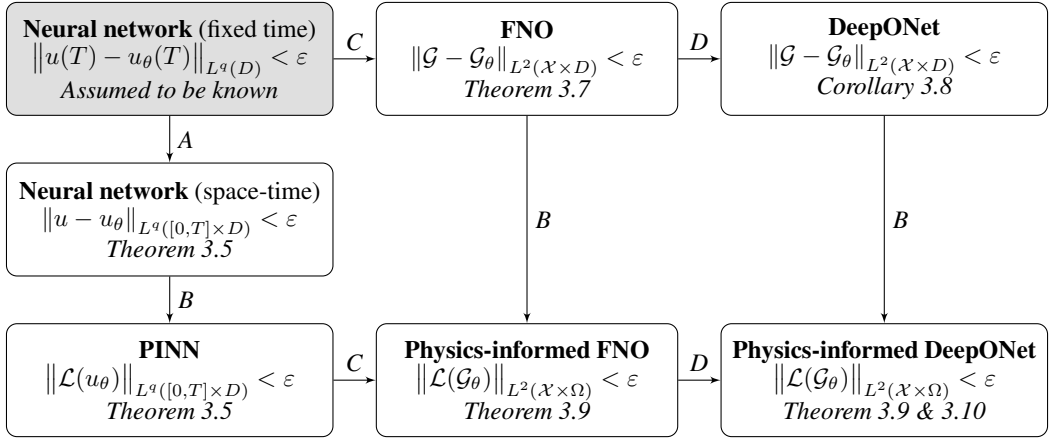
\begin{figure}
\tikzstyle{block} = [rectangle, draw, minimum height=4.5em, minimum width = 13em,
    text centered, rounded corners]
\tikzstyle{block2} = [rectangle, draw, minimum height=4.5em, minimum width = 13em,
    text centered, rounded corners,fill={rgb:black,1;white,7}]
\tikzstyle{block3} = [rectangle, draw, minimum height=4.5em, minimum width = 12em,
    text centered, rounded corners]

\tikzstyle{line} = [draw, -latex']
\centering

\begin{center}

\begin{tikzpicture}[scale=0.95, every node/.style={transform shape}]

\def\tolefta{5}
\def\toleftb{5}
\def\todown{2.25}

\node[block2, align=center] at (0,0) (nnt) {\textbf{Neural network} (fixed time) \\ $\norm{u(T)-u_\theta(T)}_{L^q(D)}<\epsilon$ \\ \emph{Assumed to be known}};
\node[block3, align=center] at (\tolefta,0) (fno) {\textbf{FNO} \\ $\norm{\cG-\cG_\theta}_{L^2(\cX\times D)}<\epsilon$ \\ \emph{Theorem \ref{thm:nn-to-fno}}};
\node[block, align=center] at (\tolefta+\toleftb,0) (deeponet) {\textbf{DeepONet} \\ $\norm{\cG-\cG_\theta}_{L^2(\cX\times D)}<\epsilon$ \\ \emph{Corollary \ref{cor:nn-to-deeponet}}};
\node[block, align=center] at (0,-\todown) (nn) {\textbf{Neural network} (space-time) \\ $\norm{u-u_\theta}_{L^q([0,T]\times D)}<\epsilon$ \\ \emph{Theorem \ref{thm:nn-to-pinn}}};
\node[block, align=center] at (0,-\todown-\todown) (pinn) {\textbf{PINN} \\ $\norm{\cL(u_\theta)}_{L^q([0,T]\times D)}<\epsilon$ \\ \emph{Theorem \ref{thm:nn-to-pinn}}};
\node[block3, align=center] at (\tolefta,-\todown-\todown) (pino) {\textbf{Physics-informed FNO} \\ $\norm{\cL(\cG_\theta)}_{L^2(\cX\times \Omega)}<\epsilon$ \\ \emph{Theorem \ref{thm:deeponet-to-pido}}};
\node[block, align=center] at (\tolefta+\toleftb,-\todown-\todown) (pido) {\textbf{Physics-informed DeepONet} \\ $\norm{\cL(\cG_\theta)}_{L^2(\cX\times\Omega)}<\epsilon$ \\ \emph{Theorem \ref{thm:deeponet-to-pido} \& \ref{thm:nn-to-pido}}};

\path [line] (nnt) -- (nn) node [midway, right] {\emph{A}};
\path [line] (nn) -- (pinn) node [midway, right] {\emph{B}};
\path [line] (fno) -- (pino) node [midway, right] {\emph{B}};
\path [line] (deeponet) -- (pido) node [midway, right] {\emph{B}};
\path [line] (nnt) -- (fno) node [midway, above] {\emph{C}};
\path [line] (pinn) -- (pino) node [midway, above] {\emph{C}};
\path [line] (fno) -- (deeponet) node [midway, above] {\emph{D}};
\path [line] (pino) -- (pido) node [midway, above] {\emph{D}};

\end{tikzpicture}
\end{center}
    \caption{Flowchart of the structure of the results in this paper, with $q\in\{2,\infty\}$. The letters reflect the techniques used in the proofs: \emph{A} uses Taylor approximations (Section \ref{sec:taylor}), \emph{B} is based on finite difference approximations (Section \ref{sec:taylor}), \emph{C} uses trigonometric polynomial interpolation (Section \ref{sec:fourier}) and \emph{D} uses the connection between FNOs and DeepONets (Section \ref{sec:fourier}). }
    \label{fig:flowchart}
\end{figure}

We propose a framework to obtain bounds on the approximation error for the various neural network architectures introduced in Section \ref{sec:pde-nn}. Figure \ref{fig:flowchart} visualizes how different types of error estimates can be obtained from one another. Every box shows the name of the network architecture, the form of the relevant loss and the theorem which proves the corresponding estimate for the approximation error. Every arrow in the flowchart represents a proof technique that allows one to transfer an error estimate from one type of method to another (see caption of Figure \ref{fig:flowchart} for an overview of those techniques). 

We give particular attention to the case where it is known that a neural network can efficiently approximate the solution to a time-dependent PDE at a fixed time. Such neural networks are usually obtained by emulating a classical numerical method. Examples include finite difference schemes, finite volume schemes, finite element methods, iterative methods and Monte Carlo methods, e.g. {\color{black} \cite{jentzen2018proof,opschoor2020deep,chen2021representation,marwah2021parametric}}.
More precisely, for $\epsilon>0$, we assume to have access to an operator {\color{black}$\cU^\epsilon:\cX\times [0,T]\to \cH$} that for any $t\in[0,T]$ maps any initial condition/parameter function $v\in \cX$ to a neural network $\cU^\epsilon(v,t)$ that approximates the PDE solution $\cG(v)(\cdot, t) = u(\cdot, t) {\color{black}\in L^q(D), q\in\{2,\infty\},}$ at time $t$, as specified below. Moreover, we will assume that we know how its size depends on the accuracy $\epsilon$. Explicit examples of the operator $\cU^\epsilon$ will be given in Section \ref{sec:applications} and \textbf{SM} \ref{app:AC}.

\begin{assumption}\label{ass:nn}
Let $q\in\{2,\infty\}$. For any $B,\epsilon>0$, $\ell \in\N$, $t\in [0,T]$ and any $v\in \cX$ with $\norm{v}_{C^\ell}\leq B$ there exist a neural network $\cU^\epsilon(v,t):D\to\R$ and a constant $C_{\epsilon,\ell}^B>0$ s.t.
\begin{equation}
    \norm{\cU^\epsilon(v,t)-\cG(v)(\cdot,t)}_{L^q(D)} \leq \epsilon \quad \text{and} \quad \max_{t\in [0,T]} \norm{\cU^\epsilon(v,t)}_{W^{\ell,q}(D)} \leq C_{\epsilon,\ell}^B.
\end{equation}
\end{assumption}

\begin{remark}
For vanilla neural networks and PINNs one can set $\cX := \{v\}$, $\cG(v) := u$ and $v:=u_0$ or $v:=a$ in Assumption \ref{ass:nn} above and Assumption \ref{ass:diff-operator} below.
\end{remark}

Under this assumption, we prove the existence of space-time neural networks and PINNs that efficiently approximate the PDE solution (Section \ref{sec:taylor}), as well as FNOs and DeepONets (Section \ref{sec:fourier}) and physics-informed FNOs and DeepONets (Section \ref{sec:FD}). Finally, we also prove a general result on the generalization error (Section \ref{sec:generalization}).

\subsection{Estimates for (physics-informed) neural networks}\label{sec:taylor}

We will construct a space-time neural network $u_\theta$ for which both $\Vert u_\theta-u\Vert_{L^q([0,T]\times D)}$ and the PINN loss $\Vert\cL(u_\theta)\Vert_{L^q([0,T]\times D)}$ are small. To accurately approximate the time derivatives of $u$ we emulate Taylor expansions, whereas for the spatial derivatives we employ finite difference (FD) operators in our proofs. 
Depending on whether forward, backward or central differences are used, a FD operator might not be defined on the whole domain $D$, e.g. for $f\in C([0,1])$ the (forward) operator $\Delta^+_h[f] := f(x+h)-f(x)$ is not well-defined for $x\in (1-h,1]$. This can be solved by resorting to piecewise-defined FD operators, e.g. a forward operator on $[0,0.5]$ and a backward operator on $(0.5,1]$. In a general domain $\Omega$ one can find a well-defined piecewise FD operator if $\Omega$ satisfies the following assumption, which is satisfied by many domains (e.g. rectangular, smooth). 
\begin{assumption}\label{ass:partition}
There exists a finite partition $\cP$ of $\Omega$ such that for all $P\in\cP$ there exists $\epsilon_P>0$ and $v_P\in B^1_\infty = \{x\in\R^{\dim(\Omega)}: \norm{x}_\infty\leq 1\}$ such that for all $x\in P$ it holds that $x+\epsilon_P(v_P+B^1_\infty) \subset \Omega$. 
\end{assumption}
Additionally, we need to assume that the PINN error can be bounded in terms of the errors related to all relevant partial derivatives, denoted by $D^{(k,\bmalpha)}:=D^k_tD^\bmalpha_x := \partial_t^k \partial^{\alpha_1}_{x_1}\ldots \partial^{\alpha_d}_{x_d}$, for $(k,\bmalpha)\in\N_0^{d+1}$. {\color{black} This assumption is valid for many classical solutions of PDEs. A few worked out examples can be found in \textbf{SM} \ref{app:pendulum} (gravity pendulum) and \textbf{SM} \ref{app:darcy} (Darcy flow). }
\begin{assumption}\label{ass:diff-operator}
Let $k,\ell\in\N$,  $q\in\{2,\infty\}$, $C>0$ be independent from $d$. For all $v\in\cX$ it holds,
\begin{equation}
    \norm{\cL(\cG_\theta(v))}_{L^q([0,T]\times D)} \leq C \cdot \poly(d)\cdot 
    \sum_{\substack{(k',\bmalpha)\in\N^{d+1}_0\\ k'\leq k, \norm{\bmalpha}_1\leq \ell }} \norm{D^{(k',\bmalpha)}(\cG-\cG_\theta)}_{L^q([0,T]\times D)}. 
\end{equation}
\end{assumption}

In this setting, we prove the following approximation result for space-time networks and PINNs.
\begin{theorem}\label{thm:nn-to-pinn}
Let $s,r\in\N$, let $u\in C^{(s,r)}([0,T]\times D)$ be the solution of the PDE \eqref{eq:PDE} and let Assumption \ref{ass:nn} be satisfied. There exists a constant $C(s,r)>0$ such that for every $M\in\N$ and $\epsilon, h>0$ there exists a tanh neural network $u_\theta:[0,T]\times D\to \R$ for which it holds that,
\begin{equation}
\label{eq:pinnerr1}
    \norm{u_\theta-u}_{L^q([0,T]\times D)} \leq C(\norm{u}_{C^{(s,0)}}M^{-s}+\epsilon). 
\end{equation}
and if additionally Assumption \ref{ass:partition} and Assumption \ref{ass:diff-operator} hold then,
\begin{equation}\label{eq:pinn-acc}
\begin{split}
     &\norm{\cL(u_\theta)}_{L^2([0,T]\times D)} + \norm{u_\theta-u}_{L^2(\partial([0,T]\times D))} \\
     &\qquad \leq C\cdot \poly(d)\cdot \ln^k(M)(\norm{u}_{C^{(s,\ell)}}M^{k-s}+M^{2k}(\epsilon h^{-\ell}+C^B_{\epsilon,\ell}h^{r-\ell})). 
\end{split}
\end{equation}
Moreover, $\depth(u_\theta)\leq C\cdot\depth(\cU^\epsilon)$ and $\width(u_\theta) \leq CM\cdot \width(\cU^\epsilon)$. 
\end{theorem}

\begin{proof}
We only provide a sketch of the full proof (\textbf{SM} \ref{app:nn-to-pinn}). The main idea is to divide $[0,T]$ into $M$ uniform subintervals and construct a neural network that approximates a Taylor approximation in time of $u$ in each subinterval. In the obtained formula, we approximate the monomials and multiplications by neural networks (\textbf{SM} \ref{app:nnt}) and approximate the derivatives of $u$ by finite differences and use \eqref{eq:accuracy-FD} of \textbf{SM} \ref{app:FD} to find an error estimate in $C^k([0,T], L^q(D))$-norm. We use again finite difference operators to prove that spatial derivatives of $u$ are accurately approximated as well. The neural network will also approximately satisfy the initial/boundary conditions as $\Vert u_\theta-u\Vert_{L^2(\partial([0,T]\times D))} \lesssim C \poly(d) \Vert u_\theta-u\Vert_{H^1([0,T]\times D)}$, which follows from a Sobolev trace inequality. 
\end{proof}
We note that the bounds \eqref{eq:pinnerr1} and \eqref{eq:pinn-acc} together imply that there exists a neural network for which the total error as well as the PINN loss can be made as small as possible, providing a solid theoretical foundation to PINNs for approximating the PDE \eqref{eq:PDE}. 

\subsection{Estimates for operator learning}\label{sec:fourier}

In this section, we use Assumption \ref{ass:nn} to prove estimates for DeepONets and FNOs. First, we prove a generic error estimate for FNOs. Using the known connection between FNOs and DeepONets ({\bf SM} Lemma \ref{lem:equivalence}) this result can then easily be applied to DeepONets
(Corollary \ref{cor:nn-to-deeponet}). 
In order to prove these error estimates, we need to assume that the operator $\cU^\epsilon$ from Assumption \ref{ass:nn} is stable with respect to its input function, as specified in Assumption \ref{ass:stability} below. Moreover, we will take the $d$-dimensional torus as domain $D=\T^d = [0,2\pi)^d$ and assume periodic boundary conditions for simplicity in what follows. This is not a restriction, as for every Lipschitz subset of $\T^d$ there exists a (linear and continuous) $\T^d$-periodic extension operator of which also the derivatives are $\T^d$-periodic \cite[Lemma 41]{kovachki2021universal}. 
\begin{assumption}\label{ass:stability}
Assumption \ref{ass:nn} is satisfied and let $p\in\{2,\infty\}$. For every $\epsilon>0$ there exists a constant $C^\epsilon_{\mathrm{stab}}>0$ such that for all $v,v'\in \cX$ it holds that,
\begin{equation}
    \norm{\cU^\epsilon(v,T)-\cU^\epsilon(v',T)}_{L^2}\leq C^\epsilon_{\mathrm{stab}} \norm{v-v'}_{L^p}. 
\end{equation}
\end{assumption}

In this setting, we prove a generic approximation result for FNOs. 

\begin{theorem}\label{thm:nn-to-fno}
Let $r\in\N$, $T>0$, let $\cG: C^r(\T^d)\to C^r(\T^d)$ be an operator that maps a function $u_0$ to the solution $u(\cdot, T)$ of the PDE \eqref{eq:PDE} with initial condition $u_0$, let Assumption \ref{ass:stability} be satisfied {\color{black}and let $p^*\in\{2,\infty\}\setminus\{p\}$}. Then there exists a constant $C>0$ such that for every $\epsilon>0$, $N\in\N$ there is an FNO $\cG_\theta: L^2_N(\T^d)\to L^2_N(\T^d)$ of depth $\bigO(\depth(\cU^\epsilon))$ and width $\bigO(N^d\width(\cU^\epsilon))$ with accuracy, 
\begin{equation}\label{eq:nn-to-fno}
    \norm{\cG-\cG_\theta}_{L^2}
\leq C(\epsilon + C^\epsilon_{\mathrm{stab}}BN^{-r+d/p^*}+ C_{\epsilon,r}^{CB}N^{-r}).
\end{equation}
\end{theorem}
\begin{proof}
We give a sketch of the proof, details can be found in \textbf{SM} \ref{app:nn-to-fno}. Given function values of $v$ on a uniform grid with grid size $1/N$, we use trigonometric polynomial interpolation (\textbf{SM} \ref{app:trig-pol}) to reconstruct $v$ and use this together with Assumption \ref{ass:nn} to construct a neural network. The resulting approximation is then projected onto {\color{black} the space $L^2_N$, of trigonometric polynomials of degree at most $N\in\N$}, again through trigonometric polynomial interpolation. 
\end{proof}

A recent result, \cite[Theorem 36]{kovachki2021universal} ({\bf SM} Lemma \ref{lem:equivalence}), shows that any error bound for FNOs also implies an error bound for DeepONets, by choosing the trunk nets as neural network approximations of the Fourier basis. We apply this result with $\epsilon\sim \poly(1/N)$ to Theorem \ref{thm:nn-to-fno} to obtain the following generic error bound for DeepONets. 

\begin{corollary}\label{cor:nn-to-deeponet}
Assume the setting of Theorem \ref{thm:nn-to-fno}. Then for every $\epsilon>0$, $N\in\N$ and every corresponding FNO $\cG_\theta$ from Theorem \ref{thm:nn-to-fno} there exists a DeepONet $\cG_\theta^*:\cX\to L^2(D)$ with $\width(\branch) = \bigO(N^d)$, $\depth(\branch) =\bigO(\depth(\cG_\theta))$, $\width(\trunk) =\bigO( N^{d+1})$ and $\depth(\trunk) \leq 3$ that satisfies \eqref{eq:nn-to-fno}. 
\end{corollary}

\subsection{Estimates for physics-informed operator learning} \label{sec:FD}

Using the techniques from previous sections, we now present the very first theoretical result for physics-informed operator learning. We demonstrate that if an error estimate for a DeepONet/FNO and the growth of its derivatives are known (see \textbf{SM} \ref{app:error-deeponet} on how to obtain these), then one can prove an error estimate for the corresponding physics-informed DeepONet/FNO. For simplicity, the following result focuses only on operators mapping to $C^r(D)$ but the generalization to e.g. $C^r([0,T]\times D)$ is immediate by considering $D' := [0,T]\times D$.

\begin{theorem}\label{thm:deeponet-to-pido}
Consider an operator $\cG:\cX\to C^r(D)$, $r\in\N$, that satisfies Assumption \ref{ass:partition} and Assumption \ref{ass:diff-operator} with $\ell\in N$. Let $\lambda^*\in (0,\infty]$, let $\lambda, C(\lambda)>0$ with $\lambda\leq\lambda^*$ and let $\sigma:\N\to\R$ be a function such that for all $p\in \N$ there is a DeepONet/FNO $\cG_\theta$ such that
\begin{equation}\label{eq:growth}
    \norm{\cG(v)-\cG_\theta(v)}_{L^2(D)}\leq Cp^{-\lambda}\qquad \text{and} \qquad \abs{\cG_\theta(v)}_{C^r(D)}\leq Cp^{\sigma(r)}\quad \forall r\in\N, v\in\cX. 
\end{equation}
Then for all $\beta\in\R$ with $0<\beta\leq \frac{(r-\ell)\lambda^*-\ell\sigma(r)}{r}$ there exists a constant $C^*>0$ such that for all $v\in \cX$ and $p\in\N$
it holds that
\begin{equation}\label{eq:deeponet-to-pido}
    \norm{\cL(\cG_\theta(v))}_{L^2(D)} \leq C^*p^{-\beta}.
\end{equation}
\end{theorem}
\begin{proof}
For suitable $D^\bmalpha$, use {\bf SM} Lemma \ref{lem:derivatives} with $q=2$, $f_1=\cG(v)$ and $f_2=\cG_\theta(v)$ together with \eqref{eq:growth} to find
\begin{equation}
    \norm{D^\bmalpha(\cG(v)-\cG_\theta(v))}_{L^2(D)} \leq C(r,\lambda)(p^{-\lambda}h^{-\ell} + p^{\sigma(r)}h^{r-\ell}). 
\end{equation}
Let $\beta\in\R$ with $0<\beta\leq \frac{(r-\ell)\lambda^*-\ell\sigma(r)}{r}$. We carefully balance terms by setting $h=p^{-\frac{\sigma(r)+\beta}{r-\ell}}$ and $\lambda = \frac{\ell}{r-\ell}\sigma(r)+\frac{r}{r-\ell}\beta$ to find \eqref{eq:deeponet-to-pido}. Conclude using Assumption \ref{ass:diff-operator}.
\end{proof}

Finally, we use Theorem \ref{thm:nn-to-pinn} to present an alternative error estimate for a physics-informed DeepONet in the case that Assumption \ref{ass:nn} is satisfied. As this assumption is different from assuming access to an error bound for the corresponding DeepONet, it is interesting to use the techniques from the previous sections rather than directly apply Theorem \ref{thm:deeponet-to-pido}. The proof of the following theorem can be found in \textbf{SM} \ref{app:nn-to-pido}. 

\begin{theorem}\label{thm:nn-to-pido}
Let $s,r\in\N$, $T>0$, let $\cG: C^r(\T^d)\to C^{(s,r)}([0,T]\times \T^d)$ be an operator that maps a function $u_0$ to the solution $u$ of the PDE \eqref{eq:PDE} with initial condition $u_0$, let Assumption \ref{ass:nn} and Assumption \ref{ass:stability} be satisfied {\color{black}and let $p^*\in\{2,\infty\}\setminus\{p\}$}. There exists a constant $C>0$ such that for every $Z,N,M\in\N$, $\epsilon, \rho>0$ there is an DeepONet $\cG_\theta: C^r(\T^d)\to L^2([0,T]\times\T^d)$ with $Z^d$ sensors with accuracy, 
\begin{equation}
    \norm{\cG(v)-\cG_\theta(v)}_{L^2([0,T]\times\T^d)} \leq CM^{\rho}(\norm{u}_{C^{(s,0)}}M^{-s}+M^{s-1}(\epsilon+C^\epsilon_{\mathrm{stab}}Z^{-r+d/p^*}+C^{CB}_{\epsilon,r}N^{-r}))
\end{equation}
and if additionally Assumption \ref{ass:partition} and Assumption \ref{ass:diff-operator} hold then,
\begin{equation}
    \norm{\cL(\cG_\theta(v))}_{L^2([0,T]\times\T^d)} \leq CM^{k+\rho}(\norm{u}_{C^{(s,\ell)}}M^{-s}+M^{s-1}N^\ell(\epsilon+C^\epsilon_{\mathrm{stab}}Z^{-r+d/p^*}+C^{CB}_{\epsilon,r}N^{-r})), 
\end{equation}
for all $v$. Moreover, it holds that, $\depth(\branch)=\depth(\cU^\epsilon)$, $\width(\branch) = \bigO(M(Z^d+N^d\width(\cU^\epsilon)))$, 
$\depth(\trunk) = 3$ and $\width(\trunk) = \bigO(MN^d(N+\ln(N)))$. 
\end{theorem}

\subsection{A posteriori bound on the generalization error}\label{sec:generalization}

Although the main focus of this paper is on the approximation error for different neural network architectures, we now demonstrate that it is possible to provide similar bounds for other sources of error, such as the generalization error. We therefore prove a general \emph{a posteriori} upper bound on the generalization error of the all the considered neural network architectures. Consider $f:\cD\to\R$ (an operator or function) and the neural network architecture $f_\theta:\cD\to\R$, $\theta\in\Theta$, which includes all architectures of Section \ref{sec:pde-nn}: neural networks ($\cD = \Omega$, $f=u$ and $f_\theta = u_\theta$), PINNs ($\cD = \Omega$, $f=0$ and $f_\theta = \cL(u_\theta)$), operator learning ($\cD = \Omega\times \cX$, $f=\cG$ and $f_\theta = \cG_\theta$) and physics-informed operator learning ($\cD = \Omega\times \cX$, $f=0$ and $f_\theta = \cL(\cG_\theta)$). 
Given a training set $\S=\{X_1, \ldots X_n\}$, where $\{X_i\}_{i=1}^n$ are iid random variables on $\cD$ (according to a measure $\mu$), the training error $\Et$ and generalization error $\Eg$ are, 
\begin{equation}
    \cE_T(\theta, \S)^2 = \frac{1}{n}\sum_{i=1}^n \abs{f(z_i)-f_\theta(z_i)}^2, \qquad \cE_G(\theta)^2 = \int_\cD\abs{f_\theta(z)-f(z)}^2d\mu(z), 
\end{equation}
where $\mu$ is a probability measure on $\cD$. The following theorem provides a computable a posteriori error bound on the expectation of the generalization error for a general class of approximators. We refer to e.g. \cite{beck2020error, deryck2021pinn} for bounds on $d_\Theta$, $c$ and $\mathfrak{L}$.

\begin{theorem}\label{thm:generalization}
For $R>0$ and $d_\Theta\in\N$, let $\Theta = [-R,R]^{d_\Theta}$ be the set of trainable parameters, and for every training set $\S$, let $\theta^*(\S) \in\Theta$ be an (approximate) minimizer of $\theta\mapsto\Et(\theta,\S)^2$, assume that $\theta\mapsto\Eg(\theta,\S)^2$ and $\theta\mapsto\Et(\theta)^2$ are bounded by $c>0$ and Lipschitz continuous with Lipschitz constant $\mathfrak{L}>0$.  
If $n\geq 2c^2e^8/(2R\mathfrak{L})^{{d_\Theta}/2}$ then it holds that 
\begin{equation}\label{eq:generalization}
    \E{\cE_G(\theta^*(\S))^2} \leq \E{\cE_T(\theta^*(\S),\S)^2} + \sqrt{\frac{2c^2(d_\Theta+1)}{n}\ln(R\mathfrak{L}\sqrt{n})}.
\end{equation}
\end{theorem}

\begin{proof}
The proof (\textbf{SM} \ref{app:generalization}) combines standard techniques, based on covering numbers and Hoeffding's inequality, with an error composition from \cite{deryck2021pinn}. 
\end{proof}

For any type of neural network architecture of depth $L$, width $W$ and weights bounded by $R$, one finds that $d_\Theta \sim LW(W+d)$. For tanh neural networks and operator learning architectures, one has that $\ln(\mathfrak{L}) \sim L\ln(dRW)$, whereas for physics-informed neural networks and DeepONets one finds that $\ln(\mathfrak{L}) \sim (k+\ell)L\ln(dRW)$ with $k$ and $\ell$ as in Assumption \ref{ass:diff-operator} \cite{LMK1, deryck2021pinn}. Taking this into account, one also finds that the imposed lower bound on $n$ is not very restrictive. Moreover, the RHS of \eqref{eq:generalization} depends at most polynomially on $L, W, R, d, k, \ell$ and $c$. For physics-informed architectures, however, upper bounds on $c$ often depend exponentially on $L$ \cite{deryck2021pinn, deryck2021navierstokes}. 

\begin{remark}
As Theorem \ref{thm:generalization} is an a posteriori error estimate, one can use the network sizes of the trained networks for $L$, $W$ and $R$. The sizes stemming from the approximation error estimates of the previous sections can be disregarded for this result. Moreover, instead of considering the expected values of $\Eg$ and $\Et$ in \eqref{eq:generalization}, one can also prove that such an inequality holds with a certain probability (see \textbf{SM} \ref{app:generalization}). 
\end{remark}

\section{Applications}\label{sec:applications}

We demonstrate the power and generality of the framework proposed in Section \ref{sec:general} by applying the presented theory to the following case studies. First, we demonstrate how these generic bounds can be used to overcome the curse of dimensionality (CoD) for linear Kolmogorov PDEs and nonlinear parabolic PDEs (Section \ref{sec:CoD}). These are the first available results that overcome the CoD for nonlinear parabolic PDEs for PINNs and (physics-informed) operator learning. Next, we apply the results of Section \ref{sec:FD} to both linear and nonlinear operators and provide bounds on the approximation error for physics-informed operator learning. 

\subsection{Overcoming the curse of dimensionality}\label{sec:CoD}

For high-dimensional PDEs, it is not possible to obtain efficient approximation results using standard neural network approximation theory \cite{yarotsky2017error, deryck2021approximation} as they will lead to convergence rates that suffer from the CoD, meaning that the neural network size scales exponentially in the input dimension. In literature, one has shown for some PDEs that their solution at a fixed time can be approximated to accuracy $\epsilon>0$ with a network that has size $\bigO(\poly(d)\epsilon^{-\beta})$, with $\beta>0$ independent of $d$, and therefore \emph{overcomes the CoD}. 

\paragraph{Linear Kolmogorov PDEs}
We consider linear time-dependent PDEs of the following form. 

\begin{setting}\label{set:kolmogorov}
Let $s,r\in\N$, $u_0\in C^2_0(\R^d)$ and let $u\in C^{(s,r)}([0,T]\times \R^d)$ be the solution of
\begin{equation}\label{eq:kolmogorov-pde}
    {\color{black}\cL (u)(x,t) = \partial_t u(x,t) - \frac{1}{2}\mathrm{Tr}(\sigma(x)\sigma(x)^T \Delta_x[u](x,t)) - \mu(x)^T \nabla_x[u](x,t) = 0, \quad u(0,x)=u_0(x)}
\end{equation}
for all $(x,t)\in D \times [0,T]$, where $\sigma:\mathbb{R}^d\to \mathbb{R}^{d\times d}$ and $\mu:\mathbb{R}^d\to \mathbb{R}^d$ are affine functions and for which $\norm{u}_{C^{(s,2)}}$ grows at most polynomially in $d$. For every $\epsilon>0$, there is a neural network $\uhat_0$ of width $\bigO(\poly(d)\epsilon^{-\beta})$ such that $\norm{u_0-\uhat_0}_{L^\infty(\R^d)}<\epsilon$. 
\end{setting}

Prototypical examples of such \emph{linear Kolmogorov PDEs} include the heat equation and the Black-Scholes equation. In \cite{grohs2018proof, berner2020analysis, jentzen2018proof} the authors construct a neural network that approximates $u(T)$ and overcomes the CoD by emulating Monte-Carlo methods based on the Feynman-Kac formula. In \cite{deryck2021pinn} one has proven that PINNs overcome the CoD as well, in the sense that the network size grows as $\bigO(\poly(d\rho_d)\epsilon^{-\beta})$, with $\rho_d$ as defined in {\bf SM} \eqref{eq:rhod}. For a subclass of Kolmogorov PDEs it is known that $\rho_d=\poly(d)$, such that the CoD is fully overcome. 

We demonstrate that the generic bounds of Section \ref{sec:general} (Theorem \ref{thm:nn-to-pinn}) can be used to provide a much shorter proof for this result. {\bf SM} Lemma \ref{lem:kolmogorov} verifies that Assumption \ref{ass:nn} is indeed satisfied. The full proof can be found in \textbf{SM} \ref{app:kolmogorov}. 

\begin{theorem}\label{thm:kolmogorov}
Assume that Setting \ref{set:kolmogorov} holds. For every $\sigma,\epsilon>0$ and $d\in\N$, there is a tanh neural network $u_\theta$ of depth $\bigO(\mathrm{depth}(\uhat_0))$ and width $\bigO(\poly(d\rho_d)\epsilon^{-(2+\beta)\frac{r+\sigma}{r-2}\frac{s+1}{s-1}-\frac{1+\sigma}{s-1}})$ such that,
\begin{equation}
    \norm{\cL(u_\theta)}_{L^2([0,T]\times [0,1]^d)} + \norm{u_\theta-u}_{L^2(\partial([0,T]\times [0,1]^d))} \leq \epsilon.
\end{equation}
\end{theorem}

\paragraph{Nonlinear parabolic PDEs}

Next, we consider nonlinear parabolic PDEs as in Section \ref{set:AC}, which  typically arise in the context of nonlinear diffusion-reaction equations that describe the change in space and time of some quantities, such as in the well-known \emph{Allen-Cahn equation} \cite{allen1979microscopic}. 
\begin{setting}\label{set:AC}
Let $s,r\in\N$ and for $u_0\in \cX\subset C^r(\T^d)$ let $u\in  C^{(s,r)}([0,T]\times \T^d)$ be the solution of
\begin{equation}\label{eq:AC}
   {\color{black}\cL (u)(x,t) =  \partial_t u(t,x) - \Delta_x u(t,x) - F(u(t,x))=0, }\qquad u(0,x)=u_0(x), 
\end{equation}
for all $(t,x)\in [0,T] \times D$, with period boundary conditions, where $F:\R\to\R$ is a polynomial and for which $\norm{u}_{C^{(s,2)}}$ grows at most polynomially in $d$. For every $\epsilon>0$, there is a neural network $\uhat_0$ of width $\bigO(\poly(d)\epsilon^{-\beta})$ such that $\norm{u_0-\uhat_0}_{L^\infty(\T^d)}<\epsilon$. Let $\mu$, resp. $\mu^*$, be the normalized Lebesgue measure on $[0,T]\times \T^d$, resp. $\partial([0,T]\times \T^d)$.  
\end{setting}
In \cite{hutzenthaler2020proof} the authors have proven that ReLU neural networks overcome the CoD in the approximation of $u(T)$. We have reproven this result in {\bf SM} Lemma \ref{lem:ass-check-ac} for tanh neural networks to show that Assumption \ref{ass:nn} is satisfied. Using Theorem \ref{thm:nn-to-pinn} we can now prove that PINNs overcome the CoD for nonlinear parabolic PDEs. The proof is analogous to that of Theorem \ref{thm:kolmogorov}.
\begin{theorem}\label{thm:AC-pinn}
Assume Setting \ref{set:AC}. For every $\sigma, \epsilon>0$ and $d\in\N$ there is a tanh neural network $u_\theta$ of depth $\bigO(\mathrm{depth}(\uhat_0)+\poly(d)\ln(1/\epsilon))$ and width $\bigO(\poly(d)\epsilon^{-(2+\beta)\frac{r+\sigma}{r-2}\frac{s+1}{s-1}-\frac{1+\sigma}{s-1}})$ such that,
\begin{equation}
     \norm{\cL(u_\theta)}_{L^2([0,T]\times\T^d, \mu)} + \norm{u-u_\theta}_{L^2(\partial([0,T]\times\T^d, \mu^*))} \leq  \epsilon. 
\end{equation}
\end{theorem}

Similarly, one can use the results from Section \ref{sec:fourier} to obtain estimates for (physics-informed) DeepONets for nonlinear parabolic PDEs \eqref{eq:AC} such as the Allen-Cahn equation. In particular, a dimension-independent convergence rate can be obtained if the solution is smooth enough, which improves upon the result of \cite{LMK1}, which incurred the CoD. For simplicity, we present results for $C^{(2,r)}$ functions, rather than $C^{(s,r)}$ functions, as we found that assuming more regularity did not necessarily further improve the convergence rate. The proof is given in \textbf{SM} \ref{app:nn-to-pido}. 


\begin{theorem}\label{thm:AC-operator}
Assume Setting \ref{set:AC} and let $\cG: \cX\to C^{r}(\T^d):u_0\mapsto u(T)$ and $\cG^*: \cX \to C^{(2,r)}([0,T]\times \T^d):u_0\mapsto u$. For every $\sigma, \epsilon>0$, there exists a DeepONets $\cG_\theta$ and $\cG_\theta^*$ such that
\begin{equation}
    \norm{\cG-\cG_\theta}_{L^2(\T^d\times \cX)} \leq \epsilon, \qquad \norm{\cL(\cG^*_\theta)}_{L^2([0,T]\times\T^d\times \cX)} \leq \epsilon. 
\end{equation}
Moreover, for $\cG_\theta$ we have $\bigO(\epsilon^{-\frac{d+\sigma}{r}})$ sensors and,
\begin{equation}
    \begin{array}{ll}
    \width(\branch)= \bigO( \epsilon^{-\frac{(d+\sigma)(2+\beta)}{r}}), & \depth(\branch) =\bigO( \ln(1/\epsilon)), \\ 
    \width(\trunk) = \bigO(\epsilon^{-\frac{d+1+\sigma}{r}}), & \depth(\trunk) = 3,
    \end{array}
\end{equation}
whereas for $\cG^*_\theta$ we have $\bigO(\epsilon^{-\frac{(3+\sigma)d}{r-2}})$ sensors and,
\begin{equation}
    \begin{array}{ll}
    \width(\branch)= \bigO( \epsilon^{-1-\frac{(3+\sigma)(d+r(2+\beta))}{r-2}}), & \depth(\branch) =\bigO( \ln(1/\epsilon)), \\ 
    \width(\trunk) = \bigO(\epsilon^{-1-\frac{(3+\sigma)(d+1)}{r-2}}), & \depth(\trunk) = 3.
    \end{array}
\end{equation}

\end{theorem}





\subsection{Error bounds for physics-informed operator learning}\label{sec:PIDO}

We demonstrate how Theorem \ref{thm:deeponet-to-pido} can be used to generalize available error estimates for DeepONets and FNOs, e.g. \cite{LMK1, kovachki2021universal} and \textbf{SM} \ref{app:error-deeponet}, to estimates for their physics-informed counterparts. 

\paragraph{Linear operators} In the simplest case, the operator $\cG$ of interest is linear. In \cite[Theorem D.2]{LMK1}, a general error bound for ReLU DeepONets for linear operators has been established, which still holds for tanh DeepONets. Using Theorem \ref{thm:deeponet-to-pido} it is then straightforward to prove convergence rates for physics-informed DeepONets for 
solution operators of linear PDEs \eqref{eq:PDE}. 

Consider an operator $\cG:\cX\to L^2(\T^d):v\mapsto u$ as in Section \ref{sec:setting}, where $v$ is the parameter/initial condition and $u$ the solution of the PDE \eqref{eq:PDE}. Following \cite{LMK1}, we fix the measure $\mu$ on $L^2(\T^d)$ as a Gaussian random field, such that $v$ allows the Karhunen-Loève expansion $v = \sum_{k\in \Z^d} \alpha_{k} X_k \fb_k$,
where $|\alpha_k|\le \exp(-\ell |k|)$ with $\ell > 0$, the $X_k\sim \cN(0,1)$ are iid Gaussian random variables and $\{\fb_k\}_{k\in \Z^d}$ is the standard Fourier basis (\textbf{SM} \ref{app:Fourier}). In this setting, we can prove the following approximation result, the proof of which can be found in \textbf{SM} \ref{app:linear-pido}. {\color{black} The result can be generalized to other data distributions $\mu$ for which a convergence result for DeepONets can be proven, as in \cite{LMK1}.}

\begin{theorem}\label{thm:linear-pido}
Assume the setting above and that of Assumption \ref{ass:diff-operator}, and assume that $\cG(v)\in C^{\ell+1}(\T^d)$ for all $v\in \cX$. For all $\beta>0$ there exists a constant $C>0$ such that for any $p\in\N$ there exists a DeepONet $\cG_\theta$ with $p$ sensors and branch and trunk nets such that
\begin{equation}
    \norm{\cL(\cG_\theta))}_{L^2(L^2(\T^d),\mu)}\leq Cp^{-\beta}.
\end{equation}
Moreover, $\size(\trunk) \leq Cp^{\frac{d+1}{d}}$, $\depth(\trunk) = 3$, $\size(\branch) \leq p$ and $\depth(\branch) = 1$. 
\end{theorem}

\paragraph{Nonlinear operators}

For nonlinear PDEs a general result like Theorem \ref{thm:linear-pido} can not be obtained from the currently available tools. Instead one needs to use Theorem \ref{thm:deeponet-to-pido} for every PDE of interest on a case-by-case basis. In the \textbf{SM}, we demonstrate this for a nonlinear ODE (gravity pendulum with external force, \textbf{SM} \ref{app:pendulum}) and an elliptic PDE (Darcy flow, \textbf{SM} \ref{app:darcy}). 

\section{Related work and discussion}\label{sec:discussion}

This is the first paper to rigorously expose the connections between the different deep learning frameworks from Section \ref{sec:pde-nn} for generic PDEs.
Until now, most available results focus on providing generic results for one specific method. In \cite{hornung2020space} and \cite{grohs2019space} one uses neural networks that approximate solutions to a generic ODE/PDE at a fixed time to construct space-time neural networks. A generalization to PINNs is not immediate as the proof involves the emulation of the forward Euler method. We have overcome this difficulty by constructing space-time neural networks using Taylor expansions instead (Theorem \ref{thm:nn-to-pinn}). To bound the approximation error of PINNs one can use the generic error bounds in Sobolev norms of e.g. \cite{guhring2020error, guhring2021approximation} for very general activation functions or the more concrete bounds \cite{deryck2021approximation} for tanh neural networks. In both approaches, the only assumption is that the solution of the PDE has sufficient Sobolev regularity. As a consequence, these results incur the curse of dimensionality and are not applicable to high-dimensional PDEs. The authors of \cite{deryck2021approximation} analyze PINNs based on three theoretical questions related to approximation, stability and generalization. Other theoretical analyses of PINNs include e.g. \cite{shin2020convergence, Zhang1, hillebrecht2022certified}. 
For DeepONets, convergence rates for advection-diffusion equations are presented in \cite{deng2021convergence} and a clear workflow for obtaining generic error estimates as well as worked out examples can be found in \cite{LMK1}. Similar results are obtained for FNOs in \cite{kovachki2021universal}. A comprehensive comparison of DeepONets and FNOs is the topic of \cite{fair}. To the best of the authors' knowledge, no theoretical results for physics-informed operator learning are currently available. Unrelated to the approximation error, we also report generic bounds on the expected value of the generalization error of all the aforementioned deep learning architectures, in the form of an a posteriori error estimate on the generalization error. 

A second goal of the paper is to prove that deep learning-based frameworks can overcome the curse of dimensionality (CoD). PDEs for which the curse of dimensionality has been overcome include linear Kolmogorov PDEs e.g. \cite{grohs2018proof, jentzen2018proof}, nonlinear parabolic PDEs \cite{hutzenthaler2020proof} and elliptic PDEs {\color{black}\cite{beck2020overcoming_elliptic, chen2021representation, marwah2021parametric}}. By assuming that the initial data lies in a Barron class, the authors of \cite{lu2021priori} proved for elliptic PDEs that the Deep Ritz Method \cite{e2018deep} can overcome the CoD. Since the Barron class is a Banach algebra \cite{chen2021representation} it is possible that our results, which mostly only involve multiplications and additions of neural networks, can be extended to Barron functions. For PINNs, it is proven that they can overcome the CoD for linear Kolmogorov PDEs \cite{deryck2021pinn}. We give an alternative proof of this result, improve the convergence rate (Theorem \ref{thm:kolmogorov}) and additionally prove that PINNs can also overcome the CoD for nonlinear parabolic PDEs (Theorem \ref{thm:AC-pinn}). DeepONets and FNOs can overcome the CoD in many cases \cite{LMK1, kovachki2021universal} but we note that this does not yet include nonlinear parabolic PDEs such as the Allen-Cahn equation. In Theorem \ref{thm:AC-operator} we prove that dimension-independent convergence rates can be obtained if the solution is sufficiently regular. {\color{black} Similar results are expected to hold for e.g. elliptic PDEs by using the results from \cite{beck2020overcoming_elliptic, chen2021representation, marwah2021parametric}.}


It is evident that the generic bounds presented here can only be obtained under suitable assumptions. These should always be checked to prevent misleading claims about mathematical guarantees for the considered deep learning methods. We briefly discuss how restrictive these are and whether they can be relaxed. Assuming the existence of a neural network that approximates the solution of PDE at a fixed time (Assumption \ref{ass:nn}) is of course essential, but such a result can usually be obtained by emulating an existing numerical method. Proving a bound on the Sobolev norm of that network is always possible as we only consider smooth networks. Assumption \ref{ass:partition} holds for many domains, including rectangular and smooth ones. Assumption \ref{ass:diff-operator} and Assumption \ref{ass:stability} also hold for a very broad class of PDEs, much like the assumption on the size of the neural network approximation in Setting \ref{set:kolmogorov} and \ref{set:AC} holds for most functions of interest. Therefore, the assumption that the PDE solution is $C^{(s,r)}$-regular seems to be the most restrictive. However, results like Theorem \ref{thm:nn-to-pinn} could be extended to e.g. Sobolev regular functions by using the Bramble-Hilbert lemma instead of Taylor expansions. Another restriction is that we exclusively focused on neural networks with the tanh activation function. This was only for simplicity of exposition. All results still hold for other sigmoidal activation functions, as well as more general smooth activation functions, which might give rise to slightly different convergence rates. A last restriction is that the obtained rates are not optimal, but this is not the goal of our framework. In particular, for PINNs for low-dimensional PDEs it is beneficial to use e.g. \cite{guhring2021approximation,deryck2021approximation}. 

Optimizing the obtained convergence rates and comparing with optimal ones is one direction for future research. Previously mentioned possibilities include extending to more general activation functions and less regular functions. 
Another direction is to make the connection between our results and that of \cite{chen2021representation} where they prove that Barron spaces are Banach algebras and use this to obtain dimension-independent convergence rates for PDEs with initial data in a Barron class by emulating numerical methods. 

Here, we have considered the approximation and generalization errors in the present analysis. It is clear that the bounds on the generalization error may not be sharp, as in traditional deep learning. Obtaining sharper bounds will be an interesting topic for further investigation. Finally, there is no explicit bound on the training (optimization) errors. Obtaining such bounds will be considered in the future.


\begin{ack}
The research of SM was performed under a project that has received funding from the European Research Council (ERC) under the European Union’s Horizon 2020 research and innovation programme (Grant Agreement No. 770880). 
\end{ack}

\bibliographystyle{abbrv}
{
\small
\bibliography{ref}
}
\section*{Checklist}

\begin{enumerate}

\item For all authors...
\begin{enumerate}
  \item Do the main claims made in the abstract and introduction accurately reflect the paper's contributions and scope?
    \answerYes{See Section \ref{sec:general} and Section \ref{sec:applications}.}
  \item Did you describe the limitations of your work?
    \answerYes{See Section \ref{sec:discussion}.}
  \item Did you discuss any potential negative societal impacts of your work?
    \answerYes{See Section \ref{sec:discussion}.}
  \item Have you read the ethics review guidelines and ensured that your paper conforms to them?
    \answerYes{}
\end{enumerate}

\item If you are including theoretical results...
\begin{enumerate}
  \item Did you state the full set of assumptions of all theoretical results?
    \answerYes{All assumptions are either stated in the theorem statement or described in the text right above the theorem statement.}
        \item Did you include complete proofs of all theoretical results?
    \answerYes{For each result we mention where the proof can be found. }
\end{enumerate}

\item If you ran experiments...
\begin{enumerate}
  \item Did you include the code, data, and instructions needed to reproduce the main experimental results (either in the supplemental material or as a URL)?
    \answerNA{}
  \item Did you specify all the training details (e.g., data splits, hyperparameters, how they were chosen)?
    \answerNA{}
        \item Did you report error bars (e.g., with respect to the random seed after running experiments multiple times)?
    \answerNA{}
        \item Did you include the total amount of compute and the type of resources used (e.g., type of GPUs, internal cluster, or cloud provider)?
    \answerNA{}
\end{enumerate}

\item If you are using existing assets (e.g., code, data, models) or curating/releasing new assets...
\begin{enumerate}
  \item If your work uses existing assets, did you cite the creators?
    \answerNA{}
  \item Did you mention the license of the assets?
    \answerNA{}
  \item Did you include any new assets either in the supplemental material or as a URL?
    \answerNA{}
  \item Did you discuss whether and how consent was obtained from people whose data you're using/curating?
    \answerNA{}
  \item Did you discuss whether the data you are using/curating contains personally identifiable information or offensive content?
    \answerNA{}
\end{enumerate}

\item If you used crowdsourcing or conducted research with human subjects...
\begin{enumerate}
  \item Did you include the full text of instructions given to participants and screenshots, if applicable?
    \answerNA{}
  \item Did you describe any potential participant risks, with links to Institutional Review Board (IRB) approvals, if applicable?
    \answerNA{}
  \item Did you include the estimated hourly wage paid to participants and the total amount spent on participant compensation?
    \answerNA{}
\end{enumerate}

\end{enumerate}


\appendix

\newpage
\section{Notation and preliminaries}\label{app:preliminaries}

We introduce notation and preliminary results regarding finite differences, Sobolev spaces, the Legendre basis, the Fourier basis, trigonometric polynomial interpolation and neural network approximation theory.  

\subsection{Overview of used notation}\label{app:glossary}

\begin{table}[h!]
  \caption{Glossary of used notation.}
  \label{tab:glossary}
  \centering
\begin{tabular}{l p{0.7\textwidth}  r}
\toprule
\textbf{Symbol} & \textbf{Description} & \textbf{Page}
\\
\midrule
$\sigma$ & tanh activation function & \\
$d$ & spatial dimension of domain &  \\
$\T^d$ & periodic torus, identified with $[0,2\pi)^d$ & \\
$D$ & general $d$-dimensional spatial domain & p.~\pageref{sec:setting} \\
$\Omega$ & general domain, either $\Omega=D$ or $\Omega=[0,T]\times D$ & p.~\pageref{sec:setting} \\
$\partial \Omega$ & boundary of $\Omega$ & \\
{\color{black} $\mathcal{H}$} & function space of PDE solution & p.~\pageref{sec:setting} \\
$\cY$ &  function space of parameters for $\cL$, e.g. $\cL_a$ with $a\in\cY$ & p.~\pageref{sec:setting} \\
$\cZ$ &  function space of initial conditions for the PDE \eqref{eq:PDE} & p.~\pageref{sec:setting} \\
$\cX$ &  input function space of the operator $\cG$ & p.~\pageref{sec:setting} \\
$\cG$ &  operator of interest, $\cG:\cX\to L^2(\Omega)$ & p.~\pageref{sec:setting} \\
$\cL$, $\cL_a$ &  differential operator that describes the PDE (with parameter $a$) & p.~\pageref{sec:setting} \\
$r,s$ &  regularity of the PDE solution, $u\in C^r(D)$ or $u\in C^{(s,r)}([0,T]\times D)$ & p.~\pageref{sec:general} \\
$D^{(k,\bmalpha)}$& $D^{(k,\bmalpha)} := D^k_tD^\bmalpha_x := \partial_t^k \partial^{\alpha_1}_{x_1}\ldots \partial^{\alpha_d}_{x_d}$, for $(k,\bmalpha)\in\N_0^{d+1}$ & p.~\pageref{ass:diff-operator} \\
$\ell$ & upper bound on $\norm{\bmalpha}_1$ &  p.~\pageref{ass:diff-operator} \\
$q$ & see Assumption \ref{ass:nn} & p.~\pageref{ass:nn} \\ 
$p$ & see Assumption \ref{ass:stability} & p.~\pageref{ass:stability} \\ 
$\poly(d)$ & a polynomial in $d$ & p.~\pageref{ass:stability} \\ 
$C^r_0$ & subset of $C^r$ functions with compact support & \\
$\Delta^{\bmalpha,r}_h$ & finite difference operator; if the variable is time: $\Delta^{\bmalpha,s}_{h,t}$ & p.~\pageref{app:FD} \\
$\cJ_N$ & grid point indices, $\cJ_N = \{0,\dots, 2N\}^d$ & p.~\pageref{app:sobolev} \\
$\cK_N$ & Fourier wavenumbers $\cK_N = \{k\in \Z^d\:\vert\: |k|_\infty \le N\}$ & \\
$L^2$ & Space of square-integrable functions &  \\
$H^s$ & Sobolev space of smoothness $s$, with norm $\Vert \slot \Vert_{H^s}$ & p.~\pageref{app:sobolev} \\
$L^2_N$ & $L^2_N\subset L^2$ trigonometric polynomials of degree $\le N$ & p.~\pageref{app:trig-pol} \\
\bottomrule
\end{tabular}

\end{table}

\subsection{Finite differences}\label{app:FD}

For $h>0$, $\bmalpha\in\N^d_0$, $r\in\N$ and $\ell := \norm{\bmalpha}_1$, we define a finite difference operator $\Delta^{\bmalpha,r}_h$ as, 
\begin{equation}
    \Delta^{\bmalpha,r}_{h}[f](t,x) = \sum_{j}
    c_{j}^{\bmalpha,r}f(t,x+h b_j^{\bmalpha,r}),
\end{equation}
for $f\in C^{r+\ell}(\R^d)$, where the number of non-zero terms in the summation can be chosen to be finite and only dependent on $\ell$ and $r$ and where the choice of $b_j^{\bmalpha,r}\in\R^d$ allows to approximate $D^\bmalpha_x f$ up to accuracy $\bigO(h^r)$. This means that for any $f\in C^{r+\ell}(\R^d)$ it holds for all $x$ that, 
\begin{equation}\label{eq:accuracy-FD}
 \abs{h^{-\ell}\cdot \Delta^{\bmalpha,r}_{h}[f](t,x) - D^\bmalpha_x f(t,x)} \leq c_{\ell,r} \abs{f(t,\cdot)}_{C^{r+\ell}}h^r \quad \text{for } h> 0, 
\end{equation}
where $c_{\ell,r}>0$ does not depend on $f$ and $h$. Similarly, we can define a finite difference operator $\Delta^{k,s}_{h,t}[f](t,x)$ to approximate $D^k_tf(t,x)$ to accuracy $\bigO(h^s)$.

\subsection{Sobolev spaces}\label{app:sobolev}

Let $d\in\mathbb{N}$, $k\in\mathbb{N}_0$, $1\leq p\leq \infty$ and let $\Omega \subseteq \mathbb{R}^d$ be open. For a function $f:\Omega\to\mathbb{R}$ and a (multi-)index $\bmalpha \in \N^d_0$ we denote by 
\begin{equation}
    D^\bmalpha f= \frac{\partial^{\abs{\bmalpha}} f}{\partial x_1^{\alpha_1}\cdots \partial x_d^{\alpha_d}}
\end{equation}
the classical or distributional (i.e. weak) derivative of $f$. 
We denote by $L^p(\Omega)$ the usual Lebesgue space and for we define the Sobolev space $W^{k,p}(\Omega)$ as
\begin{equation}
    W^{k,p}(\Omega) = \{f \in L^p(\Omega): D^\bmalpha f \in L^p(\Omega) \text{ for all } \bmalpha\in\mathbb{N}^d_0 \text{ with } \abs{\bmalpha}\leq k\}. 
\end{equation}
For $p<\infty$, we define the following seminorms on $W^{k,p}(\Omega)$, 
\begin{equation}
    \abs{f}_{W^{m,p}(\Omega)} = \left(\sum_{\abs{\bmalpha}= m}\norm{D^\bmalpha f}^p_{L^p(\Omega)}\right)^{1/p} \qquad \text{for } m=0,\ldots, k, 
\end{equation}
and for $p=\infty$ we define
\begin{equation}
    \abs{f}_{W^{m,\infty}(\Omega)} =\max_{\abs{\bmalpha}= m} \norm{D^\bmalpha f}_{L^\infty(\Omega)}\qquad \qquad \text{for } m=0,\ldots, k. 
\end{equation}
Based on these seminorms, we can define the following norm for $p<\infty$,
\begin{equation}
    \norm{f}_{W^{k,p}(\Omega)} = \left(\sum_{m=0}^k \abs{f}_{W^{m,p}(\Omega)}^p\right)^{1/p}, 
\end{equation}
and for $p=\infty$ we define the norm
\begin{equation}
    \norm{f}_{W^{k,\infty}(\Omega)} =\max_{0\leq m\leq k}  \abs{f}_{W^{m,\infty}(\Omega)}. 
\end{equation}
The space $W^{k,p}(\Omega)$ equipped with the norm $\norm{\cdot}_{W^{k,p}(\Omega)}$ is a Banach space. 

We denote by $C^k(\Omega)$ the space of functions that are $k$ times continuously differentiable and equip this space with the norm $\norm{f}_{C^k(\Omega)} = \norm{f}_{W^{k,\infty}(\Omega)}$.

\begin{lemma}[Continuous Sobolev embedding]\label{lem:sobolev-embedding}
Let $d,\ell\in\N$ and let $k\geq d/2+\ell$. Then there exists a constant $C>0$ such that for any $f\in H^{k}(\T^d)$ it holds that
\begin{equation}
    \norm{f}_{C^\ell(\T^d)}\leq C \norm{f}_{H^k(\T^d)}.
\end{equation}
\end{lemma}

\subsection{Notation for Legendre basis}\label{app:Legendre}
In a one-dimensional setting, we denote for $j\in \N_0$ the $j$-th Legendre polynomial by $L_j$. Following the notation of \cite{opschoor2019exponential}, it holds that $L_j(x) = \sum_{j=0}^\ell c^j_\ell x^\ell$ where, with $m(\ell):=(j-\ell)/2$,
\begin{equation}
    c^j_\ell = \begin{cases}
    0 & j-\ell \{0,\ldots,j\}\cup (2\Z+1),\\
    (-1)^m2^{-j}\binom{j}{m}\binom{j+\ell}{j}\sqrt{2j+1} & j-\ell \{0,\ldots,j\}\cup 2\Z, 
    \end{cases}
\end{equation}
where each polynomial is normalized in $L^2([-1,1], \lambda/2)$, where $\lambda$ is the Lebesgue measure. Similarly, the tensorized Legendre polynomials, 
\begin{equation}
    L_{\bm{\nu}}(x) = \prod_{j=1}^d L_{\nu_j}(x_j)\qquad \text{for all } \bm{\nu}\in \N^d_0, 
\end{equation}
constitute an orthonormal basis of $L^2([-1,1]^d, \lambda/2^d)$. By considering the lexicographic order on $\N_0^d$, of which we denote the enumeration by $\enum:\N\to\N_0^d$, one can defined an ordered basis $(L_j)_{j\in\N}$ by setting $L_j := L_{\enum(j)}$. 

From \cite[eq. (2.19)]{opschoor2019exponential} it also follows that, 
\begin{equation}\label{eq:legendre-cs-bound}
    \forall s\in\N_0, \bm{\nu}\in \N^d_0: \norm{L_{\bm{\nu}}}_{C^s([-1,1]^d)} \leq \prod_{j=1}^d(1+2\nu_j)^{1/2+2s}. 
\end{equation}

\subsection{Notation for Standard Fourier basis}  
\label{app:Fourier}
Using the notation from \cite{LMK1}, we introduce the following ``standard'' real Fourier basis $\{\fb_{\kappa}\}_{\kappa\in \Z^d}$ in $d$ dimensions. For $\kappa = (\kappa_1,\dots, \kappa_d)\in \Z^d$, we let $\sigma(\kappa)$ be the sign of the first non-zero component of $\kappa$ and we define
\begin{align}
\fb_{\kappa}
:=
C_{\kappa}
\begin{cases}
1,
& \sigma(\kappa)=0, \\
\cos(\langle\kappa,{x}\rangle),
& \sigma(\kappa)=1, \\
\sin(\langle\kappa,{x}\rangle),
& \sigma(\kappa)=-1,
\end{cases}
\end{align}
where the factor $C_\kappa > 0$ ensures that $\fb_{\kappa}$ is properly normalized, i.e. that $\Vert \fb_{\kappa} \Vert_{L^2(\T^d)} = 1$. Next, let $\enum: \N \to \Z^d$ be a fixed enumeration of $\Z^d$, with the property that $j \mapsto |\enum(j)|_\infty$ is monotonically increasing, i.e. such that $j\le j'$ implies that $|\enum(j)|_\infty \le |\enum(j')|_\infty$. This will allow us to introduce an $\N$-indexed version of the Fourier basis, 
\begin{equation}
\fb_j({x}) := \fb_{\enum(j)}({x}), \quad\forall j\in \N. 
\end{equation}
Finally we note that
\begin{equation}\label{eq:bound-fourier-basis}
    \norm{\fb_{\kappa}}_{C^s([0,2\pi]^d)} \leq \norm{\kappa}_\infty^s. 
\end{equation}

\subsection{Trigonometric polynomial interpolation}\label{app:trig-pol}

For $N\in\N$, let $x_j = \frac{2\pi j}{2N+1}$ and let $y_j\in\R$ for all $j\in \cJ_N = \{0, \ldots, 2N+1\}^d$. We will construct an operator 
\begin{equation}
    \cQ_N: \R^{\abs{\cJ_N}} \to L^2(\T^d): y \mapsto \cQ_N(y), 
\end{equation}
where $\cQ_N(y)$ is a trigonometric polynomial of degree at most $N$ such that $\cQ_N(y)(x_j) = y_j$ for all $j\in \cJ_N$. We construct this polynomial using the discrete Fourier transform and its inverse. 
For $k\in \cK_N = \{-N, \ldots, N\}^d$, we define the discrete Fourier transform as,
\begin{equation}
    X_k(y) = \sum_{j\in \cJ_N} y_j \exp(-i \langle k, x_j\rangle), 
\end{equation}
and the trigonometric interpolation polynomial as, 
\begin{equation}\label{eq:trig-pol}
\begin{split}
    \cQ_N(y)(z) &= \frac{1}{\abs{\cK_N}}\sum_{k\in \cK_N} X_k(y) \exp(i \langle k, z\rangle) \\
    & = \frac{1}{\abs{\cK_N}} \sum_{k\in \cK_N} \sum_{j\in \cJ_N} y_j \cos( \langle k, z-x_j\rangle)\\
    &= \frac{1}{\abs{\cK_N}} \sum_{k\in \cK_N} \sum_{j\in \cJ_N} y_j \left(\cos( \langle k, x_j\rangle)\cos( \langle k, z\rangle)-\sin( \langle k, x_j\rangle)\sin( \langle k, z\rangle)\right)\\
     &= \frac{1}{\abs{\cK_N}} \sum_{k\in \cK_N} \sum_{j\in \cJ_N} y_j a_{k,j} \fb_k(z), 
\end{split}
\end{equation}
where,
\begin{align}\label{eq:trig-pol-coeff}
a_{k,j}
=
\begin{cases}
1,
& \sigma(k)=0, \\
\cos( \langle k, x_j\rangle),
& \sigma(k)=1, \\
\sin( \langle k, x_j\rangle),
& \sigma(k)=-1,
\end{cases}
\end{align}
with $\sigma$ as in \textbf{SM} \ref{app:Fourier}.
We can also define an encoder $\cE_N$ by, 
\begin{equation}\label{eq:encoder}
    \cE_N: C(\T^d)\to \R^{\abs{\cJ_N}}: f\mapsto (f(x_j))_{j\in\cJ_N}.
\end{equation}
The composition $\cQ_N\circ \cE_N$ is called the pseudo-spectral projection onto the space of trigonometric polynomials of degree at most $N$ and has the following property \cite{kovachki2021universal}. 
\begin{lemma}\label{lem:acc-trig-pol}
For $s, k\in\N_0$ with $s>d/2$ and $s\geq k$, and $f\in C^s(\T^d)$ it holds that 
\begin{equation}
    \norm{f-(\cQ_N\circ\cE_N)(f)}_{H^k(\T^d)}\leq C(s,d)N^{-(s-k)}\norm{f}_{H^s(\T^d)}, 
\end{equation}
for a constant $C(s,d)>0$ that only depends on $s$ and $d$. 
\end{lemma}

\subsection{Neural network approximation theory}\label{app:nnt}

We recall some basic results on the approximation of functions by tanh neural networks in this section. All results are adaptations from results in \cite{deryck2021approximation}. The following two lemmas address the approximation of univariate monomials and the multiplication operator. 

\begin{lemma}[Approximation of univariate monomials, Lemma 3.2 in \cite{deryck2021approximation}]\label{lem:pol-tanh}
Let $k\in\mathbb{N}_0, s\in 2\mathbb{N}-1$,  ${\color{black}M>0}$ and define $f_p:[-M,M]\to\R: x\mapsto x^p$ for all $p\in \N$. For every $\epsilon>0$, there exists a shallow tanh neural network $\psi_{s,{\color{black}\epsilon}} : {\color{black}[-M,M]}\to \mathbb{R}^s$ of width $\frac{3(s+1)}{2}$ such that 
\begin{equation}
    \max_{p\leq s} \ck{f_p- (\psi_{s,\epsilon})_p} \leq\epsilon.
\end{equation}
\end{lemma}

\begin{lemma}[Shallow approximation of multiplication of $d$ numbers, Corollary 3.7 in \cite{deryck2021approximation}]\label{lem:mult-shallow}
Let $d\in \mathbb{N}$, $k\in \mathbb{N}_0$ and $M>0$. Then for every $\epsilon>0$, there exist a shallow tanh neural network $\widehat{\times}_d^\epsilon: [-M,M]^d\to\mathbb{R}$ of width $3\left\lceil\frac{d+1}{2}\right\rceil\abs{P_{d,d} }$ (or $4$ if $d=2$) such that
\begin{equation}
   \ck{\widehat{\times}_d^\epsilon(x)-\prod_{i=1}^d x_i} \leq \epsilon.
\end{equation}
\end{lemma}

\section{Additional material for Section \ref{sec:general}}\label{app:general}

\subsection{Auxiliary results for Section \ref{sec:general}}

\begin{lemma}\label{lem:derivatives}
Let $q\in [1,\infty]$, $r, \ell\in\N$ with $\ell\leq r$ and $f_1, f_2 \in C^{(0,r)}([0,T]\times D)$. If Assumption \ref{ass:partition} holds then there exists a constant $C(r)>0$ such that for any $\bmalpha\in\N^d_0$ with $\ell := \norm{\bmalpha}_1$ it holds that
\begin{equation}
    \norm{D^\bmalpha_x (f_1-f_2)}_{L^q} \leq C(\norm{f_1-f_2}_{L^q}h^{-\ell} + \max_{j=1,2}\abs{f_j}_{C^{(0,r)}}h^{r-\ell})\qquad \forall h>0.
\end{equation}
\end{lemma}
\begin{proof}
From the triangle inequality and \eqref{eq:accuracy-FD} the existence of a constant $C(r)>0$ follows such that,
\begin{equation}
    \begin{split}
        \norm{D^\bmalpha_x (f_1-f_2)}_{L^q} &\leq  \max_{j=1,2} \norm{D^\bmalpha f_j-h^{-\ell}\cdot\Delta^{\bmalpha,r}_h[f_j]}_{L^q} + C(r) h^{-\ell} \norm{f_1-f_2}_{L^q} \\
            &\leq c_{\ell,r} \max_{j=1,2} \abs{f_j}_{C^{(0,r)}} h^{r-\ell} + C(r) h^{-\ell} \norm{f_1-f_2}_{L^q}.\qedhere
    \end{split}
\end{equation}
\end{proof}

\begin{lemma}\label{lem:wlog-faa-di-bruno}
Using the notation of the proof of Theorem \ref{thm:nn-to-pinn} (\textbf{SM} \ref{app:nn-to-pinn}), it holds that 
\begin{equation}
    \norm{D^{(k,\bmalpha)}(\utilde-\uhat)}_{C^0} \leq \delta.
\end{equation}
\end{lemma}
\begin{proof}
Using the Faà di Bruno formula \cite{constantine1996multivariate} and its consequences for estimating the norms of derivatives of compositions \cite[Lemma A.7]{deryck2021approximation} one can prove for sufficiently regular functions $g_1,g_2,h_1,h_2$ and a suitable multi-index $\branch$ estimates of the form,
\begin{equation}
    \norm{D^{\branch}(g_1\circ h_1-g_2\circ h_2)}_{C^0} \leq C(\norm{g_1-g_2}_{C^{\norm{\branch}_1}}+\norm{h_1-h_2}_{C^{\norm{\branch}_1}}), 
\end{equation}
assuming that the compositions are well-defined and where the constant $C>0$ may depend on $g_1,g_2,h_1,h_2$ and their derivatives. Using this theorem we can prove that
\begin{equation}\label{eq:wlog1}
    \norm{D^{(k,\bmalpha)}\uhat-D^{(k,\bmalpha)}\sum_{m=1}^M \sum_{i=0}^{s-1} \frac{\Delta^{i,s-i}_{1/M,t}[\uhat^\epsilon_m](t_m,x)}{M^{-i} i!}\cdot \widehat{\varphi}_i^\delta(t-t_m)\cdot \Phi^M_m(t)} < C\delta. 
\end{equation}
Because the size of the neural network $\hattimes_\delta$ in the definition of $\uhat$ does not depend on its accuracy $\delta$ (see Lemma \ref{lem:mult-shallow}) we can rescale $\delta$ and therefore set $C=1/2$ in the above inequality. 

Next, we observe that,
\begin{align}\label{eq:wlog2}
    \begin{split}
        &D^{(k,\bmalpha)}\sum_{m=1}^M \sum_{i=0}^{s-1} \frac{\Delta^{i,s-i}_{1/M,t}[\uhat^\epsilon_m](t_m,x)}{M^{-i} i!}\cdot (\widehat{\varphi}_i^\delta-\varphi_i)(t-t_m)\cdot \Phi^M_m(t) \\
        &\quad = \sum_{m=1}^M \sum_{i=0}^{s-1} \frac{\Delta^{i,s-i}_{1/M,t}[D^\bmalpha_x \uhat^\epsilon_m](t_m,x)}{M^{-i} i!}\cdot \sum_{n=0}^k \binom{k}{n}\partial_t^n(\widehat{\varphi}_i^\delta-\varphi_i)(t-t_m)\cdot \partial_t^{k-n}\Phi^M_m(t)
    \end{split}
\end{align}
Analogously to before, because the sizes of the neural networks $\varphihat_i^\delta$ are independent of their accuracy $\delta$ we can rescale $\delta$ such that $\norm{\eqref{eq:wlog2}}_{C^0}\leq \delta/2$. The claim follows by the triangle inequality,
\begin{equation}
    \norm{D^{(k,\bmalpha)}(\utilde-\uhat)}_{C^0} \leq\norm{\eqref{eq:wlog1}}_{C^0}+ \norm{\eqref{eq:wlog2}}_{C^0}\leq \delta.
\end{equation}
\end{proof}

\begin{lemma}\label{lem:newton-pol}
Let $\Delta^{k,s}_{h,t}$ be a finite difference operator cf. Section \ref{sec:FD} and \textbf{SM} \ref{app:FD}, let $1\leq j\leq d$, let $1\leq q\leq \infty$, let $\ell \in \N_0$ and let $\bmalpha\in\N^d_0$ with $\norm{\bmalpha}_1=\ell$. Let $u,\uhat\in C^{(s,\ell)}([-2h,2h]\times D)$ such that for all $t\in [-2h,2h]$,
\begin{equation}\label{eq:newton-u-uhat}
    \norm{D^{\bmalpha}_x(u(t,\cdot)-\uhat(t,\cdot))}_{L^q(D)} \leq \epsilon. 
\end{equation}
Then there exists $c_s>0$ holds that, 
\begin{equation}
    \norm{D^{k,\bmalpha}\left(\sum_{i=0}^{s-1}\frac{\Delta^{i,s-i}_{h,t}[\uhat](0,x)}{h^i i!}t^i-u(t,\cdot)\right)}_{L^q}\leq c_{s}\left(\epsilon h^{-k}+\abs{D^{\bmalpha}_xu}_{C^{(s,0)}}h^{s-k}\right).
\end{equation}
\end{lemma}
\begin{proof}
Let $t\in[-2h,2h]$, $\bmalpha\in\N^d_0$ with $\norm{\bmalpha}_1=\ell$ and $x\in\R^d$ be arbitrary. We first observe that,
\begin{equation}
   D^{k,\bmalpha}\sum_{i=0}^{s-1}\frac{\Delta^{i,s-i}_{h,t}[u](0,x)}{h^i i!}t^i = \sum_{i=k}^{s-1}\frac{\Delta^{i,s-i}_{h,t}[D^{\bmalpha}_x u](0,x)}{h^i (i-k)!}t^{i-k}. 
\end{equation}
Taylor's theorem then guarantees the existence of $\xi_{t,x}\in[-2h,2h]$ such that
\begin{equation}
\begin{split}
    &D^{k,\bmalpha}\left(\sum_{i=0}^{s-1}\frac{\Delta^{i,s-i}_{h,t}[u](0,x)}{h^i i!}t^i-u(t,\cdot)\right)\\
    &\quad = \sum_{i=0}^{s-1-k}\left[\frac{\Delta^{i+k,s-i-k}_{h,t}[D^{\bmalpha}_x u](0,x)}{h^{i+k} i!}t^{i}-\frac{D^{i+k,\bmalpha} u(0,x)}{i!}t^i\right] + \frac{D^{s,\bmalpha}u(\xi_{t,x},x)}{(s-k)!}t^{s-k}. 
\end{split}
\end{equation}
Now observe that because of assumption \eqref{eq:newton-u-uhat} and the definition and properties \eqref{eq:accuracy-FD} of the finite difference operator, there exists a constant $C_s>0$ such that,
\begin{equation}
\begin{split}
    &\norm{\Delta^{i+k,s-i-k}_{h,t}[D^{\bmalpha}_x \uhat](0,x)-\Delta^{i+k,s-i-k}_{h,t}[D^{\bmalpha}_x u](0,x)}_{L^q} \leq C_s\epsilon,\\
    &\abs{\frac{\Delta^{i+k,s-i-k}_{h,t}[D^{\bmalpha}_x u](0,x)}{h^{i+k}} - D^{i+k,\bmalpha}u(0,x)} \leq C_s \abs{D^{\bmalpha}_xu}_{C^{(s,0)}}h^{r-i-k}.
\end{split}
\end{equation}
Combining all previous results provides us with the existence of a constant $c_s>0$ such that,
\begin{equation}
    \begin{split}
        &\norm{D^{k,\bmalpha}\left(\sum_{i=0}^{s-1}\frac{\Delta^{i,s-i}_{h,t}[\uhat](0,x)}{h^i i!}t^i-u(t,\cdot)\right)}_{L^q}\\
        &\quad \leq \sum_{i=0}^{s-1-k}\left[\frac{C_s\eps}{h^{i+k} i!}h^i + \frac{C_s }{i!}\abs{D^{\bmalpha}_xu}_{C^{(s,0)}}h^{s-i-k}h^i\right] + \frac{1}{(s-k)!}\abs{D^{\bmalpha}_xu}_{C^{(s,0)}}h^{s-k}\\
        &\quad \leq c_s\left(\epsilon h^{-k}+\abs{D^{\bmalpha}_xu}_{C^{(s,0)}}h^{s-k}\right). 
    \end{split}
\end{equation}
\end{proof}




\begin{definition}\label{def:pou}
Let $C>0$, $N\in\N$, $0<\epsilon<1$ and $\alpha=\ln(CN^k/\epsilon)$. For every $1\leq j\leq N$, we define the function $\Phi_j^N:[0,T]\to[0,1]$ by
\begin{align}\label{eq:pou-def}
\begin{split}
    \Phi_1^N(t) &= \frac{1}{2}-\frac{1}{2}\sigma\left(\alpha\left(t-\frac{T}{N}\right)\right),\\
    \Phi_j^N(t) &= \frac{1}{2}\sigma\left(\alpha\left(t-\frac{T(j-1)}{N}\right)\right) - \frac{1}{2}\sigma\left(\alpha\left(t-\frac{Tj}{N}\right)\right),\\
    \Phi_N^N(t) &= \frac{1}{2}\sigma\left(\alpha\left(t-\frac{T(N-1)}{N}\right)\right)+\frac{1}{2}.
    \end{split}
\end{align}
\end{definition}
The functions $\{\Phi^{N}_j\}_j$ approximate a partition of unity in the sense that for every $j$ it holds on $I_j^N$ that for some $\epsilon>0$,
\begin{equation}
    1-\sum_{v=-1}^1\Phi^{N}_{j+v} \lesssim \epsilon \quad \text{and} \quad \sum_{\substack{\abs{v}\geq 2,\\ j+v \in \{1,\ldots, N\}}}\Phi^{N}_{j+v} \lesssim \epsilon.
\end{equation}
This is made exact in \cite[Section 4]{deryck2021approximation}.

\begin{theorem}\label{thm:tanh-approximation}
Let $k\in\N\cup\{0\}$, $q\in\{2,\infty\}$, $\xi>0$ and $s\in \N$. Let $\mu$ be a probability measure on $D$ and let $f\in C^s([0,T], L^q(\mu))$. 
Assume that for every $0\leq \ell \leq k$ there is a constant $\cC^*_\ell>0$ for which it holds that for every $N\in\N$ there exist functions $\{p_j^N\}_{j=1}^N$ that satisfy for all $ 1\leq j\leq N$,
\begin{align}\label{eq:pkn-acc-sobolev}
\begin{split}
    \abs{f-p_j^N}_{C^\ell(J^N_j, L^q(\mu))}=\max_{t\in \left[\frac{(j-2)T}{N},\frac{(j+1)T}{N}\right]}\norm{D_t^\ell(f(t,\cdot)-p_j^N(t,\cdot))}_{L^q(\mu)} &\leq   \mathcal{C^*_{\ell}}N^{-s+\ell}+\xi.
\end{split}
\end{align}
Let $\mathcal{C}_k:=\max\{\max_{0\leq \ell \leq k}\mathcal{C}^*_\ell, \hkunit{f},1\}$. There exists a constant $C(k)>0$ that only depends on $k$ such that for all $N\geq 3$ it holds that,
\begin{equation}
\begin{split}
    \hkunit{f-\sum_{j=1}^N p_j^N\cdot \Phi^{N}_j} \quad \leq C \ln^k\left(N\right)\left[\frac{\mathcal{C}_k}{N^{s-k}} + \xi N^k\right]. 
\end{split}
\end{equation}
\end{theorem}

\begin{proof}
We follow the proof of \cite[Theorem 5.1]{deryck2021approximation}. All steps of the proofs are identical, with the only difference being that the $W^{k,\infty}([0,1]^d)$-norm of \cite{deryck2021approximation} is replaced by the ${C^k([0,T], L^2(\mu))}$-norm in this work. Following \cite{deryck2021approximation}, one divides the domain $[0,T]$ into intervals $I_i^N = [t_{i-1}, t_i]$, with $t_i=iT/N$ and $N\in\N$ large enough. On each of these intervals, $f$ locally can be  approximated (in Sobolev norm) by $p_j^N$, by virtue of the assumptions of the theorem. A global approximation can then be constructed by multiplying each $p_j^N$ with an approximation of the indicator function of the corresponding intervals and summing over all intervals. 

We now highlight the main steps in the proof. 
\textit{Step 2a} (as in \cite{deryck2021approximation}) results in the following estimate,
\begin{align}
    \begin{split}
        \hki{f-\sum_{j=1}^N f \cdot \Phi^{N}_j} \leq C \hki{f} \left(\epsilon +N^{k+1} \ln^k\left(\frac{CN^k}{\epsilon}\right)\epsilon\right). 
    \end{split}
\end{align}
\textit{Step 2b} results in the estimate, 
\begin{align}
    \begin{split}
       \hki{\sum_{j=1}^N (f -p_j^N)\cdot \Phi^{N,d}_j}\leq  C \ln^k\left(\frac{CN^k}{\epsilon}\right)\left[\frac{\mathcal{C}_k}{N^{s-k}} + \xi N^k + \mathcal{C}_k N^{k+1} {\epsilon}\right],
    \end{split}
\end{align}
Putting everything together, we find that if $CN^k\geq {\epsilon}e$,
\begin{equation}
\begin{split}
    &\hkunit{f-\sum_{j=1}^N p_j^N\cdot \Phi^{N}_j} \\&\quad \leq C \ln^k\left(\frac{CN^k}{\epsilon}\right)\left[(\hki{f}+\cC_k)N^{k+1}\epsilon+ \frac{\mathcal{C}_k}{N^{s-k}} + \xi N^k\right]. 
\end{split}
\end{equation}
In particular, if we set $N^{k+1}\epsilon = N^{-s+k}$ and $N\geq 3$, then we find that
\begin{equation}
\begin{split}
    \hkunit{f-\sum_{j=1}^N p_j^N\cdot \Phi^{N}_j} \quad \leq C \ln^k\left(N\right)\left[\frac{\hki{f}+\mathcal{C}_k}{N^{s-k}} + \xi N^k\right]. 
\end{split}
\end{equation}
\end{proof}

\begin{lemma}\label{lem:equivalence}
Let $\cG_\theta:\cX \to \cH$ be a tanh FNO with grid size $N\in\N$ and let $B>0$. For every $\epsilon>0$, there exists a tanh DeepONet $\cG_\theta^*: \cX\to\cH$ with $N^d$ sensors and $N^d$ branch and trunk nets such that
\begin{equation}
    \sup_{\norm{v}_{L^\infty}\leq B}\sup_{x \in \T^d} \abs{\cG_\theta^*(v)(x)-\cG_\theta(v)(x)}\leq \epsilon. 
\end{equation}
Furthermore, $\width(\branch) \sim N^d$, $\depth(\branch) \sim \ln(N)$, $\width(\trunk) \sim N^{d}(N+\ln(N/\epsilon))$ and $\depth(\trunk) = 3$. 
\end{lemma}
\begin{proof}
This is a consequence of \cite[Theorem 36]{kovachki2021universal} and Lemma \ref{lem:rec-fourier} with $\epsilon\leftarrow N^d\epsilon$. 
\end{proof}

\subsection{Proof of Theorem \ref{thm:nn-to-pinn}}\label{app:nn-to-pinn}

\begin{proof}
\textbf{Step 1: construction. }
To define the approximation, we divide $[0,T]$ into $M$ subintervals of the form $[t_{m-1},t_m]$, where $t_m = mT/M$ with $1\leq m\leq M$. One could approximate $u$ on every subinterval by an $s$-th order accurate Taylor approximation around $t_m$, provided that one has access to $D^i_t u(\cdot, t_m)$ for $0\leq i \leq s-1$. As those values are unknown, we resort to the finite difference approximation $D^i_t u(\cdot, t_m)\approx M^i\cdot \Delta^{i,s-i}_{1/M,t}[\cU^\epsilon(u_0,t_m)]$, which is a neural network. See \textbf{SM} \ref{app:FD} for an overview of the notation for finite difference operators. Moreover, we replace the univariate monomials $\varphi_i:[0,T]\to\R: t\mapsto t^i$ in the Taylor approximation by neural networks $\widehat{\varphi}_i^\delta:[0,T]\to\R$ with $\Vert\varphi_i-\widehat{\varphi}_i^\delta\Vert_{C^{k+1}}\lesssim \delta$. Lemma \ref{lem:pol-tanh} guarantees that the output of $(\widehat{\varphi}_i^\delta)_{i=1}^{s-1}$ can be obtained using a shallow network with width $2(s+1)$ (independent of $\delta$). The multiplication operator is replaced by a shallow neural network $\hattimes_\delta: [-a,a]^2\to\R$ (for suitable $a>0$) for which $\Vert\times-\hattimes_\delta\Vert_{C^{k+1}}\lesssim \delta$. By Lemma \ref{lem:mult-shallow} only four neurons are needed for this network. This results in the following approximation for $f\in C^0([0,T]\times D)$, 
\begin{equation}
     \hatN^{\delta}_m[f](t,x) := \sum_{i=0}^{s-1} \hattimes_\delta\left(\frac{\Delta^{i,s-i}_{1/M,t}[f](t_m,x)}{M^{-i} i!}, \widehat{\varphi}_i^\delta(t-t_m)\right) \quad \forall t\in [0,T], \: x\in D, \: 1\leq m\leq M. 
\end{equation}
Next, we patch together these individual approximations by (approximately) multiplying them with a NN approximation of a partition of unity, denoted by $\Phi^M_1, \ldots, \Phi^M_M:[0,T]\to[0,1]$, as introduced in Definition \ref{def:pou} in \textbf{SM} \ref{app:general}. Every $\Phi_m^M$ can be thought of as a NN approximation of the indicator function on $[t_{m-1},t_m]$. For any $\epsilon,\delta>0$, we then define our final neural network approximation $\uhat:[0,T]\times D\to \R$ as,
\begin{equation}\label{eq:def-pinn}
\begin{split}
     \uhat(t,x) := \sum_{m=1}^M \hattimes_\delta \left(\hatN^{\delta}_m[\cU^\epsilon(u_0,t_m)](t,x), \Phi^M_m(t)\right) \quad \forall t\in [0,T], \: x\in D.
\end{split}
\end{equation}

\textbf{Step 2: error estimate.} In order to facilitate the proof, we introduce the intermediate approximations $\utilde:[0,T]\times D\to \R$ and $N_m:C^0(D)\times [0,T]\times D\to \R$ by, 
\begin{equation}
     \utilde(t,x) := \sum_{m=1}^M N_m[\uhat^\epsilon_m](t,x)\cdot \Phi^M_m(t) := \sum_{m=1}^M \sum_{i=0}^{s-1} \frac{\Delta^{i,s-i}_{1/M,t}[\uhat^\epsilon_m](t_m,x)}{M^{-i} i!}\cdot \varphi_i(t-t_m)\cdot \Phi^M_m(t), 
\end{equation}
where $\uhat^\epsilon_m = \cU^\epsilon(u_0,t_m)$. Note that $\uhat$ can be obtained from $\utilde$ by replacing the multiplication operator and the monomials by neural networks. Since these the size of these networks are independent of their accuracy $\delta$, we can assume without loss of generality that $\Vert D^{(k,\bmalpha)}(\utilde-\uhat)(t,\cdot)\Vert_{L^q}\leq \delta$ (see Lemma \ref{lem:wlog-faa-di-bruno}) for any relevant $D^{(k,\bmalpha)}$ and $t$. 

It remains to prove that $D^{(k,\bmalpha)}\utilde \approx D^{(k,\bmalpha)}u$. 
Combining the observation that $D^{(k,\bmalpha)}N_m[\uhat^\epsilon_m] = D_t^k N_m[D^\bmalpha_x\uhat^\epsilon_m]$ with Lemma \ref{lem:newton-pol} lets us conclude that for all $0\leq k\leq s-1$ and $t\in[t_{m-2}, t_{m+2}]$,
\begin{equation}
    \norm{D^{(k,\bmalpha)}( N_m[\uhat^\epsilon_m](t,\cdot)-u(t,\cdot))}_{L^q} \leq C(r)M^k(\norm{D^\bmalpha_x(\uhat^\epsilon_m-u)(\cdot,t_m)}_{L^q}+\abs{u}_{C^{(s,\ell)}}M^{-s})
\end{equation}
We use Theorem \ref{thm:tanh-approximation} with $f\leftarrow u$, $p_j^N \leftarrow  N_m[D^\bmalpha_x\uhat^\epsilon_m]$, $\xi\leftarrow C(r)M^{k} \norm{D^\bmalpha_x(\uhat^\epsilon_m-u)(\cdot,t_m)}_{L^q}$, $\cC^*_\ell \leftarrow C(s)\abs{u}_{C^{(k,\ell)}}$, $N\leftarrow M$ to find that, 
\begin{equation}
    \norm{D^{(k,\bmalpha)}(\uhat-u)}_{L^q} \leq C\ln^k(M)(\norm{u}_{C^{(s,\ell)}}M^{k-s}+M^{2k}\norm{D^\bmalpha_x(\uhat^\epsilon_m-u)(\cdot,t_m)}_{L^q}), 
\end{equation}
where $C(r,s)>0$ only depend on $r$ and $s$. Finally, using Lemma \ref{lem:derivatives} to bound $\norm{D^\bmalpha_x(\uhat^\epsilon_m-u)(\cdot,t_m)}_{L^q}$ and combining this with  Assumption \ref{ass:nn} proves \eqref{eq:pinn-acc}. 

\textbf{Step 3: size estimate.} The following holds, 
\begin{equation}
    \depth(\uhat)\leq C\depth(\cU^\epsilon), \width(\uhat) \leq CM \width(\cU^\epsilon). 
\end{equation}
\end{proof}

\subsection{Proof of Theorem \ref{thm:nn-to-fno}}\label{app:nn-to-fno}

\begin{proof}
\textbf{Step 1: construction. }
Let $N\in\N$, let $\cE_N: C^0(T^d)\to \R^{\abs{\cJ_N}}$ be an encoder and $\cQ_N: \R^{\abs{\cJ_N}}\to L^2_N$ be a trigonometric polynomial interpolation operator, cf. \textbf{SM} \ref{app:trig-pol}. If we let $\cGhat = \cU^\epsilon \circ \cQ_N \circ \cE_N$ then we can define an FNO $\cG_\theta: L^2_N(\T^d)\to L^2_N(\T^d)$ as $\cG_\theta(u_0)(x) =  (\cQ_N\circ \cE_N \circ \cGhat)(u_0)(x)$. 

\textbf{Step 2: error estimate.}
We decompose the $L^2$-error of the FNO using the triangle inequality and the inequality $\Vert\cU^\epsilon-\cGhat\Vert_{L^2}\leq C^\epsilon_{\mathrm{stab}}\norm{u_0-\cQ_N\circ\cE_N\circ u_0}_{L^p}$, which follows from Assumption \ref{ass:stability},
\begin{align}\label{eq:triangle-deeponet}
\begin{split}
\norm{\cG-\cG_\theta}_{L^2}
&\leq \norm{\cG-\cU^\epsilon}_{L^2} + C^\epsilon_{\mathrm{stab}}\norm{u_0-\cQ_N\circ\cE_N\circ u_0}_{L^p}+ \Vert\cGhat-\cG_\theta\Vert_{L^2}.
\end{split}
\end{align}

First, we find using a Sobolev embedding result (Lemma \ref{lem:sobolev-embedding}) and Lemma \ref{lem:acc-trig-pol} that,
\begin{equation}\label{eq:QEu0}
     \norm{u_0-(\cQ_N\circ\cE_N)(u_0)}_{L^p}\leq  \norm{u_0-(\cQ_N\circ\cE_N)(u_0)}_{H^{d/p^*}} \leq C(d,r) N^{-r+d/p^*}\norm{u_0}_{H^{r}},
\end{equation}
where $p^*$ is such that $1/p+1/p^*=1/2$. Next, we observe that for any $u_0\in\cX$ with $\norm{u_0}_{C^r}\leq B$ that $\norm{(\cQ_N\circ\cE_N)(u_0)}_{H^r(\T^d)} \leq C B=:\Bar{B}$. Hence, by applying Lemma \ref{lem:acc-trig-pol} to the second and last term of \eqref{eq:triangle-deeponet} we find that,
\begin{equation}
    \norm{\cG-\cG_\theta}_{L^2}
\leq C(\epsilon + C^\epsilon_{\mathrm{stab}}BN^{-r+d/p^*}+ C_{\epsilon,r}^{\Bar{B}}N^{-r}).
\end{equation}

\textbf{Step 3: size estimate. }
As for any FNO, the width is equal to $N^d\width(\cU^\epsilon)$. The depth in this case is equal to $\depth(\cU^\epsilon)$. 

\end{proof}

\subsection{Proof of Theorem \ref{thm:nn-to-pido}}\label{app:nn-to-pido}

\begin{proof}

\textbf{Step 1: construction. }

Let $\epsilon>0$ and $n,N\in\N$. We first introduce some notation. Let $\cJ_N = \{0,\ldots, 2N+1\}^d$, $\cK_N = \{-N, \ldots, N\}^d$, let $\{\fb_j\}_{j\in\N}$ be an ordered Fourier basis, as described in \textbf{SM} \ref{app:Fourier}, and let $\{\widehat{\fb}_j\}_{j\in\N}$ be a neural network approximation of the same basis such that
\begin{equation}\label{eq:fb-nn}
    \max_{k\in\cK_N}\norm{\fb_k-\widehat{\fb}_k}_{C^r} \leq \eta, 
\end{equation}
cf. Lemma \ref{lem:rec-fourier}. Using notation from \textbf{SM} \ref{app:trig-pol}, let $\cQ_N:\R^{\abs{\cJ_N}}\to C(\T^d)$ be the trigonometric polynomial interpolation operator as in \eqref{eq:trig-pol} and let $\cE_N:C(\T^d)\to\R^{\abs{\cJ_N}}$ be the encoder as in \eqref{eq:encoder}. We define
\begin{equation}\label{eq:qhat}
    \cQhat_{N}:\R^{\abs{\cJ_N}}\to C(\T^d): y \mapsto \frac{1}{\abs{\cK_N}} \sum_{k\in \cK_N} \sum_{j\in \cJ_N} y_j a_{k,j} \widehat{\fb}_k, 
\end{equation}
with coefficients $a_{k,j}$ as in \eqref{eq:trig-pol-coeff}, as a neural network approximation of $\cQ_N$. 

Inspired by the proof of Theorem \ref{thm:nn-to-pinn} (and using its notation as well), we define $\cGhat: C(\T^d)\to L^2(\mu)$ by
\begin{equation}
    \cGhat(u_0)(t,x) =  \sum_{m=1}^M\sum_{i=0}^{s-1}  \frac{\Delta^{i,s-i}_{1/M}[\cU^\epsilon(\cQ_Z\circ\cE_Z\circ u_0, t_m)](t_m,x)}{M^{-i} i!}\cdot \varphi_i^\delta(t-t_m)   \Phi^M_m(t), 
\end{equation}
Then it holds that
\begin{equation}
\begin{split}
    (\cQ_N\circ\cE_N\circ\cGhat)(u_0)(t,x) &=  \sum_{k\in \cK_N} \sum_{j\in \cJ_N}  \sum_{m=1}^M\sum_{i=0}^{s-1} \frac{a_{k,j}}{\abs{\cK_N}} \frac{\Delta^{i,s-i}_{1/M}[\cU^\epsilon(\cQ_Z\circ\cE_Z\circ u_0, t_m)](t_m,x_j)}{M^{-i} i!}\cdot \Psi_{i,m,k}(t,x)\\
    \Psi_{i,m,k}(t,x) &= \varphi_i^\delta(t-t_m)   \Phi^M_m(t)  \fb_k(x).
\end{split}
\end{equation}
Now for every $i, m, k$ let $\Psi_{i,m,k}:[0,T]\times \T^d\to \R$ be defined as, 
\begin{equation}
    \Psihat_{i,m,k}(t,x) = \hattimes_\delta\left(\varphi_i^\delta(t-t_m),\Phi^M_m(t),\widehat{\fb}
    _k(x)\right), 
\end{equation}
where $\hattimes_\delta$ is a neural network approximation of the multiplication operator. We can then construct a DeepONet as
\begin{equation}\label{eq:def-pido}
    \begin{split}
 \cG_\theta(u_0)(t,x) &= \sum_{j=1}^p \beta_j(u_0)\tau_j(t,x)\\
 &= \sum_{i=0}^{s-1}\sum_{m=1}^M \sum_{k\in \cK_N} \left[\sum_{j\in \cJ_N}   \frac{a_{k,j}}{\abs{\cK_N}} \frac{\Delta^{i,s-i}_{1/M}[\cU^\epsilon(\cQ_Z\circ\cE_Z\circ u_0, t_m)](t_m,x_j)}{M^{-i} i!}\right] \cdot \Psihat_{i,m,k}(t,x).
    \end{split}
\end{equation}
We see that we need to set $p=sM(2N+1)^d$ and that the trunk nets are given by $\tau_j\sim \Psihat_{i,m,k}$, up to a different indexing. 

\textbf{Step 2: error estimate.}
First we use Assumption \ref{ass:diff-operator} to see that
\begin{equation}
    \norm{\cL(\cG-\cG_\theta)}_{L^2} \leq C \sum_{k,\bmalpha} \norm{D^{(k,\bmalpha)}(\cG-\cG_\theta)}_{L^2}. 
\end{equation}
Next, we observe that using Assumption \ref{ass:nn}, Assumption \ref{ass:stability} and \eqref{eq:QEu0} it holds that for all $t$,
\begin{equation}\label{eq:QEu0-Z}
    \norm{(\cU^\epsilon(\cQ_Z\circ \cE_Z\circ u_0)-\cG(u_0))(\cdot,t)}_{L^2} \leq \epsilon + C^\epsilon_{\mathrm{stab}}CBZ^{-r+d/p^*}. 
\end{equation}
One can then use Theorem \ref{thm:nn-to-pinn}, but by replacing $\epsilon$ by \eqref{eq:QEu0-Z} in the error bound \eqref{eq:pinn-acc}, to find that
\begin{equation}
    \norm{D^{(k,\bmalpha)}(\cG-\cGhat)}_{L^2}\leq C\ln^k(M)(\norm{u}_{C^{(s,\ell)}}M^{k-s}+M^{2k}((\epsilon+C^\epsilon_{\mathrm{stab}}Z^{-r+d/p^*})h^{-\ell}+C^{CB}_{\epsilon,\ell}h^{r-\ell}))
\end{equation}
Then, using the observation that $D^{(k,\bmalpha)}(\Id-\cQ_N\circ \cE_N)\cGhat = D^\bmalpha_x(\Id-\cQ_N\circ \cE_N)D^k_t\cGhat$ we find that
\begin{equation}
     \norm{D^{(k,\bmalpha)}(\Id-\cQ_N\circ \cE_N)\cGhat(u_0)}_{L^2} \leq CN^{-(r-\ell)}\norm{D^k_t\cGhat(u_0)}_{H^{r}},
\end{equation}
which can be combined with the estimate
\begin{equation}
    \norm{D^k_t\cGhat(u_0)}_{H^{r}} \leq M^{s-1}\cdot M^k \ln^k(M) \norm{\cU^\epsilon(\cQ_Z\circ \cE_Z \circ u_0)}_{H^{r}} \leq M^{s+k-1} \ln^k(M) C_{\epsilon,r}^{\Bar{B}},
\end{equation}
where we used that for $u_0\in\cX$ with $\norm{u_0}_{C^r}\leq B$ it holds $\norm{(\cQ_N\circ\cE_N)(u_0)}_{H^r(\T^d)} \leq C B=:\Bar{B}$. Next, we make the rough estimate that,
\begin{equation}
    \norm{D^{(k,\bmalpha)}(\cQhat_N-\cQ_N)\circ \cE_N)\cGhat(u_0)}_{L^2} \leq C N^d M^{s+k-1}\ln^k(M)\max_k\norm{\fb_k-\fbhat_k}_{C^r}. 
\end{equation}
Finally, using Lemma \ref{lem:wlog-faa-di-bruno} we find that
\begin{equation}
    \norm{D^{(k,\bmalpha)}(\cQhat_N\circ \cE_N\circ \cGhat-\cG_\theta)(u_0)}_{L^2}\leq \delta. 
\end{equation}
By setting $\eta = N^{\ell-r-d}$, $h=1/N$ and using that $M^{2k}\leq M^{k+s-1}$ and $C_{\epsilon,\ell}^{\Bar{B}}\leq C_{\epsilon,r}^{\Bar{B}}$ we find, 
\begin{equation}
    \norm{\cL(\cG-\cG_\theta)}_{L^2} \leq C\ln^k(M)(\norm{u}_{C^{(s,\ell)}}M^{k-s}+M^{k+s-1}((\epsilon+C^\epsilon_{\mathrm{stab}}Z^{-r+d/p^*})N^{\ell}+C^{\Bar{B}}_{\epsilon,r}N^{\ell-r})). 
\end{equation}
We conclude by using that $\ln^k(M)\leq CM^\rho$ for any $\rho>0$. 

\textbf{Step 3: size estimate.} It follows immediately that $\depth(\branch)=\depth(\cU^\epsilon)$, $\width(\branch) = \bigO(M(Z^d+N^d\width(\cU^\epsilon)))$, 
$\depth(\trunk) = 3$ and $\width(\trunk) = \bigO(MN^d(N+\ln(N)))$. 

\end{proof}

\subsection{Proof of Theorem \ref{thm:generalization}}\label{app:generalization}

\begin{proof}
Define the random variable $Y = \Eg(\theta^*(\S))^2-\Et(\theta^*(\S),\S)^2$.
Then if follows from equation (4.8) in the proof of \cite[Theorem 5]{deryck2021pinn} that 
\begin{equation}
    \mathbb{P}(Y > \epsilon^2) \leq \left(\frac{2R\mathfrak{L}}{\epsilon^2}\right)^{d_\Theta} \exp\left(\frac{-2\epsilon^4n}{c^2}\right),
\end{equation}
since $\mathbb{P}(Y > \epsilon^2) = 1 - \Prob{\cA}$, where $\cA$ is as defined in the proof of \cite[Theorem 5]{deryck2021pinn}. It follows that
\begin{equation}\label{eq:EY}
\begin{split}
    \mathbb{E}[Y] = \mathbb{E}[Y \mathbbm{1}_{Y\leq \epsilon^2}] + \mathbb{E}[Y \mathbbm{1}_{Y> \epsilon^2}] \leq \epsilon^2 + c \Prob{Y> \epsilon^2}. 
\end{split}
\end{equation}
Setting $\epsilon^2 = c \Prob{Y> \epsilon^2}$ leads to
\begin{equation}
    \mathbb{E}[Y] \leq 2\epsilon^2 = \sqrt{\frac{2c^2}{n}\ln(\frac{c}{\epsilon^2}\left(\frac{2R\mathfrak{L}}{\epsilon^2}\right)^{d_\Theta})}. 
\end{equation}
For $\epsilon<1$, and using that $\ln(x)\leq \sqrt{x}$ for all $x>0$, this equality implies that 
\begin{equation}
    \epsilon^{{d_\Theta}+1} \leq \frac{2c^3(2R\mathfrak{L})^{{d_\Theta}/2}}{n}. 
\end{equation}
Hence, we find that if $n\geq 2c^2e^8/(2R\mathfrak{L})^{{d_\Theta}/2}$ then $\epsilon^{{d_\Theta}+1}\leq ce^{-8}(2R\mathfrak{L})^{d_\Theta}$ which implies that 
\begin{equation}\label{eq:2sqrt2}
    \left[\ln(\frac{c}{\epsilon^2}\left(\frac{2R\mathfrak{L}}{\epsilon^2}\right)^{d_\Theta})\right]^{-1/2} \leq \frac{1}{2\sqrt{2}}. 
\end{equation}
Using once more that $\epsilon^2 = c \Prob{Y> \epsilon^2}$ and \eqref{eq:2sqrt2} gives us, 
\begin{equation}
    \begin{split}
\mathbb{E}[Y] &\leq \sqrt{\frac{2c^2}{n}\ln(c(2R\mathfrak{L})^{{d_\Theta}}\left(\frac{\sqrt{2n}}{c}\left[\ln(\frac{c}{\epsilon^2}\left(\frac{2R\mathfrak{L}}{\epsilon^2}\right)^{d_\Theta})\right]^{-1/2}\right)^{{d_\Theta}+1})}\\
&\leq \sqrt{\frac{2c^2}{n}\ln((a\mathfrak{L}\sqrt{n})^{{d_\Theta}+1})} = \sqrt{\frac{2c^2({d_\Theta}+1)}{n}\ln(a\mathfrak{L}\sqrt{n})}.
    \end{split}
\end{equation}
\end{proof}

\section{Additional material for Section \ref{sec:CoD}}\label{app:AC}

\subsection{Auxiliary results}

\begin{lemma}\label{lem:exp-to-prob}
Let $\epsilon>0$, let $(\Omega, \mathcal{F}, \mathcal{P})$ be a probability space, and let $X:\Omega\to\mathbb{R}$ be a random variable that satisfies $\E{\abs{X}}\leq \epsilon$. Then it holds that $\mathbb{P}(\abs{X}\leq \epsilon)>0$. 
\end{lemma}
\begin{proof}
This result is \cite[Proposition 3.3]{grohs2018proof}.
\end{proof}

\begin{lemma}
\label{lem:recursion1}
Let
$ \gamma \in \{0,1\}$,
$ \beta \in [1,\infty) $,
$
  \alpha_0, \alpha_1, x_0, x_1, x_2, \ldots \in [0,\infty) 
$
satisfy
for all $ k \in \N_0 $
that
\begin{equation}\label{eq:full_history_eq}
  x_k
\leq
  \1_{ \N }( k )
  (\alpha_0+\alpha_1k)\beta^k
  +
  \sum_{ l = 0 }^{ k - 1 }
  \left(k - l\right)^\gamma 
  \beta^{ ( k - l ) }
  \left[
    x_l
    +
    x_{ \max\{ l - 1 , 0 \} }
  \right]
  .
\end{equation}
Then it holds for all 
$ k \in \N_0 $ that
\begin{equation}\label{eq:a_bd}
\begin{split}
x_k \le 
\frac{(\alpha_0 + \alpha_1) \beta^k \1_\N(k)}{(4 + \gamma)^{1/2} (1 + 2^{(1+\gamma)/2})^{-k}}
= 
\begin{cases}
\1_{ \N }( k )(\alpha_0+\alpha_1) 2^{-1} (1+ 2^{1/2} )^k \beta^ k & \colon \gamma = 0 \\[0.5em]
\1_{ \N }( k ) (\alpha_0+\alpha_1) 5^{-1/2} (3\beta) ^ k & \colon \gamma = 1.
\end{cases}
\end{split}
\end{equation}
\end{lemma}

\begin{proof}
This result is \cite[Corollary 4.3]{aj2021mlp}.
\end{proof}

\begin{lemma}\label{lem:recursion2}
Let $\alpha\in [1,\infty)$, $x_0,x_1, \ldots \in [0,\infty)$ satisfy for all $k\in \N_0$ that
${x_{k}\leq \alpha x_{k-1}^{k}}$. Then it holds for all $k\in\N_0$ that
\begin{equation}
    x_{k}\leq \alpha^{(k+1)!}x_0^{k!}
\end{equation}
\end{lemma}
\begin{proof}
We provide a proof by induction. First of all, it is clear that $x_0 \leq \alpha x_0$. For the induction step, assume that $x_{k-1}\leq \alpha^{k!}x_0^{(k-1)!}$ for an arbitrary $k\in\N_0$. We calculate that
\begin{equation}
    x_k \leq \alpha \left(\alpha^{k!}x_0^{(k-1)!}\right)^k \leq \alpha^{(k+1)!} x^{k!}_0. 
\end{equation}
This proves the statement. 
\end{proof}

\begin{lemma}\label{lem:faa-di-bruno}
Let $\ell\in \N$, $f\in C^\ell(\R,\R)$, $h\in C^\ell(\T^d, \R)$ and let $B_\ell$ denote the $\ell$-th Bell number. Then it holds that
\begin{equation}
     \abs{ f\circ h}_{C^\ell(\R)} \leq \norm{f}_{C^\ell(\R)}\left(B_\ell\norm{h}_{C^{\ell-1}(\T^d)}^\ell+ \abs{ h}_{C^\ell(\T^d)}\right).
\end{equation}
\end{lemma}

\begin{proof}
Let $\Pi$ be the set of all partitions of the set $\{1, \ldots, \ell\}$, let $\alpha\in \N_0^d$ such that $\norm{\alpha}_1=\ell$ and let $\iota:\N^\ell\to\N^d$ be a map such that $D^\alpha = \frac{\partial^\ell}{\prod_{j=1}^\ell x_{\iota(j)}}$. Then the Faà di Bruno formula can be reformulated as \cite{constantine1996multivariate}, 
\begin{equation}
\begin{split}
    D^\alpha f(h(x)) &= \sum_{\pi\in\Pi}f^{(\abs{\pi})}(h(x))\cdot \prod_{B\in\pi} \frac{\partial^{\abs{B}}h(x)}{\prod_{j\in B} \partial x_{\iota(j)}}\\
    &= \sum_{\substack{\pi\in\Pi,\\\abs{\pi}\geq 2}}f^{(\abs{\pi})}(h(x))\cdot \prod_{B\in\pi} \frac{\partial^{\abs{B}}h(x)}{\prod_{j\in B} \partial x_{\iota(j)}} + f'(h(x)) D^\alpha h(x). 
\end{split}
\end{equation}
Combining this formula with the definition of the Bell number as $B_\ell = \abs{\Pi}$, we find the following upper bound, 
\begin{equation}
\begin{split}
    \abs{ f\circ h}_{C^\ell(\R)} &\leq \sum_{\pi\in\Pi}\norm{f}_{C^\ell(\R)}\norm{h}_{C^{\ell-1}(\R)}^\ell +  \norm{f}_{C^1(\R)} \abs{ h}_{C^\ell(\R)}\\
    &\leq \norm{f}_{C^\ell(\R)}\left(B_\ell\norm{h}_{C^{\ell-1}(\R)}^\ell+ \abs{ h}_{C^\ell(\R)}\right).
\end{split}
\end{equation}
\end{proof}

\subsection{Proof of Theorem \ref{thm:kolmogorov}}\label{app:kolmogorov}

\begin{definition}\label{def:lq}
Let $(\Omega, \mathcal{F}, \mu)$ be a measure space and let $q>0$. For every $\mathcal{F}/\mathcal{B}(\mathbb{R}^d)$-measurable function $f:\Omega\to\mathbbm{R}^d$, we define
\begin{equation}
    \norm{f}_{\mathcal{L}^q(\mu, \norm{\cdot}_{\mathbb{R}^d})} := \left[\int_\Omega\norm{f(\omega)}_{\mathbb{R}^d}^q \mu(d\omega)\right]^{1/q}.
\end{equation}
\end{definition}

Let $(\Omega, \mathcal{F}, P, (\mathbb{F}_t)_{t\in[0,T]})$ be a stochastic basis, $D\subseteq \mathbb{R}^d$ a compact set and, for every $x\in D$, let $X^x : \Omega\times[0,T]\to\mathbb{R}^d$ be the solution, in the Itô sense, of the following stochastic differential equation, 
\begin{equation}\label{eq:Itô-diffusion-sde}
    dX^x_t = \mu(X^x_t)dt + \sigma(X^x_t)dB_t, \quad X^x_0=x, \quad x\in D, t\in[0,T],
\end{equation}
where $B_t$ is a standard $d$-dimensional Brownian motion on $(\Omega, \mathcal{F}, P, (\mathbb{F}_t)_{t\in[0,T]})$. The existence of $X^x$ is guaranteed by \cite[Theorem 4.5.1]{nasode}.

As in \cite[Theorem 3.3]{deryck2021pinn} we define $\rho_d$ as
\begin{equation}\label{eq:rhod}
    \rho_d :=  \max_{x\in D}\sup_{\substack{s,t \in [0,T],\\s<t}} \frac{\norm{X^x_s-X^x_t}_{\mathcal{L}^q(P, \norm{\cdot}_{\mathbb{R}^d})}}{\abs{s-t}^{\frac{1}{p}}} <\infty,  
\end{equation}
where $X^x$ is the solution, in the Itô sense, of the SDE \eqref{eq:Itô-diffusion-sde}, $q>2$ is independent of $d$ and $\norm{\cdot}_{\mathcal{L}^q(P, \norm{\cdot}_{\mathbb{R}^d})}$ is as in Definition \ref{def:lq}.

\begin{lemma}\label{lem:kolmogorov}
In Setting \ref{set:kolmogorov}, Assumption \ref{ass:nn} and Assumption \ref{ass:stability} are satisfied with
\begin{equation}
    \norm{u(\cdot, t)-\cU^\epsilon(\varphi, t)}_{L^2(\mu)} \leq \epsilon, \qquad C^B_{\epsilon, \ell} = CB\cdot \poly(d\rho_d), \qquad C^\epsilon_{\mathrm{stab}} = 1, \qquad p=\infty,
\end{equation}
where $t\in[0,T]$ and $\varphi\in C^2_0(\R^d)$. 
Moreover, there exists $C^*>0$ (independent of $d$) for which it holds that $\depth(\cU^\epsilon) \leq C^* \depth(\varphihat^\epsilon)$ and $\{\width,\size\}(\cU^\epsilon) \leq C^* \epsilon^{-2}  \{\width,\size\}(\varphihat^\epsilon) $. 
\end{lemma}
\begin{proof}
It follows from the Feynman-Kac formula that $u(t,x) = \E{\varphi(X^x_t)}$ \cite{oksendal2003stochastic}. Replacing $\varphi$ by a neural network $\varphihat^\epsilon$ with $\norm{\varphi-\varphihat^\epsilon}_{C^0} \leq  \epsilon$ gives us for any probability measure $\mu$ that,
\begin{equation}\label{eq:kolmogorov1}
    \norm{\E{\varphi(X^x_t)}-\E{\varphihat^\epsilon(X^x_t)}}_{L^2(\mu)}\leq \norm{\varphi-\varphihat^\epsilon}_{C^0}. 
\end{equation}

Using \cite[Lemma A.2]{deryck2021pinn} (which is based on \cite{grohs2018proof}) we find, 
\begin{equation}
    \E{(I)} := \E{\left(\int_{D}\abs{\E{\varphihat^\epsilon(X^x_t)}-\frac{1}{m}\sum_{i=1}^m \varphihat^\epsilon(X^x_t(\omega_m))}^2\mu(dx)\right)^{1/2}} \leq \frac{2\norm{\varphihat^\epsilon}_{C^0}}{\sqrt{m}}.
\end{equation}

From \cite[Lemma A.5]{deryck2021pinn}, for all $x\in\mathbb{R}^d$, $t\in[0,T]$ and $\omega\in\Omega$ it holds that
    \begin{equation}
    X^x_t(\omega) = \sum_{i=1}^d \left(X^{e_i}_t(\omega)-X^0_t(\omega)\right)x_i + X^0_t(\omega).
\end{equation}
Using this equality, together with Hölder's inequality and the boundedness of $\norm{X^x_t}_{L^p}$ \cite[Lemma A.5]{deryck2021pinn} we find that,
\begin{equation}
\begin{split}
    \E{(II_\bmalpha)}&:=\E{\left(\int_{D}\abs{\E{D^\bmalpha_x \varphihat^\epsilon(X^x_t)}-\frac{1}{m}\sum_{i=1}^m D^\bmalpha_x\varphihat^\epsilon(X^x_t(\omega_m))}^2\mu(dx)\right)^{1/2}}\\
    &\leq C \cdot \poly(d\rho_d)\cdot  \E{\left(\int_{D}\abs{\E{ \varphihat^\epsilon(X^x_t)}-\frac{1}{m}\sum_{i=1}^m \varphihat^\epsilon(X^x_t(\omega_m))}^4\mu(dx)\right)^{1/4}}\\
    &\leq CB \cdot \poly(d\rho_d)
\end{split}
\end{equation}
Combining the previous results gives us,
\begin{equation}
    \E{\sqrt{m}\cdot (I) + \sum_{\norm{\bmalpha}_1\leq \ell}(II_\bmalpha)} \leq CB \cdot \poly(d\rho_d).
\end{equation}
If we combine this with Lemma \ref{lem:exp-to-prob} then we find the existence of $(\omega^*_i)_{i=1}^m$ such that
for 
\begin{equation}
    \cU^\epsilon(\varphi, t)(x) = \frac{1}{m}\sum_{i=1}^m \varphihat^\epsilon\left(\sum_{i=1}^d \left(X^{e_i}_t(\omega^*_i)-X^0_t(\omega^*_i)\right)x_i + X^0_t(\omega^*_i)\right) 
\end{equation}
it holds that
\begin{equation}
    \norm{\E{\varphihat^\epsilon(X^x_t)}-\cU^\epsilon(\varphi, t)}_{L^2(D)} \leq \frac{2C}{\sqrt{m}}. 
\end{equation}
and by setting $m = \epsilon^{-2}$ and  \eqref{eq:kolmogorov1} we find that,
\begin{equation}
    \norm{u(\cdot, t)-\cU^\epsilon(\varphi, t)}_{L^2(D)} \leq \epsilon, \qquad C^B_{\epsilon, \ell} = CB\cdot \poly(\rho_d), \qquad C^\epsilon_{\mathrm{stab}} = 1, \qquad p=\infty. 
\end{equation}
Moreover, it holds that
\begin{equation}
    \depth(\cU^\epsilon) \leq C^* \depth(\varphihat^\epsilon), \qquad \{\width,\size\}(\cU^\epsilon) \leq C^* \epsilon^{-2}  \{\width,\size\}(\varphihat^\epsilon) 
\end{equation}
where we write $C^* = C \poly(d\rho_d) $. 
\end{proof}

We can now present the proof of the actual theorem. 

\begin{proof}[Proof of Theorem \ref{thm:kolmogorov}]
We use Theorem \ref{thm:nn-to-pinn} with $k=1$ and $\ell=2$ and combine the result with Lemma \ref{lem:kolmogorov}. We find that for every $M\in\N$ and $\delta, h>0$ it holds that,
\begin{equation}
\begin{split}
     &\norm{\cL(\uhat-u)}_{L^q([0,T]\times D)} + \norm{\uhat-u}_{L^2(\partial([0,T]\times D))} \\
     &\qquad \leq CB\cdot \poly(d\rho_d)\cdot \ln(M)(\norm{u}_{C^{(1,2)}}M^{1-s}+M^{2}(\delta h^{-2}+h^{r-2})). 
\end{split}
\end{equation}
Set $\delta=h^r$, $M^{-1-s} = \delta^{1-2/r}$ we find that
\begin{equation}
    \norm{\cL(\uhat-u)}_{L^q([0,T]\times D)} + \norm{\uhat-u}_{L^2(\partial([0,T]\times D))} \leq CB\cdot \poly(d\rho_d) \ln(1/\delta) \delta^{\frac{r-2}{r}\frac{s-1}{s+1}}
\end{equation}
Using that $\ln(M)\leq CM^\sigma$ for arbitrarily small $\sigma>0$, we find that we should set
\begin{equation}
    \delta = \epsilon^{\frac{r+\sigma}{r-2}\frac{s+1}{s-1}},\qquad M = \epsilon^{\frac{-1-\sigma}{s-1}}
\end{equation}
\end{proof}

\subsection{Nonlinear parabolic equations}\label{app:sem-lin-par}

Some examples of nonlinear parabolic PDEs of the type \eqref{eq:AC} are:

\begin{itemize}
    \item The \textit{Kolmogorov–Petrovsky–Piskunov (KPP) equation} \cite{kolmogorov1937etude} is a celebrated model that is often used to model wave propagation and population genetics. The model is particularly useful for systems that exhibit phase transitions. One obtains the KPP equation if one chooses a sufficiently smooth nonlinearity $F$ that satisfies the requirements $F(0)=F(1)=0$, $F'(0)=r>0$, $F(u)>0$ and $F'(u)<r$ for all $0<u<1$. Well-known examples include the \textit{Fisher equation} \cite{fisher1937wave} with $F(u)=ru(1-u)$ and the \textit{Allen-Cahn equation} \cite{allen1979microscopic} with $F(u)=ru(1-u^2)$.
    \item \textit{Branching diffusion processes } give a probabilistic representation of the KPP equation for the case where $F(u) = \beta(\sum_{k=0}^\infty a_k u^k-u)$ with $a_k\geq 0$ and $\sum_k a_k = 1$. In this setting, the PDE \eqref{eq:AC} describes a $d$-dimensional branching Brownian motion, where every particle in the system dies in an an exponential time of parameter $\beta$ and created $k$ i.i.d. descendants with probability $a_k$ \cite{henry2014numerical, mckean1975application}. 
    \item Finally, the PDE \eqref{eq:AC} arises in the context of credit valuation adjustment when pricing derivative contracts to compute the counterparty risk valuation, e.g. \cite{henry2012counterparty}. The dimension $d$ corresponds to the number of underlying assets and can be very high. 
\end{itemize}

\subsection{Multilevel Picard approximations}

In what follows, we will provide a definition of a particular kind of MLP approximation (cf. \cite{aj2021mlp}) and a theorem that quantifies the accuracy of the approximation. First, we rigorously introduce the setting of the nonlinear parabolic PDE \eqref{eq:AC} that is under consideration, cf. \cite[Setting 3.2 with $p\leftarrow0$]{aj2021mlp}. We choose the $d$-dimensional torus $\T^d=[0,2\pi)^d$ as domain and impose periodic boundary conditions. This setting allows us to use the results of \cite{aj2021mlp}, which are set in $\R^d$, and yet still consider a bounded domain so that the error can be quantified using an uniform probability measure. 

\begin{setting}\label{set:allen-cahn}
Let $d, m \in \N$, $T, L, \cL \in [0,\infty)$, 
let $(\T^d, \cB(\T^d), \mu)$ be a probability space where $\mu$ is the rescaled Lebesgue measure,
let $g \in C(\T^d,\R)\cap L^2(\mu)$, 
let $F \in C(\R,\R)$, assume for all  $x\in\T^d$, $y,z\in\R$ that
\begin{equation}\label{eq:fn_cond}
\abs{ F(y)-F(z) } \leq L \abs{ y-z }, \qquad \max\{\abs{ F(y) }, \abs{ g(x) } \} \le \cL. 
\end{equation}
Let $u_d\in C^{1,2}([0,T]\times \R^d, \R)\cap L^2(\mu)$ satisfy for all $t\in [0,T], x\in \R^d$ that
\begin{equation}\label{eq:PDE-AC}
   ( \partial_t u_d)(t,x) = (\Delta_x u_d)(t,x) + F(u_d(t,x)), \qquad u_d(0,x) = g(x). 
\end{equation}
Assume that for every $\epsilon>0$ there exists a neural network $\Fhat_\epsilon$, a neural network $\ghat_\epsilon$ and a neural network $\cI_\epsilon$ with depth $\depth(\cI_\epsilon)=\depth(\Fhat_\epsilon)$ such that
\begin{equation}\label{eq:nn-expressive}
     \norm{\Fhat_\epsilon-F}_{C^0(\R)} \leq \epsilon, \quad \norm{\ghat_\epsilon-g}_{L^2(\mu)} \leq \epsilon, \quad \norm{\cI_\epsilon-\Id}_{C^0([-1-\cL,1+\cL])}\leq \epsilon. 
\end{equation}
\end{setting}

Note that for some of the equations introduced in Section \ref{app:sem-lin-par} the nonlinearity $F$ might not be globally Lipschitz and hence does not satisfy \eqref{eq:fn_cond}. However, it is easy to argue or rescale $g$ \cite{LMK1, beck2020overcoming} such that $u_d$ is globally bounded by some constant $C$. For instance, for the Allen-Cahn equation it holds that if $\norm{g}_{L^\infty}\leq 1$ then $\norm{u_d(t,\cdot)}_{L^\infty}\leq 1$ for any $t\in[0,T]$ \cite{yang2018uniform}.
One can then define a `smooth', globally Lipschitz, bounded function $\Tilde{F}:\R\to\R$ such that $\Tilde{F}(v) = F(v)$ for $\abs{v}\leq C$ and such that $\Tilde{F}(v)=0$ for $\abs{v}>2C$. This will then also ensure the existence of a neural network $\Fhat$ that is close to $\Tilde{F}$ in $C^0(\R)$-norm. 

In this setting, multilevel Picard approximations can be introduced. We follow the definition of \cite{aj2021mlp}.

\begin{definition}[MLP approximation]\label{def:MLP}
Assume Setting \ref{set:allen-cahn}. Let $\Theta = \bigcup_{n\in\N}\! \Z^n$, let $(\Omega,\cF,\P)$ be a probability space,
let $\fu^\theta \colon \Omega \to [0,1]$, $\theta \in \Theta$, be i.i.d.\ random variables, 
assume for all $\theta\in\Theta$, $r\in (0,1)$ that $\P(\fu^\theta \le r ) = r$,
let $\EuScript{U}^\theta \colon [0,T]\times \Omega\to [0,T]$, $\theta\in\Theta$, satisfy for all $t\in [0,T]$, $\theta\in\Theta$ that $\EuScript{U}_t^\theta = t + (T-t)\fu^\theta$, 
let $W^\theta \colon [0,T] \times \Omega \to \R^d$, $\theta\in\Theta$, be independent standard Brownian motions,
assume that $(\EuScript{U}^\theta)_{\theta\in\Theta}$ and $(W^\theta)_{\theta\in\Theta}$ are independent,
and let $\mlp_n^\theta \colon [0,T] \times \T^d\times \Omega\to \R$, 
$n\in\Z$, 
$\theta\in\Theta$, satisfy for all 
$n\in\N_0$, 
$\theta\in\Theta$, $t\in[0,T]$, $x\in\T^d$ that 
\begin{equation}\label{eq:mlp}
\begin{split}
\mlp_n^\theta(t,x) 
& = \frac{\1_\N(n)}{m^n} \br[\Bigg]{ \sum_{k=1}^{m^n} g\pr[\bigl]{x + W_{T-t}^{(\theta,0,-k)}} } \\
& \quad + \sum_{i=0}^{n-1} \frac{(T-t)}{m^{n-i}}\br[\Bigg]{ \sum_{k=1}^{m^{n-i}} \pr[]{ F(\mlp_i^{(\theta,i,k)}) - \1_\N(i) F( \mlp_{i-1}^{(\theta,-i,k)}) } \pr[]{ \EuScript{U}_t^{(\theta,i,k)}, x + W_{\EuScript{U}_t^{(\theta,i,k)}-t}^{(\theta,i,k)} } }.
\end{split}
\end{equation}
\end{definition}

\begin{example}
In order to improve the intuition of the reader regarding Definition \ref{def:MLP}, we provide explicit formulas for the multilevel Picard approximation \eqref{eq:mlp} for $n=0$ and $n=1$,
\begin{equation}
    U^\theta_0(t,x) = 0 \quad \text{and} \quad U^\theta_1(t,x) = \frac{1}{m} \br[\Bigg]{ \sum_{k=1}^{m} g\pr[\bigl]{x + W_{T-t}^{(\theta,0,-k)}} } + (T-t) F(0). 
\end{equation}
\end{example}

Finally, we provide a result on the accuracy of MLP approximations at single space-time points. 

\begin{theorem}\label{thm:mlp}
It holds for all $n\in\N_0$, $t\in [0,T]$, $x\in \T^d$ that
\begin{equation}
    \left(\E{\abs{U^0_n(t,x)-u(t,x)}^2}\right)^{1/2} \leq \frac{\cL (T+1)\exp(LT)(1+2LT)^n}{m^{n/2}\exp(-m/2)}. 
\end{equation}
\end{theorem}
\begin{proof}
This result is \cite[Corollary 3.15]{aj2021mlp} with $p\leftarrow 0$, $\mathfrak{p}\leftarrow 2$ and $\cL \leftarrow \cL/2$. 
\end{proof}

\subsection{Neural network approximation of nonlinear parabolic equations}\label{app:nnt-ac}

In this section, we will prove that the solution of the nonlinear parabolic PDE as in Setting \ref{set:allen-cahn} can be approximated with a neural network without the curse of dimensionality. At this point, we do not specify the activation function, with the only restriction being that the considered neural networks should be expressive enough to satisfy \eqref{eq:nn-expressive}. By emulating an MLP approximation and using that $F$, $g$ and the identity function can be approximated using neural networks, the following theorem can be proven. 

\begin{theorem}\label{thm:nn}
Assume Setting \ref{set:allen-cahn}. For every $\epsilon, \sigma>0$ and $t\in [0,T]$ there exists a neural network $\uhat_\epsilon:\T^d\to\R$ such that
\begin{equation}
     \norm{\uhat_\epsilon(\cdot)-u(t,\cdot)}_{L^2(\mu)} \leq \epsilon. 
\end{equation}
In addition, $\uhat$ satisfies that
\begin{equation}
    \begin{split}
          &\depth(\uhat_\epsilon) \leq \depth(\ghat_{\delta})+\log_{C_2} (3C_1\exp(m/2)/\epsilon)\depth(\Fhat_{\delta}),\\
          &\width(\uhat_\epsilon), \size(\uhat_\epsilon) \leq  (\size(\ghat_{\delta})+ \size(\Fhat_{\delta})+ \size(\cI_{\delta})) \left(\frac{4C_1\exp(m/2)}{\epsilon}\right)^{2+3\sigma}, 
    \end{split}
\end{equation}
where
\begin{equation}
    \begin{split}
        &C_1 = (T+1)(1+\cL \exp(LT)), \qquad C_2 = 5+3LT,\\
        &\delta = \frac{\epsilon^2}{9C_1^2\exp(m/2)}, \qquad m = C_2^{2(1+1/\sigma)}.
    \end{split}
\end{equation}
\end{theorem}
\begin{proof}
\textbf{Step 1: construction of the neural network. }
Let $\epsilon,\delta>0$ be arbitrary and let $\Fhat = \Fhat_\delta$, $\ghat=\ghat_\delta$ and $\cI=\cI_\delta$ as in Setting \ref{set:allen-cahn}. We then define for all $n\in\N$ and $\theta\in\Theta$, 
\begin{equation}\label{eq:nn-mlp}
\begin{split}
\Uhat_n^\theta(t,x) 
& = \frac{\1_\N(n)}{m^n} \br[\Bigg]{ \sum_{k=1}^{m^n} (\cI^{n-1}\circ\ghat)\pr[\bigl]{x + W_{T-t}^{(\theta,0,-k)}} } \\
& \quad + \sum_{i=0}^{n-1} \frac{(T-t)}{m^{n-i}}\br[\Bigg]{ \sum_{k=1}^{m^{n-i}} \pr[]{ (\cI^{n-i-1}\circ\Fhat)(\Uhat_i^{(\theta,i,k)}) - \1_\N(i) (\cI^{n-i}\circ\Fhat)( \Uhat_{i-1}^{(\theta,-i,k)}) } \pr[]{ \EuScript{U}_t^{(\theta,i,k)}, x + W_{\EuScript{U}_t^{(\theta,i,k)}-t}^{(\theta,i,k)} } }, 
\end{split}
\end{equation}
with notation and random variables cf. Definition \ref{def:MLP}. Note that for every $t\in[0,T]$, $n\in\N$, $\theta\in\Theta$, every realization of the random variable $\Uhat_n^\theta(t,\cdot)$ is a neural network that maps from $\T^d$ to $\R$. 

Let $n\in\N_0$, $m\in\N$ and $t\in [0,T]$ be arbitrary. 
Integrating the square of the error bound of Theorem \ref{thm:mlp} and Fubini's theorem tell us that
\begin{equation}\label{eq:fubini}
\begin{split}
   \E{\int_{\T^d}\abs{U^0_n(t,x)-u(t,x)}^2d\mu(x)} &= \int_{\T^d} \E{\abs{U^0_n(t,x)-u(t,x)}^2} d\mu(x)\\
   &\leq \frac{4\cL^2 (T+1)^2\exp(2LT)(1+2LT)^{2n}}{m^{n}\exp(-m)}. 
\end{split}
\end{equation}
From Lemma \ref{lem:exp-to-prob} it then follows that
\begin{equation}
    \Prob{\int_{\T^d}\abs{U^0_n(t,x)-u(t,x)}^2d\mu(x) \leq \frac{4\cL^2 (T+1)^2\exp(2LT)(1+2LT)^{2n}}{m^{n}\exp(-m)}} > 0. 
\end{equation}
As a result, there exists $\overline{\omega}=\overline{\omega}(t,n,m)\in \Omega$ and a realization $U^0_n(\overline{\omega})$ such that
\begin{equation}\label{eq:true-to-mlp}
    \norm{U^0_n(\overline{\omega})(t,\cdot)-u(t,\cdot)}_{L^2(\mu)} \leq \frac{\cL (T+1)\exp(LT)(1+2LT)^n}{m^{n/2}\exp(-m/2)}. 
\end{equation}
We define 
\begin{equation}\label{eq:omega}
    \omega : [0,T]\times \N^2\to \Omega: (t,n,m)\mapsto \overline{\omega}(t,n,m)
\end{equation}
and set for every $1\leq k\leq n$,
\begin{equation}\label{eq:def-nn}
    \Uhat_{k,\omega}^\theta(t,x) = \Uhat_k^\theta(\omega(t,n,m))(t,x)\quad \text{and} \quad \mlp_{k,\omega}^\theta(t,x) = \mlp_k^\theta(\omega(t,n,m))(t,x)
\end{equation}
for all $k\in\N_0$ and all $\theta\in\Theta$. We then define our approximation as $\Uhat_{n,\omega}^0(t,\cdot)$.

\textbf{Step 2: error estimate. }
We will quantify how well $\Uhat_{n,\omega}^0$ approximates $\mlp_{n,\omega}^0$. Using the calculation that for $f_1,f_2\in  C^1(\R)$ and $h_1, h_2\in L^2(\mu)$ it holds that
\begin{equation}
\begin{split}
\norm{f_1\circ h_1 - f_2\circ h_2}_{L^2(\mu)} &\leq \norm{f_1\circ h_1 - f_2\circ h_1}_{L^2(\mu)} + \norm{f_2\circ h_1 - f_2\circ h_2}_{L^2(\mu)}\\
&\leq \norm{f_1-f_2}_{C^0(\R)} + \abs{f_2}_{Lip(\R)}\norm{h_1-h_2}_{L^2(\mu)}, 
\end{split}
\end{equation}
and the fact that $2n\leq 2^n$ for $n\in\N$ we find that it holds for every $\theta\in\Theta$ that,
\begin{equation}\label{eq:uhat-un-calc}
\begin{split}
    &\norm{\Uhat_{n,\omega}^\theta(t,\cdot)-U^\theta_{n,\omega}(t,\cdot)}_{L^2(\mu)} \\ & \quad \leq \1_\N(n) \left(\norm{\ghat-g}_{L^2(\mu)}+(n-1)\norm{\cI-\Id}_{C^0}\right)
    + T \sum_{i=0}^{n-1} \norm{\cI^{n-i-1}\circ\Fhat\circ \Uhat_{i,\omega}^{(\theta,i,k)}-F\circ U_{i,\omega}^{(\theta,i,k)}}_{L^2(\mu)} \\& \qquad + T \sum_{i=0}^{n-1}\1_\N(i) \norm{\cI^{n-i}\circ\Fhat\circ \Uhat_{i-1,\omega}^{(\theta,-i,k)}-F\circ U_{i-1,\omega}^{(\theta,-i,k)}}_{L^2(\mu)} \\
    & \quad \leq \1_\N(n) \left[\norm{\ghat-g}_{L^2(\mu)} + 2Tn\norm{\Fhat-F}_{C^0(\R)}+ \left((n-1)+\frac{(n-1)n}{2}+\frac{(n+1)n}{2}\right)\norm{\cI-\Id}_{C^0}\right]\\
    &\qquad + LT \sum_{i=0}^{n-1}\left(\norm{\Uhat_i^{(\theta,i,k)}- U_i^{(\theta,i,k)}}_{L^2(\mu)} + \1_\N(i)\norm{ \Uhat_{i-1}^{(\theta,-i,k)}- U_{i-1}^{(\theta,-i,k)}}_{L^2(\mu)} \right)\\
    & \quad \leq \1_\N(n)2^n \left[\norm{\ghat-g}_{L^2(\mu)} + T\norm{\Fhat-F}_{C^0(\R)}+ n\norm{\cI-\Id}_{C^0}\right]\\
    &\qquad + \sum_{i=0}^{n-1} (\max\{1,LT\})^{n-i} \left(\norm{\Uhat_i^{(\theta,i,k)}- U_i^{(\theta,i,k)}}_{L^2(\mu)} + \1_\N(i)\norm{ \Uhat_{i-1}^{(\theta,-i,k)}- U_{i-1}^{(\theta,-i,k)}}_{L^2(\mu)} \right).
\end{split}, 
\end{equation}

Now let us set for every $k\in\N_0$, 
\begin{equation}
    x_k = \sup_{\theta\in\Theta} \norm{\Uhat_{k,\omega}^\theta(t,\cdot)-U^\theta_{k,\omega}(t,\cdot)}_{L^2(\mu)},
\end{equation}
and in addition we define $\alpha_0 = \norm{\ghat-g}_{L^2(\mu)} +T\norm{\Fhat-F}_{C^0(\R)}$, $\alpha_1 =  \norm{\cI-\Id}_{C^0}$ and $\beta = 2+LT$.  
Taking the supremum over all $\theta\in\Theta$ in \eqref{eq:uhat-un-calc} gives us for all $k\in\N_0$ that,
\begin{equation}
\begin{split}
    x_k 
    &\leq \1_\N(k)(\alpha_0+\alpha_1k)\beta^k + \sum_{i=0}^{k-1} \beta^{k-i} (x_{i}+x_{\max\{i-1,0\}}). 
\end{split}
\end{equation}
Therefore, we can use Lemma \ref{lem:recursion1} with $\gamma\leftarrow 0$ then gives us that for all $k\in\N_0$ it holds that,
\begin{equation}\label{eq:mlp-to-nn}
\begin{split}
    &\sup_{\theta\in\Theta} \norm{\Uhat_{k,\omega}^\theta(t,\cdot)-U^\theta_{k,\omega}(t,\cdot)}_{L^2(\mu)}\\ &\qquad\leq \1_\N(k) \frac{(1+\sqrt{2})^k}{2} \left(\norm{\ghat-g}_{L^2(\mu)}+T\norm{\Fhat-F}_{C^0(\R)}+ \norm{\cI-\Id}_{C^0}\right)(2+LT)^k.
\end{split}
\end{equation}
Next we define
\begin{equation}\label{eq:C1-C2}
    C_1 = (T+1)(1+\cL \exp(LT)), \qquad C_2 = 5+3LT.
\end{equation}
Combining \eqref{eq:true-to-mlp} with \eqref{eq:mlp-to-nn} then gives us that, 
\begin{equation}\label{eq:nn-error-bound}
\begin{split}
     &\norm{\Uhat^0_{n,\omega}(t,\cdot)-u(t,\cdot)}_{L^2(\mu)} \\
     &\quad \leq \norm{\Uhat^0_{n,\omega}(t,\cdot)-\mlp^0_{n,\omega}(t,\cdot)}_{L^2(\mu)}+\norm{\mlp^0_{n,\omega}(t,\cdot)-u(t,\cdot)}_{L^2(\mu)}\\
     &\quad \leq C_1 C_2^n \left( \norm{\ghat-g}_{L^2(\mu)}+\norm{\Fhat-F}_{C^0(\R)} + \norm{\cI-\Id}_{C^0}+ m^{-n/2}\exp(m/2)\right). 
\end{split}
\end{equation}
For an arbitrary $\sigma>0$, we choose
\begin{equation}\label{eq:mn-choice}
    m = C_2^{2(1+1/\sigma)}, \qquad n = \sigma\log_{C_2} (4C_1\exp(m/2)/\epsilon)
\end{equation}
and if we choose $\ghat=\ghat_\delta$ and $\Fhat = \Fhat_\delta$ such that,
\begin{equation}\label{eq:delta-eta}
\begin{split}
    \norm{\ghat-g}_{L^2(\mu)} \leq \delta = \frac{\epsilon}{4C_1C_2^n} = \frac{\epsilon^{1+\sigma}}{(4C_1)^{1+\sigma}\exp(\sigma m/2)}, 
\end{split}
\end{equation}
then we obtain that
\begin{equation}
    \norm{\Uhat^0_{n,\omega}(t,\cdot)-u(t,\cdot)}_{L^2(\mu)} \leq \epsilon. 
\end{equation}

\textbf{Step 3: size estimate. }
We now provide estimates on the size of the network constructed in Step 1. 
First of all, it is straightforward to see that the depth of the network can be bounded by
\begin{equation}\label{eq:L-nn}
     \cL_\epsilon(\Uhat^0_{n,\omega}) \leq \cL_\delta(\ghat)+(n-1)\cL_\delta(\Fhat) \leq \cL_{\delta}(\ghat)+\log_{C_2} (3C_1\exp(m/2)/\epsilon)\cL_\delta(\Fhat). 
\end{equation}
Next we prove an estimate on the number of needed neurons. For notation, we write $\cM_{n} = \cM_\epsilon(\Uhat^0_{n,\omega})$. We find that for all $0\leq k\leq n$,
\begin{equation}\label{eq:mlp-cost-recursion}
\begin{split}
    \cM_k &\leq \1_\N(k) m^k (\cM_\delta(\ghat)+(k-1)\cM_\delta(\cI)) \\
    &\quad+ \sum_{i=0}^{k-1} m^{k-i} (2\cM_\delta(\Fhat) + (2k-2i-1)\cM_\delta(\cI)+ \cM_i + \cM_{\max\{i-1,0\}})\\
    &\leq \1_\N(k)(\cM_\delta(\ghat)+ \cM_\delta(\Fhat)+k\cM_\delta(\cI))(2m)^k+ \sum_{i=0}^{k-1} m^{k-i} (\cM_i + \cM_{\max\{i-1,0\}}).
\end{split}
\end{equation}
Applying Lemma \ref{lem:recursion1} to \eqref{eq:mlp-cost-recursion} (i.e. $\alpha_0\leftarrow \cM_\delta(\ghat)+\cM_\delta(\Fhat)$, $\alpha_1\leftarrow \cM_\delta(\cI)$ and $\beta\leftarrow 2m$) then gives us that
\begin{equation}\label{eq:M-nn}
    \cM_n \leq \frac{1}{2}(\cM_\delta(\ghat)+ \cM_\delta(\Fhat)+ \cM_\delta(\cI)) (1+ \sqrt{2} )^n (2m)^n. 
\end{equation}
Observing that $2+2\sqrt{2} \leq C_2$ and recalling that $m=C_2^{2(1+1/\sigma)}$ we find that
\begin{equation}
\begin{split}
    \cM_n &\leq \frac{1}{2}(\cM_\delta(\ghat)+ \cM_\delta(\Fhat)+ \cM_\delta(\cI)) C_2^{(3\sigma+2)n/\sigma}\\
    &= \frac{1}{2}(\cM_\delta(\ghat)+ \cM_\delta(\Fhat)+ \cM_\delta(\cI)) \left(\frac{4C_1\exp(m/2)}{\epsilon}\right)^{2+3\sigma}. 
\end{split}
\end{equation}
For the width, we make the estimate $\width_\epsilon(\Uhat^0_{n,\omega}) \leq \cM_n$. 

\end{proof}


\subsection{PINN approximation of nonlinear parabolic equations}

\begin{setting}\label{set:pinn}
Assume Setting \ref{set:allen-cahn}, let $\ghat \in C(\T^d,\R)\cap L^2(\mu)$\footnote{The function $\ghat$ can but need not be the same as the function $\ghat_\epsilon$, for some $\epsilon>0$, of Setting \ref{set:allen-cahn}. } and let $\omega:[0,T]\times\N^2\to\Omega$ be defined as in \eqref{eq:omega} in the proof of Theorem \ref{thm:nn}. Let $\Uhat_{n,\omega}^\theta \colon [0,T] \times \T^d\times \Omega\to \R$, 
$n\in\Z$, 
$\theta\in\Theta$, satisfy for all 
$n\in\N_0$, 
$\epsilon>0$,
$\theta\in\Theta$, $t\in[0,T]$, $x\in\T^d$ that
\begin{equation}\label{eq:nn-pido}
\begin{split}
\Uhat_{n,\omega}^\theta(t,x) 
& = \frac{\1_\N(n)}{m^n} \br[\Bigg]{ \sum_{k=1}^{m^n} (\cI_\epsilon^{n-1}\circ\ghat)\pr[\bigl]{x + W_{T-t}^{(\theta,0,-k)}(\omega(t,n,m))} } \\
& \quad + \sum_{i=0}^{n-1} \frac{(T-t)}{m^{n-i}}\br[\Bigg]{ \sum_{k=1}^{m^{n-i}} \pr[\Big]{ (\cI_\epsilon^{n-i-1}\circ\Fhat_\epsilon)(\Uhat_{i,\omega}^{(\theta,i,k)}) \\
&\quad - \1_\N(i) (\cI_\epsilon^{n-i}\circ\Fhat_\epsilon)( \Uhat_{i-1,\omega}^{(\theta,-i,k)}) } \pr[\big]{ \EuScript{U}_t^{(\theta,i,k)}(\omega(t,n,m)), x + W_{\EuScript{U}_t^{(\theta,i,k)}-t}^{(\theta,i,k)} (\omega(t,n,m))} }.
\end{split}
\end{equation}
\end{setting}

\begin{lemma}\label{lem:space-der-nn}
Assume Setting \ref{set:pinn}. Under the assumption that,
\begin{equation}
\max_{1\leq j\leq k}\norm{\cI^j}_{C^k([-\cL-1, \cL+1])}\leq 2, 
\end{equation}
where $\cI^j$ denotes $j$ compositions of $\cI$, it holds for all $\ell, k\in\N_0$ that,
\begin{equation}\label{eq:pinn-x-claim}
\begin{split}
     &\sup_{\theta\in\Theta} \norm{ \Uhat^\theta_{k,\omega}}_{C^{(0,\ell)}([0,T]\times\T^d)} \leq C_{k,\ell} := \left[\norm{\ghat}_{C^\ell(\T^d)}+2B_\ell (1+\sqrt{2})^k (1+2B_\ell T\norm{\Fhat}_{C^\ell(\R)}^\ell)^k\right]^{2(\ell+1)!},
\end{split}
\end{equation}
and where $B_\ell$ denote the $\ell$-th Bell number i.e., the number of possible partitions of a set with $\ell$ elements. 
\end{lemma}
\begin{proof}
We prove the claim by induction on $\ell$. 

\textit{Base case.} From Definition \ref{def:MLP}, we find that for $\ell = 0$ and all $k\in\N_0$ it holds that
\begin{equation}\label{eq:base-case}
    \sup_{\theta\in\Theta} \norm{\Uhat^\theta_{k,\omega}}_{C^{0}([0,T]\times\T^d)} \leq \norm{\ghat}_{C^{0}(\T^d)} + 2T\norm{\Fhat}_{C^{0}(\R)}k. 
\end{equation}
Claim \eqref{eq:pinn-x-claim} follows immediately for $\ell=0$. 

\textit{Induction step.} We assume that claim \eqref{eq:pinn-x-claim} holds true for all $0\leq \ell^*\leq \ell-1$ and $k\in\N_0$. From this assumption, we will deduce that \eqref{eq:pinn-x-claim} holds true for $\ell$ and all $k\in\N_0$. We first observe that it follows from Lemma \ref{lem:faa-di-bruno}, the induction hypothesis and the fact that $(C_{k,\ell})_{\ell\geq0}$ is non-decreasing for any $k$, that for all $\theta\in\Theta$ and $0\leq i,j\leq k$ it holds that,
\begin{equation}\label{eq:composition-chain-rule}
\begin{split}
    \abs{ (\cI^j\circ\Fhat)(\Uhat^{\theta}_{i,\omega})}_{C^\ell([0,T]\times\T^d)} \leq \norm{\cI^j\circ\Fhat}_{C^{\ell}(\R)}\left(B_\ell C_{k,\ell-1}^\ell+ \abs{ \Uhat^{\theta}_{i,\omega}}_{C^{(0,\ell)}([0,T]\times\T^d)}\right),
\end{split}
\end{equation}
and where (again using Lemma \ref{lem:faa-di-bruno}) it holds that $\norm{\cI^j\circ\Fhat}_{C^{\ell}(\R)} \leq 2B_\ell \norm{\Fhat}_{C^{\ell}}^\ell$. 

Using this estimate and the fact that $(C_{k,\ell})_{k\geq0}$ is non-decreasing for any $\ell$, we can make the following calculation for every $k\in\N_0$,
\begin{equation}
    \begin{split}
        &\sup_{\theta\in\Theta} \abs{ \Uhat^\theta_{k,\omega}}_{C^{(0,\ell)}([0,T]\times\T^d)} \\&\leq \1_{\N}(k)\abs{\ghat}_{C^\ell(\T^d)} + T \sum_{i=0}^{k-1} \sup_{\theta\in\Theta}\abs{ (\cI^{k-i-1}\circ\Fhat)(\Uhat^{\theta}_{i,\omega})}_{C^{(0,\ell)}(([0,T]\times\T^d))}\\&\quad +T \sum_{i=0}^{k-1}\1_{\N}(i)\sup_{\theta\in\Theta}\abs{ (\cI^{k-i}\circ\Fhat)(\Uhat^{\theta}_{i-1,\omega})}_{C^{(0,\ell)}(([0,T]\times\T^d))}\\
        &\leq \1_{\N}(k)\abs{\ghat}_{C^\ell(\T^d)} + 2B_\ell T\norm{\Fhat}_{C^\ell(\R)}^\ell \sum_{i=0}^{k-1} \left(B_\ell C_{k,\ell-1}^\ell+\sup_{\theta\in\Theta}\abs{ \Uhat^{\theta}_{i,\omega}}_{C^{(0,\ell)}(([0,T]\times\T^d))}\right)\\
        &\quad +2B_\ell T\norm{\Fhat}_{C^\ell(\R)}^\ell \sum_{i=0}^{k-1} \1_{\N}(i)\left(B_\ell C_{k,\ell-1}^\ell+\sup_{\theta\in\Theta}\abs{ \Uhat^{\theta}_{i-1,\omega}}_{C^{(0,\ell)}(([0,T]\times\T^d))}\right)\\
        &\leq \1_{\N}(k)(\abs{\ghat}_{C^\ell(\T^d)} + 2B_\ell T \norm{\Fhat}_{C^\ell(\R)}^\ell C_{k,\ell-1}^\ell k)\\
        &\quad + \sum_{i=0}^{k-1}2B_\ell T\norm{\Fhat}_{C^\ell(\R)}^\ell\left(\sup_{\theta\in\Theta}\abs{ \Uhat^{\theta}_{i,\omega}}_{C^{(0,\ell)}(([0,T]\times\T^d))} + \1_{\N}(i)\sup_{\theta\in\Theta}\abs{ \Uhat^{\theta}_{i-1,\omega}}_{C^{(0,\ell)}(([0,T]\times\T^d))}\right)
    \end{split}
\end{equation}

Application of Lemma \ref{lem:recursion1} with $\alpha_0\leftarrow \abs{\ghat}_{C^\ell(\T^d)}$, $\alpha_1 \leftarrow  2B_\ell C_{k,\ell-1}^\ell $, $\beta \leftarrow (1+2B_\ell T\norm{\Fhat}_{C^\ell(\R)}^\ell)$ and $\gamma \leftarrow 0$ gives us
\begin{equation}
\begin{split}
    \sup_{\theta\in\Theta} \norm{ \Uhat^\theta_{k,\omega}}_{C^{0}([0,T]\times\T^d)} &\leq \frac{\abs{\ghat}_{C^\ell(\T^d)} + 2B_\ell C_{k,\ell-1}^\ell}{2}(1+\sqrt{2})^k (1+2B_\ell T\norm{\Fhat}_{C^\ell(\R)}^\ell)^k \\
    &\leq \left[\abs{\ghat}_{C^\ell(\T^d)}+2B_\ell (1+\sqrt{2})^k (1+2B_\ell T\norm{\Fhat}_{C^\ell(\R)}^\ell)^k\right]C_{k,\ell-1}^\ell. 
\end{split}
\end{equation}
Filling in the definition of $C_{k,\ell-1}$ indeed gives us the formula as stated in \eqref{eq:pinn-x-claim}, thereby concluding the proof of the claim. 
\end{proof}

\begin{lemma}\label{lem:ass-check-ac}
Let $F$ be a polynomial. For every $\sigma, \epsilon>0$ there is an operator $\cU^\epsilon$ as in Assumption \ref{ass:nn} such that for every $t\in [0,T]$,
\begin{equation}
    \norm{\cU^\epsilon(u_0,t)-\cG(v)(u_0,t)}_{L^2(\T^d)} \leq \epsilon, \qquad C_{\epsilon,\ell}^B \leq C(B\epsilon^{-\sigma})^{2(l+1)!}, \qquad C^\epsilon_{\mathrm{stab}} \leq C\epsilon^{-\sigma}. 
\end{equation}
Moreover it holds that $\depth(\cU^\epsilon(u_0,t)) \leq \depth(\uhat_0)+C\ln(\epsilon^{-1})$, $\width(\cU^\epsilon(u_0,t)) \leq \width(\uhat_0)\epsilon^{-2-\sigma}$ and $\size(\cU^\epsilon(u_0,t)) \leq \size(\uhat_0)\epsilon^{-2-\sigma}$. 
\end{lemma}
\begin{proof}
The three bounds are a consequence of, respectively, Theorem \ref{thm:nn} and Lemma \ref{lem:space-der-nn} and \eqref{eq:nn-error-bound}. The size estimates follow from Theorem \ref{thm:nn}. Note that one might have to rescale the constant $\sigma>0$.
\end{proof}

\section{Additional material for Section \ref{sec:PIDO}}\label{app:PIDO}

\subsection{Errors of DeepONets}\label{app:error-deeponet}

In \cite{LMK1}, numerous error estimates for DeepONets are proven, with a focus on DeepONets that use the ReLU activation function. In order to quantify this error, the authors fix a probability measure $\mu \in \cP(\inspace)$ and define the error as,
\begin{align} \label{eq:approxerr}
\Err
=
\left(
\int\limits_{\inspace} \int\limits_{U}
\left|
\G(u)(y) 
-
\cG_\theta(u)(y)
\right|^2 
\, dy
\, d\mu(u)
\right)^{1/2},
\end{align}
assuming that there exist embeddings $\inspace \embeds L^2(D)$ and  $\outspace\embeds L^2(U)$. 
From \cite[Lemma 3.4]{LMK1}, it then follows that $\Err$ \eqref{eq:approxerr} can be bounded as,
\begin{gather} \label{eq:error}
\begin{aligned}
\Err 
&\le \Lip_\alpha(\G) \Lip(\cR \circ \cP) \, (\Err_{\cE})^\alpha + \Lip(\cR) \Err_{\cA} + \Err_{\cR}, 
\end{aligned}
\end{gather}
where $\Lip_\alpha(\cdot)$ denotes the $\alpha$-Hölder coefficient of an operator and where $\Err_\cE$ quantifies the encoding error, where $\Err_\cA$ is the error incurred in approximating the approximator $\cA$ and where $\Err_\cR$ quantifies the reconstruction error. Assuming that all Hölder coefficients are finite, one can prove that $\Err$ is small if $\Err_\cE$, $\Err_\cA$ and $\Err_\cR$ are all small. 
We summarize how each of these three errors can be bounded using the results from \cite{LMK1}. 

\begin{itemize}
    \item The upper bound on the encoding error $\erre$ depends on the chosen sensors and the spectral decay rate for the covariance operator associated to the measure $\mu$. Use bespoke sensor points to obtain optimals bounds when possible, otherwise use random sensors to obtain almost optimal bounds. More information can be found in \cite[Section 3.5]{LMK1}. 
    \item The upper bound on the reconstruction error $\errr$ depends on the smoothness of the operator and the chosen basis functions $\trunk$ i.e., neural networks, for the reconstruction operator $\cR$. Following \cite[Section 3.4]{LMK1}, one first chooses a standard basis $\Tilde{\trunk}$ of which the properties are well-known. We denote the corresponding reconstruction by $\Tilde{\cR}$ and the corresponding reconstruction error by $\Err_{\Tilde{\cR}}$. In this work, we focus on Fourier and Legendre basis function, both of which are introduced in \textbf{SM} \ref{app:preliminaries}. One then proceeds by constructing the neural network basis $\trunk$ i.e., the trunk nets, that satisfy for some $\epsilon>0$ and $p\geq 1$ the condition
    \begin{align} \label{eq:tdiff-ass}
\max_{k=1,\dots, p} \Vert \tr_k - \tilde{\tr}_k \Vert_{L^2}
\le 
\frac{\epsilon}{p^{3/2}},
\end{align}
which is shown to imply that, 
\begin{align} \label{eq:err-rec-comp}
\Err_{\cR}
\le
\Err_{\tR}
+ 
C \epsilon,
\end{align}
where $C\ge 1$ depends only on $\int_{L^2} \Vert u \Vert^2 \, d\G_\#\mu(u)$. Using standard approximation theory, one can calculate an upper bound on $\Err_{\Tilde{\cR}}$ and using neural network theory one can quantify the network size of $\trunk$ needed such that \eqref{eq:tdiff-ass} is satisfied. For the Fourier and Legendre bases such results are presented in Lemma \ref{lem:rec-fourier} and Lemma \ref{lem:rec-legendre}, respectively. 

\item The upper bound on the approximation error $\erra$ depends on the regularity of the operator $\G$. We present the tanh counterparts of some results of \cite[Section 3.6]{LMK1} in the following sections, with the main result being Theorem \ref{thm:holomorphic-approx}.
\end{itemize}

For bounded linear operators, these calculations are rather straightforward and are presented in \cite[\textbf{SM} D]{LMK1}. For nonlinear operators, one has to complete all the above steps for each specific case. In \cite[Section 4]{LMK1}, this has been done for four types of differential equations. 

\subsection{Auxiliary results for linear operators}

Following Section \ref{app:error-deeponet}, we need results on the required neural network size to approximate the reconstruction basis to a certain accuracy \eqref{eq:tdiff-ass}. The following lemma provides such a result for the Fourier basis introduced in \textbf{SM} \ref{app:Fourier}. 

\begin{lemma}\label{lem:rec-fourier}
Let $s, d, p\in \N$. For any $\epsilon>0$, there exists a trunk net $\trunk: \R^d \to \R^p$ with 2 hidden layers of width $\bigO(p^{\frac{d+1}{d}}+ps\ln(ps\epsilon^{-1}))$ and such that 
\begin{align} \label{eq:Fourier-NN}
p^{3/2} \max_{j=1,\dots, p} \Vert \tr_j - \fb_j \Vert_{C^s([0,2\pi]^d)}
\le \epsilon,
\end{align}
where $\fb_1,\dots, \fb_p$ denote the first $p$ elements of the Fourier basis, as in \textbf{SM} \ref{app:Fourier}.

\end{lemma}
\begin{proof}
We note that each element in the (real) trigonometric basis $\fb_1,\dots, \fb_p$ can be expressed in the form 
\begin{equation}
\fb_j({x}) 
=
\cos(\kappa\cdot {x}),
\quad
\text{or}
\quad
\fb_j({x}) = \sin(\kappa\cdot {x}),
\end{equation}
for $\kappa = \kappa(j)\in \Z^d$ with $|\kappa|_\infty \le N$, where $N$ is chosen as the smallest natural number such that $p \le (2N+1)^d$. We focus only focus on the first form, as the proof for the second form is entirely similar. Define $f:[0,2\pi]^d\to\R:x\mapsto 
\kappa\cdot x$ and $g:[-2\pi dN, 2\pi dN]\to\R:x\mapsto \cos(x)$. As $f([0,2\pi]^d) \subset [-2\pi dN, 2\pi dN]$, the composition $g\circ f$ is well-defined and one can see that it coincides with a trigonometric basis function $\fb_j$. 
Moreover, the linear map $f$ is a trivial neural network without hidden layers. Approximating $\fb_j$ by a neural network $\tau_j$ therefore boils down to approximating $g$ by a suitable neural network. 

From \cite[Theorem 5.1]{deryck2021approximation} it follows that the function $g$ there exists an independent constant $R>0$ such that for large enough $t\in\N$ there is a tanh neural network $\hg_t$ with two hidden layers and $\bigO(t+N)$ neurons such that
\begin{equation}
    \norm{g-\hg_t}_{C^s([-2\pi dN, 2\pi dN])} \leq 4 (8(s+1)^3R)^{s} \exp(t-s). 
\end{equation}
This can be proven from \cite[eq. (74)]{deryck2021approximation} by setting $\delta\leftarrow\frac{1}{3}$, $k\leftarrow s$, $s\leftarrow t$, $N\leftarrow 2$ and using $\norm{g}_{C^s}=1$ and Stirling's approximation to obtain
\begin{equation}
    \frac{1}{(t-s)!}\left(\frac{3}{2\cdot 2}\right)^{t-s} \leq \frac{1}{\sqrt{2\pi(t-s)}} \left(\frac{e}{t-s}\right)^{t-s} \leq \exp(s-t) \quad \text{for } t>s+e^2. 
\end{equation}
Setting $t = \bigO(\ln(\delta^{-1})+s\ln(s))$ then gives a neural network $\hg_t$ with $\norm{g-\hg_t}_{C^s}<\eta$. Next, it follows from \cite[Lemma A.7]{deryck2021approximation} that
\begin{align}
\begin{split}
    \norm{g\circ f-\hg_t\circ f}_{C^s([0,2\pi]^d)} &\leq 16(e^2s^4d^2)^s \norm{g-\hg_t}_{C^s([-2\pi dN, 2\pi dN])} \norm{f}_{C^s([0,2\pi]^d)}^s\\
    &\leq 16(e^2s^4d^2)^s \eta (2\pi dN)^s. 
\end{split}
\end{align}
From this follows that we can obtain the desired accuracy \eqref{eq:Fourier-NN} if we set $\tau_j = \hg_{t(\eta)}\circ f$ with
\begin{equation}
    \eta = \frac{\epsilon p^{-3/2}}{16(2\pi Nd^3e^2s^4)^{s}},
\end{equation}
which amounts to $t = \bigO(s\ln(sN\epsilon^{-1}))$. As a consequence, the tanh neural network $\tau_j$ has two hidden layers with $\bigO(s\ln(sN\epsilon^{-1})+N)$ neurons and therefore, by recalling that $p\sim N^d$, the combined network $\bm{\tau}$ has two hidden layers with
\begin{equation}
    \bigO(p(s\ln(sN\epsilon^{-1})+N)) = \bigO(ps\ln(ps\epsilon^{-1})+p^{\frac{d+1}{d}})
\end{equation}
neurons. 
\end{proof}

\subsection{Proof of Theorem \ref{thm:linear-pido}}\label{app:linear-pido}
\begin{proof}
Consider the setting of Theorem \ref{thm:linear-pido}. Using \cite[Theorem D.3]{LMK1}, the reasoning as in \cite[Example D.4]{LMK1} and Lemma \ref{lem:rec-fourier} we find that there exists a constant $C = C(d,\ell)>0$, such that for any $m, p, s \in \N$ there exists a DeepONet with trunk net $\trunk$ and branch net $\branch$, such that 
\begin{gather}
\size(\trunk) \le C(p^{\frac{d+1}{d}}+ps\ln(ps\epsilon^{-1})),
\quad
\depth(\trunk) = 3,
\end{gather}
and where 
\begin{gather}
\size(\branch) \le p, 
\quad
\depth(\branch) \le 1,
\end{gather}
and such that the DeepONet approximation error \eqref{eq:approxerr} is bounded by 
\begin{equation}
\norm{\cG(v)-\cG_\theta(v)}_{L^2(\mu\times\lambda)}
\le \epsilon + 
C\exp\left(-c\, p^{1/d}\right) + 
C \exp\left(-\frac{c\,m^{1/d}}{\log(m)^{1/d}}\right). 
\end{equation}
Moreover, it holds that
\begin{equation}
     \abs{\mathcal{N}(u)(\cdot)}_{C^s} \leq C p^{s/d}, 
\end{equation}
since in this case $\trunk$ approximates the Fourier basis (\textbf{SM} \ref{app:Fourier}). From \eqref{eq:bound-fourier-basis}, one can then deduce the estimate on the $C^s$-norm of the DeepONet. This proves that \eqref{eq:growth} in Theorem \ref{thm:deeponet-to-pido} holds with $\sigma(s)=s/d$. This concludes the proof. 
\end{proof}

\subsection{Auxiliary results for nonlinear operators}
We provide a neural network approximation result for the Legendre basis from \textbf{SM} \ref{app:Legendre}.

\begin{lemma}\label{lem:rec-legendre}
Let $n, p\in \N$. For any $\epsilon>0$, there exists a trunk net $\trunk: \R^d \to \R^p$ with two hidden layers of width $\bigO(p)$ such that 
\begin{align} \label{eq:Legendre-NN}
p^{3/2} \max_{j=1,\dots, p} \Vert \tr_j - L_j \Vert_{C^s([-1,1]^d)}
\le \epsilon,
\end{align}
where $L_1,\dots, L_p$ denote the first $p$ elements of the Legendre basis, as in \textbf{SM} \ref{app:Legendre}. 
\end{lemma}

\begin{proof}
Let $j\in{1,\ldots,p}$. It holds by definition of Legendre polynomials and the corresponding enumeration (\textbf{SM} \ref{app:Legendre}) that the degree in every variable is at most $p$. Therefore, $L_j$ is a product of $d$ univariate polynomials of degree at most $p$. 
From \cite[Lemma 3.2]{deryck2021approximation} it follows that one needs a shallow tanh neural network with $O(p)$ neurons to approximate a univariate polynomial to any accuracy. The result from \cite[Corollary 3.7]{deryck2021approximation} can be used to construct a shallow tanh network that approximates the product of the $d$ univariate polynomials. Note that its size only depends on the dimension $d$ and not on the polynomial degree $p$ or the accuracy. 
Finally, \cite[Lemma A.7]{deryck2021approximation} ensures the accuracy of the composition of the two subnetworks. It then follows that there exist a tanh neural network of width $\bigO(p)$ and two hidden layers that achieves the wanted error estimate. 
\end{proof}

In our proofs, we require tanh counterparts to the results for DeepONets with ReLU activation function from \cite{LMK1}. We present these adapted results below for completeness. 

The first lemma considers the neural network approximation of the map $\bm{u}\mapsto \hat{Y}(\bm{u})$, as defined in \cite[Eq. (3.59)]{LMK1}.

\begin{lemma} \label{lem:Yhat-NN}
Let $N,d \in \N$, and denote $m := (2N+1)^d$. There exists a constant $C>0$, independent of $N$, such that for every $N$ there exists a tanh neural network $\Psi: \R^m \to \R^m$, with 
\begin{equation}
   \size(\Psi) \le C(1+m\log(m)), 
\quad
\depth(\Psi) \le C(1+\log(m)), 
\end{equation}
and such that $\Psi(\bm{u}) = (\hat{Y}_{1}(\bm{u}),\dots, \hat{Y}_{m}(\bm{u}))$, for all $\bm{u}\in \R^m$.
\end{lemma}
\begin{proof}
The proof is identical to that of \cite[Lemma 3.28]{LMK1}.
\end{proof}

We can now state the following result  \cite[Theorem 3.10]{SZ1} which is the counterpart of \cite[Theorem 3.32]{LMK1} for tanh neural networks.

\begin{theorem} \label{thm:be-expansion}
Let $V$ be a Banach space and let $\cJ$ be a  countable index set. Let $\cF: [-1,1]^\cJ \to V$ be a $(\bm{b},\epsilon,\kappa)$-holomorphic map for some $\bm{b}\in \ell^q(\N)$ and $q\in (0,1)$, and an enumeration $\kappa: \N \to \cJ$. Then there exists a constant $C>0$, such that for every $N\in \N$, there exists an index set 
\begin{equation}
\Lambda_N
\subset
\set{
\bm{\nu} = (\nu_1,\nu_2,\dots) \in \textstyle{\prod}_{j\in \cJ} \N_0
\:\vert\:
\nu_j\ne 0 \; \text{for finitely many $j\in \cJ$}
},    
\end{equation}
with $|\Lambda_N|=N$, a finite set of coefficients $\{c_{\bm{\nu}}\}_{\bm{\nu}\in \Lambda_N} \subset V$, and a tanh network $\Psi: \R^N \to \R^{\Lambda_N}$, $y\mapsto \{\Psi_{\bm{\nu}}(y)\}_{\bm{\nu}\in \Lambda_N}$ with
\begin{gather}
\size(\Psi) \le C(1+N\log(N) ), 
\quad
\depth(\Psi) \le C(1+\log\log(N)),
\end{gather}
and such that
\begin{align} \label{eq:be-expansion}
\sup_{\bm{y}\in [-1,1]^\cJ} 
\left\Vert
\cF(\bm{y}) - \sum_{\bm{\nu}\in \Lambda_N} c_{\bm{\nu}} \Psi_{\bm{\nu}}(y_{\kappa(1)},\dots, y_{\kappa(N)})
\right\Vert_{V} 
\le C N^{1-1/q}.
\end{align}
\end{theorem}

Using this theorem, we can state the tanh counterpart to \cite[Corollary 3.33]{LMK1}. 

\begin{corollary} \label{cor:holomorphic-NN}
Let $V$ be a Banach space. Let $\cF: [-1,1]^\cJ \to V$ be a $(\bm{b},\epsilon,\kappa)$-holomorphic map for some $\bm{b}\in \ell^q(\N)$ and $q\in (0,1)$, where $\kappa: \N\to \cJ$ is an enumeration of $\cJ$. In particular, it is assumed that $\{b_{j}\}_{j\in \N}$ is a monotonically decreasing sequence. If $\cP: V \to \R^p$ is a continuous linear mapping, then there exists a constant $C>0$, such that for every $m\in \N$, there exists a tanh network $\Psi: \R^m \to \R^p$, with
\begin{gather}
\size(\Psi) \le C(1+pm\log(m) ), 
\quad
\depth(\Psi) \le C(1+ \log\log(m)),
\end{gather}
and such that
\begin{equation}
\sup_{\bm{y}\in [-1,1]^\cJ} \Vert \cP \circ \cF(\bm{y}) - \Psi(y_{\kappa(1)},\dots, y_{\kappa(m)})\Vert_{\ell^2(\R^p)} \le C \Vert \cP \Vert \,  m^{-s}, 
\end{equation}
where $s := q^{-1} - 1 > 0$ and $\Vert \cP \Vert = \Vert \cP \Vert_{V\to \ell^2}$ denotes the operator norm.
\end{corollary}
\begin{proof}
The proof is identical to the one presented in \cite[Appendix C.18]{LMK1}.
\end{proof}

Finally, we use this result to state the counterpart to \cite[Theorem 3.34]{LMK1}, which considers the approximation of a parametrized version of the operator $\cG$, defined as a mapping
\begin{equation}\label{eq:G-parametrized}
    \cF: [-1,1]^\cJ\to L^2(U): \bm{y} \mapsto \cG(u(\cdot; \bm{y})). 
\end{equation}
A more detailled discussion can be found in \cite[Section 3.6.2]{LMK1}.

\begin{theorem} \label{thm:holomorphic-approx}
Let $\cF: [-1,1]^\cJ \to L^2(U)$ be $(\bm{b},\epsilon,\kappa)$-holomorphic with $\bm{b}\in \ell^q(\N)$ and $\kappa: \N \to \cJ$ an enumeration, and assume that $\cF$ is given by \eqref{eq:G-parametrized}. Assume that the encoder/decoder pair is constructed as in \cite[Section 3.5.3]{LMK1}, so that \cite[Eq. (3.69)]{LMK1} holds. Given an affine reconstruction $\cR: \R^p\to L^2(U)$, let $\cP: L^2(U) \to \R^p$ denote the corresponding optimal linear projection \cite[Eq. (3.17)]{LMK1}. Then given $k\in\N$, there exists a constant $C_k>0$, independent of $m$, $p$ and an approximator $\cA: \R^m \to \R^p$ that can be represented by a neural network with
\[
\begin{gathered}
\size(\cA) 
\le
C_k(1+pm \log(m) ), 
\quad
\depth(\cA) \le C_k(1+\log(m)).
\end{gathered}
\]
and such that the approximation error $\Err_{\cA}$ can be estimated by
\[
\Err_{\cA} 
\le
C_k \Vert \cP \Vert \, m^{-k},
\]
where $\Vert \cP \Vert =  \Vert \cP \Vert_{L^2(U) \to \R^p}$ is the operator norm of $\cP$.
\end{theorem}
\begin{proof}
The proof is as in \cite[Appendix C.19.1]{LMK1}.
\end{proof}

\subsection{Gravity pendulum with external force}\label{app:pendulum}

Next, we consider the following nonlinear ODE system, already considered in the context of approximation by DeepONets in \cite{deeponets} and \cite{LMK1},
\begin{gather} \label{eq:pendulum0}
\left\{
\begin{aligned}
\frac{dv_1}{dt} &= v_2,
\\
\frac{dv_2}{dt} &= -\gamma\sin(v_1) + u(t).
\end{aligned}
\right.
\end{gather}
with  initial condition $v(0) = 0$ and where $\gamma>0$ is a parameter. Let us denote $v = (v_1, v_2)$ and
\begin{equation}
    g(v) := 
\begin{pmatrix}
v_2 \\
-\gamma\sin(v_1)
\end{pmatrix},
\quad
U(t) := 
\begin{pmatrix}
0 \\
u(t)
\end{pmatrix},
\end{equation}
so that equation \eqref{eq:pendulum0} can be written in the form
\begin{align} \label{eq:pendulum}
\cL_u(v) := \frac{dv}{dt} - g(v) + U = 0, \quad v(0) = 0.
\end{align}
In \eqref{eq:pendulum}, $v_1,v_2$ are the angle and angular velocity of the pendulum and the constant $\gamma$
denotes a frequency parameter. The dynamics of the pendulum is driven by an external force $u$. 
With the external force $u$ as the input, the output of the system is the solution vector $v$ and the underlying nonlinear operator is given by $\G: L^2([0,T]) \to L^2([0,T]): u \mapsto \cG(u) = v$. Following the discussion in \cite{LMK1}, we choose an underlying (parametrized) measure $\mu \in \cP(L^2([0,T]))$ as a law of a random field $u$, that can be expanded in the form 
\begin{align} \label{eq:pendulum-law}
u(t; Y) 
=
\sum_{k\in \Z} Y_k \alpha_k \fb_k\left(\frac{2\pi t}{T}\right),
\quad
t \in [0,T],
\end{align}
where $\fb_k(x)$, $k\in \Z$, denotes the one-dimensional standard Fourier basis (\ref{app:Fourier}) and where the coefficients $\alpha_k\ge 0$ decay to zero as
$\alpha_k \le C_\alpha \exp(-|k|\ell)$ for some constants $C_\alpha, \ell>0$. 
Furthermore, we assume that the $\{Y_k\}_{k\in \Z}$ are iid random variables on $[-1,1]$.

Assuming the described setting, the following lemma gives an error bound of tanh DeepONets in terms of the sizes of the corresponding branch and trunk nets. 

\begin{lemma} \label{lem:pendulum}
Consider the DeepONet approximation problem for the gravity pendulum \eqref{eq:pendulum0}, where the forcing $u(t)$ is distributed according to a probability measure $\mu \in \cP(L^2([0,T]))$ given as the law of the random field \eqref{eq:pendulum-law}. For any $k,r\in \N$, there exists a constant $C = C(k,r)>0$, and a constant $c>0$, independent of $m$, $p$, such that for any $m, \, p \in \N$, there exists a DeepONet $\cG_\theta$ with trunk net $\trunk$ and branch net $\branch$, such that 
\begin{gather}
\size(\trunk) \le Cp,
\quad
\depth(\trunk) = 2,
\end{gather}
and
\begin{gather}
\size(\branch) \le C(1+pm \log(m)), 
\quad
\depth(\branch) \le C(1+ \log(m)),
\end{gather}
and such that the DeepONet approximation error \eqref{eq:approxerr} is bounded by 
\begin{equation}
\Err 
\le 
C e^{-c\ell m}
+
C m^{-k} 
+
C p^{-r},
\end{equation}
and that for all $s\in N$,
\begin{equation}
     \abs{\cG_\theta(u)(\cdot)}_{C^s} \leq C p^{d/2+2sd}. 
\end{equation}
\end{lemma}
\begin{proof}
The proof of the statement is identical to that of \cite[Theorem 4.10]{LMK1}, with the only difference that we consider tanh neural networks instead of ReLU neural networks. As a result, the proof comes down to determining the size of the trunk net $\trunk$ using Lemma \ref{lem:rec-legendre} instead of \cite[Proposition 2.10]{opschoor2019exponential}, thereby proving the tanh counterpart of \cite[Proposition 4.5]{LMK1}, and replacing \cite[Proposition 4.9]{LMK1} by Theorem \ref{thm:holomorphic-approx}. The $C^s$-bound of the DeepONet follows from the $C^s$-bound of Legendre polynomials \eqref{eq:legendre-cs-bound} and Lemma \ref{lem:rec-legendre}.
\end{proof}

We can again follow Theorem \ref{thm:deeponet-to-pido} to obtain error bounds for physics-informed DeepONets. Assumption \ref{ass:partition} is satisfied for $[0,T]$. As a result, we can apply Theorem \ref{thm:deeponet-to-pido} to obtain the following result. 

\begin{theorem}
Consider the setting of Lemma \ref{lem:pendulum}. For every $\beta>0$, there exists a constant $C>0$ such that for any $p\in N$, there exists a DeepONet $\cG_\theta$ with a trunk net $\trunk = (0,\tr_1, \dots, \tr_p)$ with $p$ outputs and branch net $\branch = (0,\beta_1,\dots, \beta_p)$, such that 
\begin{gather}
\size(\trunk) \le Cp,
\quad
\depth(\trunk) = 2,
\end{gather}
and
\begin{gather}
\size(\branch) \le C(1+p^2 \log(p)), 
\quad
\depth(\branch) \le C(1+ \log(p)),
\end{gather}
and such that
\begin{equation}
    \norm{\frac{d\cG_\theta(u)_1}{dt} - \cG_\theta(u)_2}_{L^2(\mu)} + \norm{\frac{d\cG_\theta(u)_2}{dt} + \gamma\sin(\cG_\theta(u)_1) - u(t)}_{L^2(\mu)} \leq Cp^{-\beta}.
\end{equation}
\end{theorem}

\begin{proof}
Lemma \ref{lem:pendulum} with $s\leftarrow1$, $k\leftarrow r$ and $m\leftarrow p$ then provides a DeepONet that satisfies the conditions of Theorem \ref{thm:deeponet-to-pido} with $r^*=+\infty$ and equation \eqref{eq:growth} with $\sigma(s)=d/2+2sd$. The smoothness of $v$ is guaranteed by \cite[Lemma 4.3]{LMK1}. Moreover, it holds that, 
\begin{equation}
     \norm{\frac{d\cG_\theta(u)_1}{dt} - \cG_\theta(u)_2}_{L^2(\mu)} \leq \norm{\frac{d\cG_\theta(u)_1}{dt} - \frac{d\cG(u)_1}{dt}}_{L^2(\mu)} + \norm{\cG(u)_2 - \cG_\theta(u)_2}_{L^2(\mu)},
\end{equation}
and also that,
\begin{align}
\begin{split}
&\norm{\frac{d\cG_\theta(u)_2}{dt} + \gamma\sin(\cG_\theta(u)_1) - u(t)}_{L^2(\mu)} \\
     &\qquad\leq \norm{\frac{d\cG_\theta(u)_2}{dt} - \frac{d\cG(u)_2}{dt}}_{L^2(\mu)}+ \gamma\norm{\sin(\cG(u)_1)-\sin(\cG_\theta(u)_1) }_{L^2(\mu)}\\
     &\qquad\leq \norm{\frac{d\cG_\theta(u)_2}{dt} - \frac{d\cG(u)_2}{dt}}_{L^2(\mu)}+ \gamma\norm{\cG(u)_1-\cG_\theta(u)_1 }_{L^2(\mu)}.
\end{split}
\end{align}
Combining this estimate with Theorem \ref{thm:deeponet-to-pido} with $k=2$ then gives the wanted result. 
\end{proof}

\subsection{An elliptic PDE: Multi-d diffusion with variable coefficients}\label{app:darcy}

Next, again following \cite{LMK1}, we consider a popular model problem for elliptic PDEs with unknown diffusion coefficients \cite{CDS2011} and references therein. For the sake of definiteness and simplicity, we shall assume a periodic domain $D = \T^d$ in the following. For $b\in\N_0$, we consider an elliptic PDE with variable coefficients $a$,
\begin{align} \label{eq:random-elliptic}
\cL_a(u) := \nabla \cdot (a(x) \nabla u(x)) + f(x) = 0,
\end{align}
for $u\in C^{b+2}(D)$ with suitable boundary conditions, and for fixed $f \in C^{b}(D)$.
Similar to the previous examples, we fix a probability measure $\mu$ on the coefficient $a$ by assuming that every $a$ can be written as
\begin{align} \label{eq:law-elliptic}
a(x,Y) 
=
\overline{a}(x) 
+
\sum_{{k}\in \Z^d}
\alpha_{k} Y_k \fb_k(x),
\end{align}
with notation from \textbf{SM} \ref{app:Fourier}, and where for simplicity $\bar{a}(x) \equiv 1$ is assumed to be constant. Furthermore, we will consider the case of smooth coefficients $x \mapsto a(x;Y)$, which is ensured by requiring that there exist constants $C_\alpha>0$ and $\ell > 1$, such that $|\alpha_k|\le C_\alpha \exp(-\ell |k|_\infty)$ for all $k \in \Z^d$. Still following \cite{LMK1}, we define $\bm{b} = (b_1,b_2,\dots)\in \ell^1(\N)$ by 
\begin{align} \label{eq:elliptic-b}
b_j := C_\alpha \exp(-\ell |\kappa(j)|_\infty),
\end{align}
where $\kappa: \N \to \Z^d$ is the enumeration for the standard Fourier basis, (\textbf{SM} \ref{app:Fourier}). Note that by assumption on the enumeration $\kappa$, we have that $\bm{b}$ is a monotonically decreasing sequence. In the following, we will assume throughout that $\Vert \bm{b} \Vert_{\ell^1} < 1$, ensuring a uniform coercivity condition on all random coefficients $a = a(\slot; Y)$ in \eqref{eq:random-elliptic}. Finally, we assume that the $Y_j \in [-1,1]$ are centered random variables and we let $\mu \in \cP(L^2(\T^d))$ denote the law of the random coefficient \eqref{eq:law-elliptic}. 

The following lemma provides an error estimate for DeepONets approximating the operator $\cG$ that maps the input coefficient $a$ into the solution field $u$ of the PDE \eqref{eq:random-elliptic}. 

\begin{lemma}\label{lem:elliptic}
For any $k,r \in \N$, there exists a constant $C>0$, such that for any $m,p\in \N$, there exists a DeepONet $\cG_\theta = \cR \circ \cA \circ \cE$ with $m$ sensors, a trunk net $\trunk = (0,\tr_1, \dots, \tr_p)$ with $p$ outputs and branch net $\branch = (0,\beta_1,\dots, \beta_p)$, such that 
\begin{gather}
\size(\branch) \le C(1+pm \log(m)), \quad
\depth(\branch) \le C(1+ \log(m)),
\end{gather}
and
\begin{gather}
\size(\trunk) \le Cp^{\frac{d+1}{d}} \quad
\depth(\trunk) \le 2
\end{gather}
such that the DeepONet approximation error \eqref{eq:approxerr} satisfies
\begin{equation}
\label{eq:elliptic-err}
\Err
\le
C e^{-c\ell m^{\frac{1}{d}}}
+
C m^{-k}
+
C p^{-r},
\end{equation}
and that for all $s\in N$
\begin{equation}
     \abs{\cG_\theta(u)(\cdot)}_{C^s} \leq C p^{s/d}. 
\end{equation}
\end{lemma}

\begin{proof}
This statement is the tanh counterpart of \cite[Theorem 4.19]{LMK1}, which addresses ReLU DeepONets. We only highlight the differences in the proof. First, one should use Lemma \ref{lem:rec-fourier} instead of \cite[Lemma 3.13]{LMK1}, which then results in different network sizes in \cite[Lemma 3.14, Proposition 3.17, Corollary 3.18, Proposition 4.17]{LMK1}. Second, one needs to replace \cite[Proposition 4.18]{LMK1} with Theorem \ref{thm:holomorphic-approx}.

Moreover, in this case the trunk net $\trunk$ approximates the Fourier basis (\textbf{SM} \ref{app:Fourier}). From \eqref{eq:bound-fourier-basis}, one can then deduce the estimate on the $C^s$-norm of the DeepONet. 
\end{proof}

It is straightforward to verify that the conditions of Theorem \ref{thm:deeponet-to-pido} are satisfied in the current setting. Applying Theorem \ref{thm:deeponet-to-pido} then results in the following theorem on the error of physics-informed DeepONets for \eqref{eq:random-elliptic}.

\begin{theorem}\label{thm:pido-elliptic}
Consider the elliptic equation \eqref{eq:random-elliptic} with $b\geq1$. For every $\beta>0$, there exists a constant $C>0$ such that for any $p\in N$, there exists a DeepONet $\cG_\theta$ with a trunk net $\trunk = (0,\tr_1, \dots, \tr_p)$ with $p$ outputs and branch net $\branch = (0,\beta_1,\dots, \beta_p)$, such that 
\begin{gather}
\size(\branch) \le C(1+p^2 \log(p)), \quad
\depth(\branch) \le C(1+ \log(p)),
\end{gather}
and
\begin{gather}
\size(\trunk) \le Cp^2 \quad
\depth(\trunk) \le 2
\end{gather}
such that
\begin{equation}
    \norm{\nabla \cdot (a(x) \nabla \cG_\theta(a)(x)) - f(x)}_{L^2(\mu)} \leq C p^{-\beta}. 
\end{equation}
\end{theorem}

\begin{proof}
We first check the conditions of Theorem \ref{thm:deeponet-to-pido}. Lemma \ref{lem:elliptic} with $s\leftarrow 1$, $k\leftarrow r$ and $m\leftarrow p$ then provides a DeepONet that satisfies the conditions of Theorem \ref{thm:deeponet-to-pido}  with $r^*=+\infty$ and equation \eqref{eq:growth} with $\sigma(s)=s/d$. Moreover, the following estimate holds, 
\begin{align}
\begin{split}
    &\norm{\nabla \cdot (a(x) \nabla \cG_\theta(a)(x)) - f(x)}_{L^2(\mu)}\\
    &\leq \norm{\nabla \cdot (a(x) \nabla \cG_\theta(a)(x)) - \nabla \cdot (a(x) \nabla \cG(a)(x))}_{L^2(\mu)}\\
    &\leq \sum_{j=1}^d \norm{a}_{C^0} \norm{\partial_j^2 (\cG_\theta-\cG)(a)(x)}_{L^2(\mu)} + \sum_{j=1}^d \norm{a}_{C^1} \norm{\partial_j (\cG_\theta-\cG)(a)(x)}_{L^2(\mu)}. 
\end{split}
\end{align}
Combining this estimate with Theorem \ref{thm:deeponet-to-pido} with $k=2$ then gives the wanted result. 
\end{proof}

\end{document}

%% file: custom_definitions.tex
\usepackage{amsthm,amssymb}
\usepackage{mathtools,thmtools}
\usepackage{bm,esint}
\usepackage{color}
\usepackage{float,graphicx,subcaption}
\usepackage{cancel}
\usepackage[shortlabels]{enumitem}
\usepackage{mathrsfs}  
\usepackage{bbm}
\usepackage{euscript}
\usepackage{hyperref}
\usepackage{cleveref}
\usepackage{upgreek}

\usepackage{todonotes}

\newcommand{\poly}{\mathrm{poly}}
\newcommand{\size}{\mathrm{size}}
\newcommand{\width}{\mathrm{width}}
\newcommand{\depth}{\mathrm{depth}}

\newcommand{\fb}{\bm{\mathrm{e}}}
\newcommand{\fbhat}{\widehat{\bm{\mathrm{e}}}}
\renewcommand{\tilde}{\widetilde}
\renewcommand{\hat}{\widehat}

\newcommand{\Id}{\mathrm{Id}}

\newcommand{\Err}{\widehat{\mathscr{E}}}

\newcommand{\embeds}{{\hookrightarrow}}

\renewcommand{\bar}{\overline}

\newcommand{\slot}{{\,\cdot\,}}

\newcommand{\T}{\mathbb{T}}
\newcommand{\R}{\mathbb{R}}

\newcommand{\N}{\mathbb{N}}

\newcommand{\Z}{\mathbb{Z}}
\renewcommand{\S}{\mathscr{S}}

\newcommand{\G}{\mathcal{G}}

\newcommand{\Lip}{\mathrm{Lip}}

\renewcommand{\P}{\mathbb{P}}
\newcommand{\cA}{\mathcal{A}}
\newcommand{\cB}{\mathcal{B}}
\newcommand{\cC}{\mathcal{C}}
\newcommand{\cD}{\mathcal{D}}
\newcommand{\cE}{\mathcal{E}}
\newcommand{\cF}{\mathcal{F}}
\newcommand{\cG}{\mathcal{G}}
\newcommand{\cH}{\mathcal{H}}
\newcommand{\cI}{\mathcal{I}}
\newcommand{\cJ}{\mathcal{J}}
\newcommand{\cK}{\mathcal{K}}
\newcommand{\cL}{\mathcal{L}}
\newcommand{\cM}{\mathcal{M}}
\newcommand{\cN}{\mathcal{N}}

\newcommand{\cP}{\mathcal{P}}
\newcommand{\cQ}{\mathcal{Q}}
\newcommand{\cR}{\mathcal{R}}

\newcommand{\cU}{\mathcal{U}}

\newcommand{\cX}{\mathcal{X}}
\newcommand{\cY}{\mathcal{Y}}
\newcommand{\cZ}{\mathcal{Z}}

\newcommand{\tR}{\tilde{\cR}}

\newcommand{\hg}{\widehat{g}}

\newcommand{\bmalpha}{{\bm{\alpha}}}

\newcommand{\eps}{\epsilon}

\newcommand{\inspace}{\mathcal{X}}
\newcommand{\outspace}{\mathcal{Y}}
\newcommand{\uhat}{{\widehat{u}}}

\renewcommand{\hat}{\widehat}

\renewcommand{\tr}{{{\tau}}}
\newcommand{\trunk}{{\bm{\tau}}}
\newcommand{\branch}{{\bm{\beta}}}
\newcommand{\erre}{\Err_{\cE}}
\newcommand{\errr}{\Err_{\cR}}
\newcommand{\erra}{\Err_{\cA}}

\newcommand{\enum}{{\upkappa}}

\renewcommand{\epsilon}{\varepsilon}

\newcommand{\fu}{Y}
\newcommand{\mlp}{U}
\newcommand{\1}{\mathbbm{1}}
\DeclarePairedDelimiter{\br}{[}{]}
\DeclarePairedDelimiter{\pr}{(}{)}

\newcommand{\Fhat}{\widehat{F}}
\newcommand{\Uhat}{\widehat{U}}

\newcommand{\ghat}{\widehat{g}}

\newcommand{\Psihat}{\widehat{\Psi}}

\newcommand{\hattimes}{\widehat{\times}}

\newcommand{\hatN}{\widehat{N}}

\newcommand{\utilde}{\Tilde{u}}

\newcommand{\cGhat}{{\widehat{\cG}}}

\newcommand{\cQhat}{{\widehat{\cQ}}}

\newcommand{\varphihat}{{\widehat{\varphi}}}

\newcommand{\ck}[1]{\norm{#1}_{W^{k,\infty}}}

\newcommand{\hki}[1]{\norm{#1}_{C^k(I^N_i, L^q(\mu))}}
\newcommand{\hkunit}[1]{\norm{#1}_{C^k([0,T], L^q(\mu))}}

\newcommand{\s}{s} 
\renewcommand{\S}{\mathcal{S}} 
\newcommand{\Et}{\mathcal{E}_T} 
\newcommand{\Eg}{\mathcal{E}_G} 

\newcommand{\bigO}{\mathcal{O}}
\newcommand{\E}[1]{{\mathbb{E}\left[ #1 \right]}} 
\newcommand{\Prob}[1]{{\mathbb{P}\left( #1 \right)}} 

\newtheorem{theorem}{Theorem}[section]
\newtheorem{assumption}[theorem]{Assumption}
\newtheorem{example}[theorem]{Example}
\newtheorem{remark}[theorem]{Remark}
\newtheorem{definition}[theorem]{Definition}
\newtheorem{lemma}[theorem]{Lemma}

\newtheorem{corollary}[theorem]{Corollary}
\newtheorem{setting}[theorem]{Setting}